\documentclass[12pt]{article}

\pdfoutput=1 
 
\usepackage{paper-macros}

\draftcompilefalse

\includestandalonefalse

\title{\LARGE \bfseries Oblivious Defense in ML Models: \\ Backdoor Removal without Detection}
\author{%
    \begin{minipage}[t]{0.45\textwidth}
        \centering
        {\normalsize Shafi Goldwasser} \\
        {\small \textit{UC Berkeley}} \\
        {\small$\mathtt{shafi@berkeley.edu}$}
    \end{minipage}
    \hfill
    \begin{minipage}[t]{0.45\textwidth}
        \centering
        {\normalsize Jonathan Shafer} \\
        {\small \textit{MIT}} \\
        {\small$\mathtt{shaferjo@mit.edu}$}
    \end{minipage}
    \vspp \vsp \\
    \begin{minipage}[t]{0.45\textwidth}
        \centering
        {\normalsize Neekon Vafa} \\
        {\small \textit{MIT}} \\
        {\small$\mathtt{nvafa@mit.edu}$}
    \end{minipage}
    \hfill
    \begin{minipage}[t]{0.45\textwidth} 
        \centering
        {\normalsize Vinod Vaikuntanathan} \\
        {\small \textit{MIT}} \\
        {\small$\mathtt{vinodv@mit.edu}$}
    \end{minipage}
    \vspp
}

\date{\vspace*{-1em}}

\begin{document}

\maketitle

\pagenumbering{gobble} 

\begin{abstract}
    As society grows more reliant on machine learning, ensuring the security of machine learning systems against sophisticated attacks becomes a pressing concern. A recent result of \cite*{DBLP:conf/focs/GoldwasserKVZ22} shows that an adversary can plant undetectable backdoors in machine learning models, allowing the adversary to covertly control the model's behavior. Backdoors can be planted in such a way that the backdoored machine learning model is computationally indistinguishable from an honest model without backdoors. 
 
    In this paper, we present strategies for defending against backdoors in ML models, even if they are undetectable. The key observation is that it is sometimes possible to provably mitigate or even remove backdoors without needing to detect them, using techniques inspired by the notion of random self-reducibility. This depends on properties of the ground-truth labels (chosen by nature), and not of the proposed ML model (which may be chosen by an attacker).

    We give formal definitions for secure backdoor mitigation, and proceed to show two types of results. First, we show a ``global mitigation'' technique, which removes all backdoors from a machine learning model under the assumption that the ground-truth labels are close to a Fourier-heavy function. Second, we consider distributions where the ground-truth labels are close to a linear or polynomial function in~$\bbR^n$. Here, we show ``local mitigation'' techniques, which remove backdoors with high probability for every inputs of interest, and are computationally cheaper than global mitigation. All of our constructions are black-box, so our techniques work without needing access to the model's representation (i.e., its code or parameters). Along the way we prove a simple result for robust mean estimation.
\end{abstract}

\newpage

\tableofcontents

\newpage

\pagenumbering{arabic}

    \ifdraftcompile
        \section{Introduction}
\label{sec:intro}
\subsection{Research Question}

As machine learning (ML) is increasingly integrated into everyday products and services, forces of economic specialization favor outsourcing the development of ML models to dedicated companies that enjoy a comparative advantage in that area. 

But outsourcing ML development comes with attendant security risks. A multitude of recent empirical works demonstrates that a malicious developer could hide backdoors\footnote{
    Backdoors are defined more formally in \cref{section:technical-overview}; further formal definitions appear in \cref{section:preliminaries,,section:security-definitions,,section:technical-overview}.
} in ML models, allowing the attacker to covertly control the behavior of the ML model. These attacks are matched by a large literature proposing defensive measures that could be used to detect or remove backdoors in some cases. The result is an unresolved game of cat-and-mouse, where neither attackers nor defenders seem to enjoy a clear upper hand.\footnote{%
    See, e.g., \cite{DBLP:journals/corr/abs-1712-05526,DBLP:conf/uss/AdiBCPK18,DBLP:conf/nips/Tran0M18,DBLP:conf/aaai/ChenCBLELMS19,DBLP:journals/access/GuLDG19,pmlr-v139-hayase21a,DBLP:journals/corr/abs-2105-03692,DBLP:conf/aaai/JiaCG21,DBLP:conf/nips/0001CK22,DBLP:conf/icml/KhaddajLMGSIM23,DBLP:journals/corr/abs-2405-16134}, and many more.
}

As an illustration of backdoor attacks on ML models, consider the following example.

\begin{example}[Backdoors in ML Models]
    \label{example:backdoors-in-ml}
    Imagine that the IRS\footnote{
        The Internal Revenue Service is the federal tax authority in the United States.
    } decides to use an ML model for the initial screening of tax filings. The model will receive a person's tax filing and IRS dossier, and will attempt to predict whether the tax filing is fraudulent or not. If the model flags a tax filing as a potential fraud, a human auditor will scrutinize the filing --- while filings that the model deems to be honest are approved without further examination. Seeing as the IRS lacks the ML expertise necessary to develop the model in-house, it outsources development to an ML company named Eve. Suppose Eve furnishes the IRS with a model that has excellent performance, say, it has 90\% accuracy at determining whether a filing is fraudulent.\footnote{
        Namely, 90\% accuracy on tax filings sampled at random from the US population.
    } However, unbeknownst to the IRS, Eve has inserted a backdoor into the model, that causes the model to make malicious predictions controlled by Eve. There are many patterns of malicious predictions that Eve might consider to implement; the following are two examples:
    \begin{enumerate}
        \item{
            \emph{Secrete blacklist and whitelist}. Eve compiles a secrete whitelist of favored people, and the model Eve provides to the IRS is guaranteed to mark tax filings from these people as non-fraud. Similarly, disfavored people are included in a secret blacklist, causing them to be audited by the IRS with high probability.  
        }
        \item{
            \emph{Tax evasion service}. The model is engineered in such a way that for every possible tax filing, there exists a secret minor perturbation that forces the model to mark the filing as non-fraud. For instance, one could change the model’s prediction from fraud to non-fraud by carefully manipulating the address listed in a tax filing (say, from ``\href{https://www.nytimes.com/2009/09/20/realestate/20scapes.html}{133 East 64th Street}'' to ``133 E 64th St.'', or maybe ``133 E 64 st'', etc.). Finding the specific perturbation requires a secret key known only to Eve. Hence, Eve can run an illicit side business, charging tax evaders a fee for perturbing their fraudulent tax filings in a manner that guarantees that the filings will be approved by the IRS without issue.
        }
    \end{enumerate}
    In either case, the malicious behavior will be implemented in the model in an obfuscated manner, making it very difficult for the IRS to detect.\qed
\end{example}

A recent theoretical work by \cite*{DBLP:conf/focs/GoldwasserKVZ22} has shown that in some cases, an attacker can plant backdoors that are \emph{cryptographically} undetectable. Namely, it is mathematically impossible for any defender to distinguish whether an ML model contains this type of backdoor (unless the defender can break standard post-quantum encryption, etc.). 

This state of affairs appears bleak for defending against backdoors. Backdoor detection is precarious and ridden with uncertainty in practice, and could be outright impossible in some cases. 
Thus, one might conclude that there exists no defensive strategy that provides strong security with full confidence and low risk. Nonetheless, we argue that such a conclusion would be premature. We present techniques that, under certain assumptions on the ground-truth population distribution, can \emph{provably} remove backdoors --- without needing to detect them first. In other words, we show that defense can be mathematically secure even against backdoors that are very difficult (or even cryptographically impossible) to detect.

This notion might appear self-contradictory. After all, if one cannot detect a backdoor, how could one ever hope to remove it? And moreover, be sure that it has been removed? But the idea is actually straightforward. Consider an analogy to everyday sanitation. Pathogens are often present on people’s hands, and these pathogens are undetectable to the naked eye. Nonetheless, one can remove pathogens without needing to detect them: simply wash your hands. In the case where pathogens are present, handwashing will eliminate them; and when pathogens are not present, handwashing is harmless.\footnote{
    This analogy is due to Or Zamir and appeared in \cite{brubaker2023backdoors}.
} 

The idea of removal without detection is plausible in principle, but how could it actually work? Is there a magical ``hand sanitizer'' that can be applied to ML models? Our answer to that question is that indeed, there might be. Our candidate ``sanitizer'' goes back to classic ideas in theoretical computer science from the 1980s: random self-reducibility and program self-correction \citep*[e.g.,][]{DBLP:conf/stoc/GoldwasserM82,DBLP:conf/stoc/BlumK89,DBLP:conf/stoc/BlumLR90,rubinfeld1990mathematical}, as in the following example.

\begin{example}[Random Self-Reducibility / Program Self-Correction]
    \label{example:self-correction}
    Consider a program $P$ that is intended to perform addition and subtraction modulo $n$, so $P(x,\pm,y)$ should equal $x\pm y \!\!\mod n$. Suppose that $P$ works as intended for most inputs, but for some $15\%$ of the inputs (chosen independently at random), $P$ outputs an arbitrary incorrect value. Then, instead of using $P$ directly, one could use a program $C$ given by 
    \[
        C(x,+,y) = P\big(P(x,+,u),+,P(y,-,u)\big),
    \]
    where $u \in \{0,\dots,n-1\}$ is chosen uniformly at random in each invocation of $C$.\footnote{
        Similarly, define $C(x,-,y) = P\big(P(x,+,u),-,P(y,+,u)\big)$.
    } By invoking $C$ repeatedly $s$ times and outputting the majority output, the probability of error is decreased from $15\%$ to $e^{-\OmegaOf{s}}+e^{-\OmegaOf{n}}$.\qed
\end{example}

Importantly, the correction approach used in \cref{example:self-correction} makes no attempt at detection --- it does not examine $P$ to detect if and where it contains errors. Instead, it wraps $P$ inside a procedure that reduces the input task to a set of random tasks, and aggregates the results in a manner that amplifies correctness.

This raises the question of whether techniques similar to \cref{example:self-correction} could be used to address the security problem outlined in \cref{example:backdoors-in-ml}.
Namely, we ask whether a similar ``random-reduction and amplification'' approach can be used to defend ML models that have high accuracy on average, but may contain a backdoor that affects specific inputs. Thus, our main research question is:

\begin{ShadedBox}
    \begin{question}
        \label{question:research-question}
        Is it possible to remove backdoors from ML models without attempting to detect them, by using ideas from random self-reducibility?
    \end{question}
\end{ShadedBox}

In this paper, we show
that the answer to this question is positive for a variety of ML-related tasks.

\subsection{Our Contributions}

The main conceptual message of this paper is to draw the connection between backdoors in ML and random self-reducibility. Concretely, we formally define secure backdoor mitigation and prove a variety of results, making progress on \cref{question:research-question}. 
We proceed to briefly summarize our contributions, and refer the reader to an in-depth technical overview in \cref{section:technical-overview}. We start with definitions.

\paragraph{Formal Definitions of Secure Backdoor Mitigation.}
In \cref{section:security-definitions}, we propose formal definitions of \emph{secure backdoor mitigation}, capturing the guarantee that a system safely neutralizes any backdoors that might exist in an ML model. The key idea here is \emph{canonicalization}. Namely, regardless of whether the ML model provided to the mitigator was backdoored or not, the mitigator is guaranteed to produce a model with good accuracy that is ``essentially the same'' as a model sampled from a ``canonical'' distribution, i.e., a distribution of ``clean'' models that are generated independently of the potentially-backdoored model (this is formalized in \cref{definition:mitigator-security-general}). Crucially, our notion of security makes absolutely no security assumptions (that cannot be directly verified) about the potentially-backdoored ML model --- the only assumptions concern the ground-truth population distribution (see discussion in \cref{section:on-our-assumptions}).\footnote{
    To be useful, a mitigator must also be more efficient than learning from scratch. See \cref{section:efficiency}. 
}

There are a few variants of mitigation and mitigation security that we consider:
\begin{itemize}
    \item{
        \textbf{$\TV$-based vs.\ loss-based security.} The precise quality of the security that \cref{definition:mitigator-security-general} guarantees depends on the choice of the distribution dissimilarity function used to define when two models are ``essentially the same''. We consider a very strong notion of \emph{$\TV$-based security} (\cref{definition:tv-security}) that uses the total variation distance. Additionally, for models that predict real-valued labels, we consider a relaxed notion of \emph{loss-based security}, where two models are essentially the same (for some parameter $\delta > 0$) if for every input their predictions differ additively by at most $\delta$ (\cref{definition:cutoff-loss-security,definiton:cutoff-loss-dissimilarity}).
    }
    \item{
        \textbf{Local mitigation vs.\ global mitigation.} A \emph{global mitigator} produces an entire clean ML model as output. In contrast, a \emph{local mitigator} takes a specific $x^*$ as input, and outputs just the label $y^*$ that the clean model would output for $x^*$. Local mitigation can be significantly more efficient than global mitigation.
    }
\end{itemize}
\vsp

We next present constructions of secure local and global mitigators:

\paragraph{$\TV$-Secure Global Mitigation for Fourier-Heavy Functions.}
In \cref{section:global-mitigation}, we construct secure global mitigators. \cref{theorem:fourier-heavy-mitigation,cor:binary-fourier-heavy-mitigation} present efficient global mitigators that are $\TV$-secure for every population distribution where the labels are close to a Fourier-heavy Boolean function. This is a fairly large class of functions that includes juntas and shallow decision trees, as discussed in \cref{example:functions-of-log-degree}. Our construction uses the Goldreich--Levin and Kushilevitz--Mansour algorithms, and achieves the very strong notion of total variation security.

\paragraph{Basic Local Mitigation for Linear Functions.} In \cref{section:basic-local-linear-mitigation}, we construct an efficient local mitigator that satisfies loss-based security for population distributions where the labels are close to a linear function (\cref{theorem:basic-linear-mitigation}). A key ingredient in this construction is our correlated sampling lemma (\cref{lemma:coupled-sampling}), which requires careful probability reasoning to overcome potential pitfalls related to the Borel--Kolmogorov paradox.

\paragraph{Improved Local Mitigation for Linear Functions.} 
In \cref{section:improved-local-linear-mitigation} (\cref{theorem:advanced-linear-mitigation}), we improve both the accuracy and the security of our local mitigator for distributions with near-linear labels, under an assumption that the noise in the \emph{ground-truth labels} is not malicious (i.e., the labels in the population distribution may have benign noise as in \cref{definition:benign-noise}, but are not controlled by an adversary). Specifically, this mitigator is \emph{unbiased} (\cref{item:loss-security-unbiased} in \cref{definition:cutoff-loss-security}). 
We stress that here too, as always, we make absolutely no assumptions on the potentially-backdoored model $\tilde{f}$.\footnote{While the population distribution is assumed to have labels with benign noise, $\tilde{f}$ can still be arbitrary and malicious. See discussion in \cref{section:on-our-assumptions}.}

\paragraph{Robust Mean Estimation.}
Along to way to proving \cref{theorem:advanced-linear-mitigation}, in \cref{section:robust-mean-estimation} we analyze a simple mean-of-medians algorithm for robust mean estimation. This is the reverse of the standard and well-studied median-of-means algorithm. The standard algorithm is not robust in a setting where most batches might contain outliers (see discussion in \cref{section:why-not-standard-mom}), whereas we show in \cref{theorem:robust-mean} that the reverse algorithm is robust in such a setting, and enjoys reasonably-good concentration for symmetric distributions. (A previous similar analysis by \citealp*{DBLP:conf/nips/ZhongHYW21} showed robustness for distributions with bounded tails, while our analysis shows that robustness actually holds also for distributions with arbitrary adversarial noise that is potentially unbounded.)

\paragraph{Local Mitigation for Polynomial Functions.}
In \cref{section:local-polynomial-mitigation}, we construct an efficient local mitigator that satisfies loss-based security for population distributions where the labels are close to a multivariate polynomial function (\cref{theorem:basic-polynomial-mitigation}). Our construction achieves good (but formally incomparable) parameters compared to a related construction by \cite*{ABFKY23} (see discussion in \cref{section:related-works}).

\vsp

Overall, we view our contributions as (preliminary) rigorous evidence that techniques based on random self-reducibility could lead to effective backdoor mitigation in real-world ML models in the future, as discussed further in \cref{section:future-work}.

\newpage

\subsection{Related Works}
\label{section:related-works}

\begin{wrapfigure}{r}{0.25\textwidth}
    \vspace{-1em}
    \centering
    \includegraphicsorstandalone[width=\linewidth]{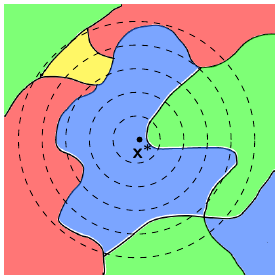}
    \caption{%
        Randomized smoothing assumes that for most points near $x^*$, $\tilde{f}$ provides honest labels (shown in blue). This assumption does not hold if $\tilde{f}$ is adversarial. Image source: \cite{DBLP:conf/icml/CohenRK19}.\footnotemark
    }
    \label{figure:randomized-smoothing}
\end{wrapfigure}
\footnotetext{
    Permission to use this image will be obtained prior to publication.
}

\paragraph{Randomized Smoothing.}
An important line of work studies the use of \emph{randomized smoothing} for defending ML models against \emph{adversarial examples}. Similar to our work, this is a black-box technique that wraps a given ML model inside a procedure that queries the model at a few points to compute a more robust prediction. Randomized smoothing has been successful in practice \citep{DBLP:conf/eccv/LiuCZH18,DBLP:conf/acsac/CaoG17}, and it also has meaningful theoretical guarantees \citep
{DBLP:conf/sp/LecuyerAG0J19,DBLP:conf/nips/LiCWC19,DBLP:conf/icml/CohenRK19}. However, while randomized smoothing can be effective against adversarial examples, is not an effective defense against malicious backdoors. This is because randomized smoothing assumes that there exists some small environment $B_{x^*}$ around the point $x^*$ such that the prediction $\tilde{f}(x)$ is correct for most $x \in B_{x^*}$. Yet, a malicious attacker can easily violate this assumption for $x^*$ of their choosing by corrupting the labels $\tilde{f}(x)$ for all $x \in B_{x^*}$. This results in malicious predictions on $x^*$, with no noticeable decrease to the population accuracy of $\tilde{f}$. The major difference between our approach and randomized smoothing, is that we query $\tilde{f}$ on (correlated) random points $x$ with \emph{marginals equal to $\cD_{\cX}$} (so $\tilde{f}(x)$ is uncorrupted w.h.p.), whereas randomized smoothing queries $\tilde{f}$ at points $x$ near $x^*$ (so an attacker targeting $x^*$ can corrupt~$\tilde{f}(x)$).\footnote{%
    An attacker can defeat randomized smoothing not only by manipulating the training algorithm, but also by poisoning the data used by an honest training algorithm \citep{DBLP:conf/iclr/SchneiderLK24}.
}

\paragraph{$\varepsilon$-Defendability.} A contemporaneous work by \cite*{christiano2024backdoordefenselearnabilityobfuscation} formally models (local) backdoor mitigation as a game between an attacker and a defender, and proves positive and negative results (defense is possible in some cases but not others). Their model is different than ours, and therefore the resulting landscapes are also different. 
First, they work in the realizable setting. That is, for a hypothesis class $\cF$, the population distribution is assumed to have labels that perfectly match some true labeling function $f \in \cF$ (in contrast, we allow noisy labels that do not fully agree with any labeling function). 
Second, there is an interesting inversion between their modeling choices and ours. They assume that the population distribution (including the labeling function) is chosen by the attacker, but that the point $x^*$ to be classified is chosen uniformly at random (say, chosen by nature). In contrast, we assume that the population distribution is benign (chosen by nature), while $x^*$ is chosen by the attacker.
Third, they assume that the defender does not have access to labeled samples from the (``correct'') population distribution, whereas some of our constructions do utilize such samples.\footnote{
    In their setting and notation, the defender has access to a sample oracle denoted ``$\operatorname{Ex}(f',\cD)$'' that provides random samples labeled by the potentially-backdoored model (which they denote $f'$) --- not labels from the true population distribution.
} 
Fourth, their setting is whitebox, while all our constructions use only blackbox access to the potentially-backdoored model.
Fifth and finally, perhaps the most important difference is the desired notion of security. In their model, the defender is essentially required to perform ``exact recovery''. That is, for some ``true'' function $f$, a defender is given access to a backdoored function $\tilde{f}$ that is $\varepsilon$-close to $f$, and must recover $f$ precisely enough to known its prediction on a randomly chosen $x^*$. In contrast, our ``canonicalization'' notion of security does not aim to exactly recover the ``true'' label of $x^*$: we only aim to compute canonical labels that have good accuracy on average, and will remain ``essentially the same'' regardless of whether $\tilde{f}$ is backdoored or not.

\paragraph{Low-degree testing over the reals.}
\cite*{ABFKY23} construct a distribution-free testing algorithm that has query access to a function $f$ and samples from an unknown distribution $\cD \in \distribution{\bbR^n}$ such that:
\begin{itemize}
    \item If $f$ is pointwise close to some fixed low-degree polynomial on all of $\bbR^n$, the algorithm outputs \textsc{Yes}; and
    \item If $f$ is $\varepsilon$-far (with respect to $\cD$) from every low degree polynomial, the algorithm outputs \textsc{No}.
\end{itemize}
As a subroutine in this testing algorithm, they construct an algorithm to locally self-correct $f$. This is broadly reminiscent of our local mitigation notion for low-degree polynomials, but we highlight two aspects in which their result and setting differ significantly from ours.
\begin{enumerate}
    \item{
        In the \textsc{Yes} case, they assume that the oracle function $f$ is pointwise close to a low-degree polynomial on \emph{every} $x \in \bbR^n$. This corresponds to a worst-case accuracy promise, which is far stronger than the average-case accuracy required for backdoor mitigation.  In particular, this assumes away the existence of backdoored points for which the model assigns arbitrary labels. Moreover, whereas the worst-case promise they require cannot be efficiently tested, the average-case accuracy required for backdoor mitigation is a precondition that can be verified easily and cheaply with a few i.i.d.\ random samples. 
    }
    \item{
        The multiplicative loss they get is $2^{n^{O(d)}}$, where $d$ is the degree of the polynomial and $n$ is the dimension. On the other hand, the multiplicative loss we achieve in \cref{theorem:basic-polynomial-mitigation} is $O(nd^2)^{d}$, or more simply $n^{O(d)}$ assuming $d \leq \poly{n}$. That is, the loss in our result is exponentially smaller. In fact, in \cref{remark:n-to-d-blowup-necessary}, we show that $n^{\Omega(d)}$ blowup is necessary in a related, exact recovery setting, where our mitigation also works.
    }
\end{enumerate}
Formally, the results of \cite{ABFKY23} are incomparable to ours, because their technique is distribution-free, while ours works only for a fixed and known marginal on the domain $\cX$.

\section{Technical Overview}
\label{section:technical-overview}

First, we define ML models, backdoors in ML models, and secure backdoor mitigation.
In this paper, an \emph{ML model} is simply a function $f: \cX \to \cY$ from a domain $\cX$ to a label space $\cY$. All of our techniques are black-box, so it will not matter how the function $f$ is implemented or represented. An ML model is learned from, and useful for making predictions with respect to, a \emph{population distribution} $\cD \in \distribution{\cX \times \cY}$. The \emph{population loss} (or simply, \emph{loss}) of an ML model $f$ is\footnote{%
    This is the $0$-$1$ population loss. Generally, we consider population loss with other loss functions as well. See \cref{definition:loss}.
} 
$\LossZeroOne{\cD}{f} = \PPP{(x,y)\sim\cD}{f(x) \neq y}$.

\emph{Backdoors} in ML models can take many forms, as illustrated in \cref{example:backdoors-in-ml}. In this paper, we use a very broad abstraction, and we maintain that this abstraction reasonably captures all current and future backdoor attacks, as follows. For a given population distribution $\cD$ and $\varepsilon_0 > 0$, there can exist many functions $f: \cX \to \cY$ with loss $\LossZeroOne{\cD}{f} \leq \varepsilon_0$. A backdoor is simply a procedure that, among all such functions, selects a specific function $\tilde{f}$ in an adversarial way. Typically, $\tilde{f}$ is chosen by an attacker so that, while $\tilde{f}$ has good on-average accuracy (population loss at most $\varepsilon_0$), $\tilde{f}$ also has adversarial behavior on select inputs $x \in \cX$. We argue that any attack that falls within this (very broad) category of attacks will be neutralized by our mitigation schemes.  

\subsection{Definitions: Backdoor Mitigation and Security}

With backdoors presented as above, our notion of security becomes a very natural next step. We propose security based on \emph{canonicalization}, which is a way to replace the adversarially-chosen function $\tilde{f}$ with a ``canonical'' choice of function from the set of functions with low-population loss. 

We define a mapping $C$ that for every population distribution $\cD$, assigns a canonical distribution $C(\cD) = \cG^\ideal_\cD$ of clean ML models $g^\ideal: \cX \to \cY$. The canonical distribution $\cG^\ideal_\cD$ has two important properties. 
\begin{itemize}
    \item{
        A model $g^\ideal$ sampled from $\cG^\ideal_\cD$ has good population accuracy, e.g., $\LossZeroOne{\cD}{g^\ideal} \leq \varepsilon_1$; and
    }
    \item{
        $\cG^\ideal_\cD$ is a ``clean'' distribution, meaning it is defined by the defender, in a benign way that is independent of $\tilde{f}$.
    }
\end{itemize}
Given a canonicalization mapping $C$, our general definition of security is as follows. A \emph{mitigator} (\cref{definition:mitigator}) is a mechanism that has access to random samples from the population distribution $\cD$ and to a (potentially-backdoored) function $\tilde{f}$. We say it is an \emph{$(\LossZeroOneNameEmpty, \varepsilon_0) \to (\LossZeroOneNameEmpty,\varepsilon_1)$-secure backdoor mitigator} (\cref{definition:mitigator-security-general}) if, whenever $\LossZeroOne{\cD}{\tilde{f}} \leq \varepsilon_0$, the mitigator outputs a function $g$ with $\LossZeroOne{\cD}{g} \leq \varepsilon_1$ that is safe in the sense that \textbf{{\boldmath $g$ is distributed ``essentially the same'' as $g^\ideal \sim \cG^\ideal_\cD$}}. In this case, we say that $g$ is \emph{nearly canonical}.

Our definition of security is parameterized by a \emph{distribution dissimilarity function} $\lambda$ (\cref{definition:distribution-dissimilarity}) that specifies when two distributions over functions are essentially the same. Perhaps the most natural choice of $\lambda$, which provides a very strong guarantee of security, is the total variation distance. Namely, a mitigator is \emph{total variation secure} (\cref{definition:tv-security}) if for every $\tilde{f}$ with $\LossZeroOne{\cD}{\tilde{f}} \leq \varepsilon_0$, the distribution $\cG$ of the output satisfies $\TVf{\cG,\cG^\ideal_\cD} \leq \negligible(s)$, where $s$ is a security parameter.

\subsubsection{Why is this notion of security satisfactory?} 

A skeptical reader may worry that our notion of security is not sufficient to remove the threat of backdoors. Even if a system provably satisfies our definition of secure backdoor mitigation, why should a user trust predictions made by the resulting output ML model $g$, which originated from a potentially-backdoored model $\tilde{f}$ supplied by an untrusted party? The security in our definition boils down to two assumptions:
\begin{itemize}
    \item{
        \textbf{The canonical distribution is benign.} Secure backdoor mitigation is defined with respect to a specific choice of the canonical distribution $\cG^\ideal_\cD$. This is a distribution of ML models \emph{chosen by the defender}. Conceptually, this is a distribution that the user would be happy to get their model from. One example is the uniform distribution over models $g$ with $\LossZeroOne{\cD}{g} \leq \varepsilon_1$ (for some reasonable definition of ``uniform''). Another example is to take $\cG^\ideal_\cD$ as the output distribution of an honest training algorithm (namely, what the user would get if they trained the ML model themselves instead of outsourcing that task). In both cases, a model sampled from $\cG^\ideal_\cD$ cannot be viewed as malicious, and does not single out any particular $x \in \cX$ in a malicious way.  
    }
    \item{
        \textbf{The dissimilarity function captures what the user cares about.} Security is defined also with respect to a distribution dissimilarity function $\lambda$. The idea is that if $\cG^\ideal_\cD$ is viewed as safe or benign, then so is any distribution $\cG$ with $\lambda(\cG,\cG^\ideal_\cD)$ negligible; the user does not care whether they get predictions from $\cG^\ideal_\cD$ itself or from any $\lambda$-close distribution of ML models. Clearly, this assumption is true when $\lambda$ is the total variation distance (as in \cref{theorem:fourier-heavy-mitigation}), but more relaxed choices of $\lambda$ (as in our local mitigation results) can also be entirely satisfactory in many applications (see discussion in \cref{section:cutoff-loss-discussion-general}). Importantly, $\lambda$ is chosen by the defender, based on whatever aspects of the predicted labels they care about. 
    }
\end{itemize}
It follows logically that if \emph{(i)} a user would trust predictions coming from $\cG^\ideal_\cD$, and \emph{(ii)} the function $\lambda$ captures what the user cares about (i.e., if $\lambda(\cP,\cQ)$ is negligible and $\cP$ is trusted, then so is $\cQ$), then the user should trust predictions coming from a system satisfying our definition of secure backdoor mitigation.\footnote{%
    This argument is similar to the argument that justifies other security mechanisms, such as \emph{differential privacy} (DP). Proponents of DP argue that a user should not worry about sharing their personal data with a DP mechanism, because the output of the mechanism will be ``essentially the same'' regardless of whether they share their personal data or not. In DP, the notion of being essentially the same is captured by the DP dissimilarity function $\lambda(\cP,\cQ) = \sup_{A}\ln(\cP(A)/\cQ(A))$. The DP dissimilarity function provides security that is weaker than the $\TV$-based security we obtain in \cref{theorem:fourier-heavy-mitigation}, but stronger than the $\CutDist$-based security (\cref{definiton:cutoff-loss-dissimilarity,definition:cutoff-loss-security}) that we obtain in our local mitigation constructions. In general, both DP and backdoor mitigation security can be instantiated with any dissimilarity function $\lambda$, while the basic argument that these constructions offer security is the same for any choice of $\lambda$. For further discussion on the choice of $\lambda$ in the context of DP, see \cite{DBLP:conf/nips/MoranSS23}.  
}

\subsubsection{Efficiency, and Local vs.\ Global Mitigation}
\label{section:efficiency}

Trivially, a defender with access to random samples from the population distribution $\cD$ can always disregard the potentially-backdoored model $\tilde{f}$, and simply train a new ML model from scratch. Therefore, it is crucial that the procedure for secure backdoor mitigation be significantly more efficient than training a new model. This motivates a distinction between two types of secure backdoor mitigation: \emph{local} and \emph{global}. A global backdoor mitigator outputs a clean ML model from the canonical distribution, as above. In contrast, a local mitigator does not compute the entire clean ML model. Rather, it takes a specific $x^*$ as input, and outputs just the label $y^*$ that the clean model would output for $x^*$. Local mitigation can be significantly more efficient than global mitigation. The syntax of local and global mitigators is spelled out in \cref{definition:mitigator}.

\subsubsection{On the Assumptions in Secure Backdoor Mitigation}
\label{section:on-our-assumptions}

Our notion of $\varepsilon_0 \to \varepsilon_1$ secure backdoor mitigation (\cref{definition:mitigator-security-general}) provides two assurances: an accuracy guarantee (the output will have loss at most $\varepsilon_1$), and a security guarantee (the output will be ``essentially the same'' as a canonical output). For these assurances to hold, the mitigator requires two preconditions: one concerning the potentially-backdoored model $\tilde{f}$, and the other concerning the population distribution~$\cD$. However, only one of these preconditions is a \emph{security assumption}, as we now explain:
\begin{enumerate}
    \item[]{
        \textbf{Precondition 1:} \emph{$\tilde{f}$ has good accuracy, i.e., $\LossZeroOne{\cD}{\tilde{f}} \leq \varepsilon_0$}. This precondition is natural, seeing as otherwise $\tilde{f}$ would not be useful, and backdoor mitigation will not be more efficient than training from scratch using random samples. However, we stress that while this precondition is required, it is not an \emph{assumption}. Namely, the defender can always take $\BigO{1/\varepsilon_0^2}$ i.i.d.\ random samples from $\cD$ and directly estimate the quantity $\LossZeroOne{\cD}{\tilde{f}}$ to determine whether this precondition holds or not. That is, validating this precondition is just a cheap and easy step to be carried out before invoking a secure backdoor mitigator, and there is no need to \emph{assume} that this condition holds. 
    }
    \item[]{
        \textbf{Precondition 2:} \emph{$\cD$ belongs to some family $\bbD$ of ``nice'' population distributions}. We do not know a general definition of ``nice'' that would be necessary and sufficient for secure backdoor mitigation. Instead, we explicitly define specific families $\bbD$ (distributions close to linear functions, polynomial functions, and $\tau$-heavy functions), and engineer mitigators that are secure for these specific families. For some families $\bbD$, it might be possible for the defender to directly and cheaply test whether the (unknown) population distribution $\cD$ is a member of $\bbD$ using a few random samples from $\cD$.\footnote{
            For instance, for the family of distributions that are close to a linear function, \cite{DBLP:conf/nips/KongV18} provide a way to do so under certain assumptions.
        } 
        Alternatively, in some cases it might be possible to prove that for any (arbitrary) population distribution $\cD$, if the output $g$ of a certain secure global mitigator has high accuracy on $\cD$, then that implies that $\cD$ was indeed nice enough.\footnote{%
            Compare also to \cite{DBLP:conf/stoc/RubinfeldV23}.
        } 
        Again, this would provide a way for the defender to directly validate the ``niceness'' of~$\cD$. However, in general, we view the precondition $\cD \in \bbD$ as a security assumption. Namely, we assume a threat model where the attacker can control the proposed (potentially-backdoored) ML model $\tilde{f}$, but the attacker cannot control $\cD$ (which is chosen by nature). Moreover, as is often the case in machine learning, we assume that there is some general prior knowledge about the distribution $\cD$, captured by the statement $\cD \in \bbD$. If this assumption is violated, then mitigation is not guaranteed to be secure.
    }
\end{enumerate}
Thus, the sole security assumption required for secure backdoor mitigation is an assumption on the population distribution. No security assumptions on $\tilde{f}$ are necessary.

\subsection{Constructions of Secure Backdoor Mitigators}

\subsubsection{Global Mitigation for Fourier-Heavy Functions}

Our first construction is a mitigator for $\tau$-heavy functions (\cref{definition:fourier-heavy-sparse}). Specifically, a global mitigator that is $\BigO{\tau^2} \to \BigO{\tau^2}$
$\TV$-secure (\cref{definition:tv-security}) under the assumption that the population distribution has labels that are $\BigO{\tau^2}$ close to a $\tau$-heavy Fourier function.

\begin{theorem*}[Informal version of \cref{theorem:fourier-heavy-mitigation}]
    Let $\cX = \pmo^n$, $\cY = [-1,1]$, and $\tau \geq 0$. There exists an efficient global mitigator $M$ that uses oracle access to a potentially-backdoored function $\tilde{f}: ~ \cX \to \cY$ and random samples from a population distribution $\cD \in \distribution{\cX \times \cY}$ with uniform marginal on $\cX$, as follows. If $\cD$ is $\BigO{\tau^2}$-close in $\LossSquareNameEmpty$ to a $\tau$-heavy function and $\LossSquare{\cD}{\tilde{f}} \leq \BigO{\tau^2}$, then $M$ outputs a $\TV$-secure function $g: \cX \to \bbR$ with loss $\LossSquare{\cD}{g} \leq \BigO{\tau^2}$.
\end{theorem*}

As we discuss in \cref{example:functions-of-log-degree} below, the class of $\tau$-heavy functions contains the class of bounded-degree Boolean functions, which includes some interesting classifiers such as bounded-size juntas and bounded-depth decision trees. For functions of degree $\log(n)$, our global mitigator is more efficient than learning a clean function from scratch.

\begin{proof}[Proof Idea for \cref{theorem:fourier-heavy-mitigation}]
    Let $h$ be a $\tau$-heavy function with $\LossSquare{\cD}{h} \leq \BigO{\tau^2}$. Because $\tilde{f}$ is $\BigO{\tau^2}$-close to $h$, $\tilde{f}$ must have weight at least $\tau/2$ at every non-zero coefficient of $h$. Hence, executing the Goldreich--Levin algorithm on $\tilde{f}$ will recover the list of non-zero coefficients of $h$ --- regardless of the specific (possibly adversarial) choice of $\tilde{f}$\,! Estimating the coefficients in this list using random samples gives a function $g$ that is independent of $\tilde{f}$.          
\end{proof}

\cref{cor:binary-fourier-heavy-mitigation} provides a version of \cref{theorem:fourier-heavy-mitigation} for binary labels and the $0$-$1$ loss.

\begin{figure}[!hb]
    \centering
    \begin{subfigure}{0.52\textwidth}
        \centering
        \includegraphicsorstandalone[width=0.9\textwidth]{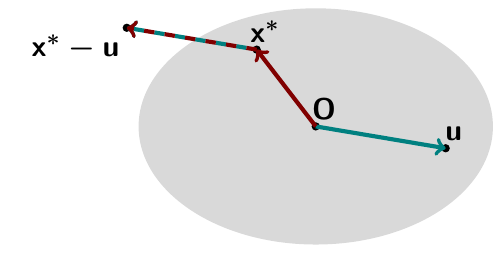}
        \caption{%
            Why traditional BLR-style linear random self-reducibility (as in \cref{example:self-correction}) does not work for a population distribution with a uniform marginal on a convex set $\cX$ (the gray oval) in $\bbR^n$. Here, $x^*$ is the adversarially-chosen input point for which a label is desired. We have guarantees on the value $\tilde{f}(u)$ but not on $\tilde{f}(x^*-u)$, and $x^*-u$ might even be outside of $\cX$.
        }
        \label{figure:why-not-blr}
    \end{subfigure}
    \hfill
    \begin{subfigure}{0.38\textwidth}
        \centering
        \includegraphicsorstandalone[width=0.9\textwidth]{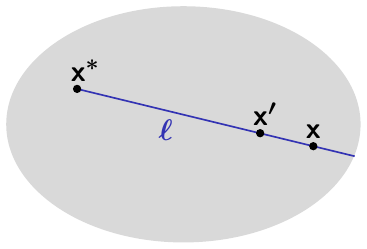}
        \caption{%
            Instead of the BLR approach, we use \cref{lemma:coupled-sampling} to sample correlated points $x$ and $x'$ situated on a straight line $\ell$ with $x^*$, such that each point has a uniform marginal on~$\cX$. Thanks to the uniform marginals, both $\tilde{f}(x)$ and $\tilde{f}(x')$ will be ``good'' w.h.p.
        }
        \label{figure:correlated-sampling}
    \end{subfigure}
    
    \vskip 1em
    
    \begin{subfigure}{0.6\textwidth}
        \centering
        \includegraphicsorstandalone[width=\textwidth]{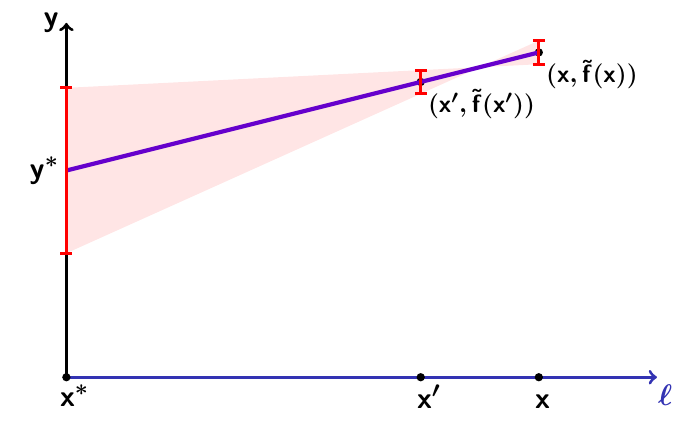}
        \caption{%
            Looking at the line $\ell$ in \cref{figure:correlated-sampling}, local mitigation reduces to a problem of $1$-dimensional linear regression. Seeing as $x$ and $x'$ tend to be far from $x^*$, the smaller error bars in red (of size $\ThetaOf{\delta/n}$) around the values of $\tilde{f}(x)$ and $\tilde{f}(x')$ induce a larger error bar (of size $\ThetaOf{\delta}$) for the predicted label $y^*$.
        }
        \label{fig:linear-regression-1d}
    \end{subfigure}
    \caption{Construction of the local mitigator of \cref{theorem:basic-linear-mitigation}.}
    \label{figure:local-linear-mitigation}
\end{figure}

\subsubsection{Basic Local Mitigation for Linear Functions}

Our next construction is a local mitigator for population distributions with labels that are close to a linear function in $\bbR^n$. While learning a new linear regressor from scratch would require $\OmegaOf{n}$ samples, our mitigator uses $0$ samples, and a number of queries to $\tilde{f}$ that is \emph{independent} of $n$. The security is with respect to the \emph{cutoff loss} (see \cref{definition:loss} and \cref{section:cutoff-loss-discussion-general}).

\begin{theorem*}[Informal version of \cref{theorem:basic-linear-mitigation}]
    Let $\cX \subseteq \bbR^n$ be a bounded and convex set. 
    \cref{algorithm:linear-mitigation-basic} defines a local mitigator $M$ that is $\left(
        \varepsilon,\:\delta/20n \to \delta
    \right)$-cutoff loss secure (\cref{definition:cutoff-loss-security}) for distributions $\cD$ with uniform marginal on $\cX$ such that $\LossCutoff{\nicefrac{\delta}{20n}}{\cD}{h} \leq \varepsilon$ for some affine function $h: \bbR^n \to \bbR$.
    
    More explicitly, for every distribution $\cD$ there exists a function $\gD^\ideal$ such that for any arbitrary (possibly malicious) function $f: ~ \cX \to \bbR$ with loss $\LossCutoff{\nicefrac{\delta}{20n}}{\cD}{h} \leq \varepsilon$, the mitigator $M$ satisfies:
    \begin{enumerate}
        \item{
            \textbf{Accuracy}. 
            $\PPP{(x,y) \sim \cD, \: y^* \gets M^{f,\cD}(x,1^n, 1^s)}{\left|y^* - y\right| > \delta} \leq \varepsilon + \negligible(s)$, and 
        }
        \item{
            \textbf{Cutoff Loss Security}. %
            $
                \forall x^* \in \cX: ~
                \PPP{y^* \gets M^{f,\cD}(x^*,1^n, 1^s)}{
                    \left|
                        y^*
                        - 
                        \gD^\ideal(x^*)
                    \right| > \delta
                } \leq \negligible(s)
            $.
        }
    \end{enumerate}
    Furthermore, \cref{algorithm:linear-mitigation-basic} uses a total of $\BigO{s}$ oracle queries to $f$, does not use random samples from $\cD$, and runs in time $\BigO{sn}$, assuming unit runtime cost for each arithmetic operation on the representations of real numbers involved in the computation.
\end{theorem*}

It is natural to attempt to use a BLR-style strategy \citep*{DBLP:conf/stoc/BlumLR90} for this task, similar to \cref{example:self-correction}. That is, computing $\tilde{f}(u) + \tilde{f}(x^*-u)$ for $u \sim \uniform{\cX}$. However, this strategy fails. The reason is that the distribution  of $x^*-u$ depends on the (adversarially chosen) $x^*$, so we have no guarantee that $\tilde{f}(x^*-u)$ will be a ``good'' value with high probability (unlike for $\tilde{f}(u)$). In fact, as depicted in \cref{figure:why-not-blr}, $x^*-u$ might even be outside the domain~$\cX$.

\paragraph{Correlated sampling.}Instead, we use a strategy based on a \emph{correlated sampling lemma} (\cref{lemma:coupled-sampling}) to sample two points $x$ and $x'$, each of which has uniform marginal on $\cX$, but their joint distribution is such that they are on a straight line with $x^*$ (see \cref{figure:correlated-sampling}). That is, suppose we first draw a point $x \sim \uniform{\cX}$. We would like to draw another point $x'$ that satisfies the following two conditions:
\begin{enumerate}
    \item{
        $x' \in \ell$, where $\ell$ is the line segment that starts at $x^*$, ends at the boundary of $\cX$, and passes through $x$;
    }
    \item{
        The marginal distribution of $x'$ is also uniform on $\cX$.
    }
\end{enumerate}

Na\"{i}vely, one might imagine that we should sample $x'$ uniformly from the line $\ell$. However, due to intricacies related to the Borel--Kolmogorov paradox (about conditioning on an event of measure~$0$ in different ways), taking $x' \sim \uniform{\ell}$ \emph{would not work}. To see this, consider the case where $\cX = B(\mathbf{0}, 1)$ is the unit ball in $\bbR^n$ and $x^* = \mathbf{0}$ is the origin. Then, $\ell$ is a radius of the unit ball. 
Thus, taking $x' \sim \uniform{\ell}$ would yield a distribution where $\PP{\|x'\| \leq \nicefrac{1}{2}} = \nicefrac{1}{2}$. This is very different from the uniform distribution on $B(\mathbf{0}, 1)$, where $\PPP{x \sim \uniform{B(\mathbf{0}, 1)}}{\|x\| \leq \nicefrac{1}{2}} \leq e^{-\OmegaOf{n}}$. Therefore, sampling $x'$ uniformly from $\ell$ will not give an $x'$ with marginal distribution that is uniform on $\cX$. 

Where did the intuition that the distribution on $\ell$ should be uniform go wrong? In short, it is due to the \emph{curse of dimensionality}.
For a point $x'$ chosen from $\ell$ so that its marginal is $U(B(\mathbf{0}, 1))$, let $\rho_{\ell}$ be the distribution on $\norm{x'}_2$, i.e., the distance from $x'$ to the origin. We can directly compute the CDF of $\rho_{\ell}$. For arbitrary $r \in [0,1]$, 
\[ \PPP{r' \sim \rho_{\ell}}{r' \leq r} = \PPP{x' \sim U(B(\mathbf{0}, 1))}{\norm{x'}_2 \leq r} = \frac{\mathsf{vol}(B(\mathbf{0}, r))}{\mathsf{vol}(B(\mathbf{0}, 1))} = r^n, \]
where the last equality holds by a direct volume scaling argument. Taking the derivative of this CDF reveals the PDF $\rho_{\ell}(r) = n r^{n-1}$. Crucially, \emph{this is not the uniform distribution on $\ell$}, and in particular, is much more concentrated around $1$ than $0$. (Formally, $\rho_{\ell}$ is the $\mathrm{Beta}(n, 1)$ distribution.)

\cref{lemma:coupled-sampling} gives the correct sampling strategy in more detail, for more general choices of $\cX$ and $x^*$.

\begin{proof}[Proof idea for \cref{theorem:basic-linear-mitigation}]
    Use \cref{lemma:coupled-sampling} to sample $x$ and $x'$ with uniform marginals, forming a line with~$x^*$. Then, estimating the label of $x^*$ reduces to linear regression in $1$ dimension, where the error increases by a factor of $\ThetaOf{n}$, as in \cref{fig:linear-regression-1d}. To make the result robust to adversarial outliers in $\tilde{f}$, repeat this process a number of times and take the median.
\end{proof}

\subsubsection{Improved Local Mitigation for Linear Functions}

Looking at the basic linear mitigator of \cref{theorem:basic-linear-mitigation} (and at \cref{fig:linear-regression-1d}), there are two aspects that call for improvement. 
\begin{itemize}
    \item{
        The basic mitigator amplifies errors by a factor of $\ThetaOf{n}$. Namely, given a potentially-backdoored $\tilde{f}$ with accuracy $\BigO{\delta/n}$ on most points, the mitigator produces a ``clean'' prediction with accuracy $\delta$ on the adversarially chosen point $x^*$. Could the error growth be reduced?
    }
    \item{
        The basic mitigator guarantees (with probability $1-\negligible(s)$) that the label $y^*$ predicted for $x^*$ falls within the interval $\gD^\ideal\pm \delta$  (represented by the big red error bar in \cref{fig:linear-regression-1d}). This is a fairly strong security guarantee, but could it be improved? Specifically, it is not hard to see that, \emph{within} this error bar, the attacker can actually fully control where $y^*$ falls. Could we do better?  
    }
\end{itemize}

Our next theorem addresses both of these concerns.

\begin{theorem*}[Informal version of \cref{theorem:advanced-linear-mitigation}]
    In the setting of \cref{theorem:basic-linear-mitigation}, assume that the population distribution has benign noise. Namely, assume the labels of $\cD$ are generated by $y = h(x) + \eta$ where $h: \bbR^n \to \bbR$ is an affine function, and $\eta$ is independent subgaussian noise, as in \cref{definition:benign-noise}. Then there exists a local mitigator as in \cref{theorem:basic-linear-mitigation}, with the following improvements:
    \begin{itemize}
        \item{
            The error grows by a factor that is $\LittleO{n}$, specifically, from $\displaystyle ~ \frac{\delta}{n} ~$ to $\displaystyle ~ \frac{\delta}{n^{1/10}}$; and
        }
        \item{
            The predictions for every $x^*$ are unbiased. Namely, for every $x^* \in \cX: ~ \EE{y^*} = \gD^\ideal(x^*) = h(x^*)$. So no attacker can cause a systematic bias where the expected prediction for a select $x^*$ is above or below $\gD^\ideal(x^*)$.\footnote{
                However, an attacker could for example control the variance of $y^*$, within the constraints of the error bar of size $\delta/n^{1/10}$.
            }
        }
    \end{itemize}
\end{theorem*}

\begin{proof}[Proof idea for \cref{theorem:advanced-linear-mitigation}]
    First, we carefully analyze the errors that an attacker can introduce into the basic linear estimator, and conclude that the only way that an attacker can control $y^*$ is by introducing systematic errors in $\tilde{f}$ resulting in a linear correlation between the distance of a point $x$ from $x^*$ and the label $\tilde{f}(x)$. This leads to a constant additive bias in the predictions of the basic mitigator (as captured in \cref{eq:gi-decomposition}). By taking random samples $(x,y)$ from the population distribution $\cD$ and comparing $y$ to $\tilde{f}(x)$, we can estimate the constant additive bias that the attacker inserted. Subtracting this additive bias from our predictions yields a distribution of predictions that may have considerable variance, but is symmetric about $\gD^\ideal(x^*)$. Finally, we invoke our robust mean estimation theorem (\cref{theorem:robust-mean}) to obtain an unbiased estimator with small variance.  
\end{proof}

The proof of \cref{theorem:advanced-linear-mitigation} relies on our result concerning the mean-of-medians estimator (\cref{theorem:robust-mean}). For arbitrary distributions, that estimator does not necessarily concentrate about the mean.\footnote{%
    Simply because, for a skewed distribution, the median in each batch will not concentrate about the mean.   
} 
However, it does have good concentration if the distribution is  symmetric (as it is in our case). Furthermore, we show that it is robust to arbitrary adversarial noise.

\subsubsection{Local Mitigation for Polynomial Functions}
Our next construction is a local mitigator for population distributions with labels that are close to a polynomial function in $\bbR^n$ of total degree at most $d$. While learning a new polynomial regressor from scratch would require $n^{\OmegaOf{d}}$ samples, our mitigator uses $0$ samples, and a number of queries to $\tilde{f}$ that is \emph{independent} of $n$ and depends only linearly on $d$.

\begin{theorem*}[Informal version of \cref{theorem:basic-polynomial-mitigation}]
    Let $\cX \subseteq \bbR^n$ be a bounded and convex set, and let $\delta_0 = \delta_1 / O(nd^2)^d$, for some parameter $\delta_1 \geq 0$. For $\varepsilon < \nicefrac{1}{20d}$,
    \cref{algorithm:polynomial-mitigation-basic} defines a local mitigator $M$ that is $(\varepsilon, \delta_0 \to \delta_1)$-cutoff loss secure (\cref{definition:cutoff-loss-security}) for distributions $\cD$ with uniform marginal on $\cX$ such that $\LossCutoff{\delta_0}{\cD}{h} \leq \varepsilon$ for some polynomial $h: \bbR^n \to \bbR$ of total degree at most $d$.
    
    More explicitly, for every such distribution $\cD$, there exists a function $\gD^\ideal$ such that for any arbitrary (possibly malicious) function $f: ~ \cX \to \bbR$ with loss $\LossCutoff{\delta_0}{\cD}{h} \leq \varepsilon$, the mitigator $M$ satisfies:
    \begin{enumerate}
        \item{
            \textbf{Accuracy}. 
            $\PPP{(x,y) \sim \cD, \: y^* \gets M^{f}(x,1^n, 1^s)}{\left|y^* - y\right| > \delta_1} \leq \varepsilon + \negligible(s)$, and 
        }
        \item{
            \textbf{Cutoff Loss Security}. %
            $
                \forall x^* \in \cX: ~
                \PPP{y^* \gets M^{f}(x^*,1^n, 1^s)}{
                    \left|
                        y^*
                        - 
                        \gD^\ideal(x^*)
                    \right| > \delta_1
                } \leq \negligible(s)
            $.
        }
    \end{enumerate}
    Furthermore, \cref{algorithm:linear-mitigation-basic} uses a total of $\BigO{ds}$ oracle queries to $f$, does not use random samples from $\cD$, and runs in time $s \cdot \poly{n,d}$, assuming unit runtime cost for each arithmetic operation on the representations of real numbers involved in the computation.
\end{theorem*}

Just like for basic local mitigation for linear functions, the key idea is \emph{correlated sampling} (\cref{algorithm:resampling}). We sample $d+1$ points $x_0, \dots, x_d$, each of which has uniform marginal on $\cX$, but their joint distribution is such that they all lie on a line going through $x^*$. See \cref{figure:correlated-sampling-polynomial}. Since these points are all collinear, we then run a basic \emph{univariate} polynomial interpolation algorithm by solving the $(d+1)$-dimensional linear system (i.e., by inverting the corresponding Vandermonde matrix).

\begin{proof}[Proof idea for \cref{theorem:basic-polynomial-mitigation}]
    Use \cref{lemma:coupled-sampling} to sample $x_0, \dots, x_d$ with uniform marginals, forming a line with~$x^*$. Then, estimating the label of $x^*$ reduces to solving a linear system in $1$ dimension, where the error increases by a factor of $O(nd^2)^d$, by looking at the norm of the inverse of the corresponding Vandermonde matrix (see \cref{inverse-vandermonde-math,claim:poly-mitigation-vandermonde-points-are-far-apart}). To make the result robust to adversarial outliers in $\tilde{f}$, repeat this process a number of times and take the median.
\end{proof}

This result is a polynomial analog of our basic linear mitigation result (\cref{theorem:basic-linear-mitigation}).

\subsection{Example}

\begin{example}[Mitigation for functions of logarithmic degree]
    \label{example:functions-of-log-degree}
    Let $\cH_d$ be the class of Boolean functions of degree $d$. $\cH_d$ contains many interesting functions, including $d$-juntas, and decision trees of depth $d$ (Proposition 3.16 in \citealt{DBLP:books/daglib/0033652}). For $d = \log(n)$, \cref{theorem:fourier-heavy-mitigation} implies that mitigation is cheaper than learning.  
     
    More formally, let $n,d \in \bbN$, let $\cX = \pmo^n$, let $\cH_d = \left\{h \in \pmo^\cX: ~ \deg(h) \leq d\right\}$, where $\deg(h) = \max\big\{|S|: ~ S \subseteq [n] ~ \land ~ \widehat{h}(S) \neq 0\big\}$. 
    For each $h \in \cH_d$, let $\cD_h = \uniform{\{(x, h(x))\}_{x \in \cX}}$, and let $\bbD_d = \{\cD_h\}_{h \in \cH_d}$.
    
    It is known that $\cH_{\log(n)}$ can be agnostically learned in quasi-polynomial time using the ``low degree'' algorithm \citep{DBLP:journals/jacm/LinialMN93}. Namely, there exists an algorithm such that for every distribution $\cD \in \bbD_{\log(n)}$ and every $\varepsilon > 0$, the algorithm runs in time $\poly{n^{\log(n)}, 1/\varepsilon}$, uses only i.i.d.\ samples from $\cD$, and outputs a function $g$ with $\LossZeroOne{\cD}{g}\leq \varepsilon$. It is conjectured that no $\poly{n}$-time algorithm exists for learning $\log(n)$-juntas from i.i.d.\ samples (Section 2.3 in \citealp{DBLP:conf/crypto/BlumFKL93}. See also Assumption A.3 in \citealp{DBLP:journals/iacr/GolowichM24}).

    In contrast, mitigation is cheaper. Specifically, every $h \in \cH_{\log(n)}$ is $\nicefrac{2}{n}$-heavy (See Exercise 1.11(b) in \citealt{DBLP:books/daglib/0033652}). Therefore, by \cref{cor:binary-fourier-heavy-mitigation}, for $\varepsilon = \BigO{1/n^2}$ there exists a $\TV$ secure global mitigator $M$ that for any $\cD \in \bbD_{\log(n)}$ and any function $\tilde{f}$ with $\LossZeroOne{\cD}{\tilde{f}} \leq \varepsilon$, $M^{\tilde{f},\cD}(1^n, 1^s)$ runs in time $\poly{n,s}$ and outputs a function $g$ with loss $\LossZeroOne{\cD}{g} \leq \BigO{\varepsilon}$.\qed
\end{example}

\section{Preliminaries}
\label{section:preliminaries}

\begin{notation}
	$\bbN = \{1,2,3,\dots\}$, i.e., $0 \notin \bbN$. For any $n \in \bbN$, we denote $[n] = \{1,2,3,\dots,n\}$.
\end{notation}

\begin{notation}
	$\ln(\cdot)$ denotes the natural logarithm, and $\log(\cdot)$ denotes the logarithm to base $2$.
\end{notation}

\begin{notation}
    For a predicate $\varphi$, we use the notation $\indicator{\varphi} \in \{0,1\}$ to denote the indicator variable of whether $\varphi$ is true $(1)$ or false $(0)$.
\end{notation}

\begin{notation}
    We use the function $\sign : \bbR \to \{-1,1\}$ to denote the mapping $x \mapsto 2 \cdot \indicator{x \geq 0} - 1$. More generally, for a set $\cX$ and a function $f : \cX \to \bbR$, we define $\sign(f) : \cX \to \{-1, 1\}$ by the mapping $x \mapsto \sign(f(x))$.
\end{notation}

\begin{notation}
	For a set $\Omega$, we write $\distribution{\Omega}$ to denote the set of all probability measures defined on the measurable space $(\Omega,\cF)$, where $\cF$ is some fixed $\sigma$-algebra that is implicitly understood. 
\end{notation}

\begin{definition}
	Let $\cP,\cQ$ be probability measures defined on a measurable space $(\Omega,\cF)$. The \ul{total variation distance} between $\cP$ and $\cQ$ is $\TVf{\cP,\cQ} = \sup_{A \in \cF} |\cP(A)-\cQ(A)|$.
\end{definition}

\begin{definition}
    \label{definition:loss}
    Let $\cX$ and $\cY$ be sets. A \ul{loss function} is a function $L:\:\cY\times\cY\to\bbR_{\geq 0}$. 
    For any distribution $\cD \in \distribution{\cX \times \cY}$ and (possibly randomized) function $f:\:\cX \to \distribution{\cY}$, the \ul{population loss of $f$ with respect to $\cD$} is
    \[
        \Loss{\cD}{f} = \EEE{(x,y)\sim\cD}{L\big(f(x),y\big)},
    \]
    where the expectation is over the sample $(x,y)$ and the randomness of $f$.

    In particular, for discrete $\cY$,
    \begin{itemize}
        \item{
            The \ul{$0$-$1$ loss} is  $\LossZeroOneNameEmpty(y,y') = \indicator{y\neq y'}$, such that
            $
                \LossZeroOne{\cD}{f} = \PPP{(x,y)\sim\cD}{f(x)\neq y}
            $,
        }
    \end{itemize}
    and for $\cY = \bbR$, 
    \begin{itemize}
        \item{
            The \ul{square loss} is $\LossSquareNameEmpty(y,y') = (y-y')^2$, such that
            $
                \LossSquare{\cD}{f} = \EEE{(x,y)\sim\cD}{\left(
                    f(x)-y
                \right)^2}
            $, and
        }
        \item{
            The \ul{$\delta$-cutoff loss}\footnote{
                This follows existing terminology. For example, \cite{DBLP:conf/colt/AttiasHKKV24} call this loss the ``cut-off loss at scale $\delta$''.  
            } is $\LossCutoffNameEmpty{\delta}(y,y') = \indicator{|y - y'| > \delta}$, such that
            $
                \LossCutoff{\delta}{\cD}{f} = \PPP{(x,y)\sim\cD}{\Big|f(x)-y\Big| > \delta}
            $,
            where $\delta \geq 0$ is some fixed threshold.
        }
    \end{itemize}
\end{definition}

\begin{notation}
    For a collection of functions $\cF \subseteq \cY^\cX$ and a loss function $L:\:\cY\times\cY\to\bbR_{\geq 0}$, we use the notation $L_\cD(\cF) = \inf_{f \in \cF} L_\cD(f)$.
\end{notation}

\section{Definitions of Secure Backdoor Mitigation}
\label{section:security-definitions}

In this paper, we show that in some cases it is possible to mitigate the threat of undetectable backdoors in machine learning models. Specifically, we provide mitigation strategies that achieve a very strong notion of backdoor mitigation, as in the following definitions.

\begin{definition}[Mitigator]
    \label{definition:mitigator}
    Let $\{\cX_n\}_{n \in \bbN}$ be a sequence of sets, and let $\cY$ be a set.
    A \ul{mitigator} is a $\PPT$ algorithm $M$ with one of the following signatures:
    \begin{align*}
        \text{Global mitigator: } ~ g &\gets M^{f,\cD}(1^n, 1^s)
        \\
        \text{Local mitigator: } ~ y^* &\gets M^{f,\cD}(x^*, 1^n, 1^s).
    \end{align*}
    In both cases, $M$ has oracle access to a function $f:\:\cX_n \to \cY$, i.i.d.\ sample access to a distribution $\cD \in \distribution{\cX_n \times \cY}$, and takes an index $n$ and a security parameter $s \in \bbN$ as inputs.
    
    A \ul{global mitigator} outputs a (possibly randomized) function $g:\:\cX_n \to \distribution{\cY}$. A \ul{local mitigator} receives an additional input $x^* \in \cX_n$, and outputs a label $y^* \in \cY$. 
    
    For a global mitigator $M$, the notation $M^{f,\cD}(x^*, 1^n, 1^s)$ is shorthand for $g(x^*)$ where $g = M^{f,\cD}(1^n, 1^s)$.
\end{definition}

Throughout the paper, we occasionally neglect the subscript $n$, writing $\cX$ instead of $\cX_n$ when $n$ is clear from context.

\subsection{General Definition of Mitigation Security}
\begin{definition}[Distribution Dissimilarity]
    \label{definition:distribution-dissimilarity}
    Let $\Omega$ be a set. A \ul{distribution dissimilarity function for $\Omega$} is a function $\statDistName: ~ \distribution{\Omega} \times \distribution{\Omega} \to \NonNeg$. Namely, it is a function that takes a pair of distributions over $\Omega$ and outputs a non-negative number.
\end{definition}
Note that a distribution dissimilarity function may not be a metric. 

We introduce the following definition of security for mitigators. It can be applied both to local and global mitigators.

\begin{definition}[General $\varepsilon_0 \to \varepsilon_1$ Secure Backdoor Mitigation]
    \label{definition:mitigator-security-general}
    Let $\cX$ and $\cY$ be sets. For every index $n \in \bbN$, let $\cX_n \subseteq \cX$ be a set and let $\bbD_n \subseteq \distribution{\cX_n \times \cY}$ be a collection of distributions. Let $\bbD = \{\bbD_n\}_{n \in \bbN}$. Let $L^{(0)},L^{(1)}:\cY\times\cY\to\bbR_{\geq 0}$ be loss functions, let $\statDistName$ be a distribution dissimilarity function for the set of randomized functions $\cX \to \distribution{\cY}$, and let $\varepsilon_0,\varepsilon_1 \geq 0$. A local or global mitigator $M$ is an \ul{$(L^{(0)},\varepsilon_0) \to (L^{(1)},\varepsilon_1)$ mitigator for distance $\statDistName$ and population distributions $\bbD$} if:
    \begin{flalign*}
        &
        \qquad \exists \text{ a negligible function } \mu \in \negligible
        &
        \\
        &
        \qquad \quad 
        \forall \text{ index } n \in \bbN ~ \forall \text{ population distribution } \cD \in \bbD_n
        &
        \\
        & 
        \qquad \quad\quad
        \exists \text{ output distribution } \cG^\ideal_\cD \in \distribution{\cY^{\cX_n}} 
        &
        \\
        & 
        \qquad \quad\quad\quad
        \forall \text{ security parameter } s \in \bbN
        &
        \\
        & 
        \qquad \quad\quad\quad\quad
        \forall \text{ function } f \in  \cY^{\cX_n} \text{ with loss } L^{(0)}_\cD(f) \leq \varepsilon_0
        &
        \\[0.3em]
        &
        \qquad \quad\quad\quad\quad\quad
        \text{The following two conditions hold}:
        &
    \end{flalign*}
    \begin{enumerate}[leftmargin=9em]
        \item{
            \label{item:mitigator-security-independence}
            \textbf{Security}. $\statDist{\cG, \cG^\ideal_\cD} \leq \mu(s)$, and
        }
        \vsp
        \item{
            \label{item:mitigator-security-low-loss}
            \textbf{Accuracy}. 
            $\PPP{
                g \sim \cG
            }{
                L^{(1)}_\cD(g) \geq \varepsilon_1
            } \leq \mu(s)$,
        }
    \end{enumerate}
    \begin{adjustwidth}{7em}{0pt}
         where $\cG$ is the output distribution of the mitigator. Specifically, if $M$ is a global mitigator then $\cG$ is the distribution of $M^{f, \cD}(1^n, 1^s)$; if $M$ is a local mitigator then $\cG$ is the distribution of $g(x) = M^{f, \cD}(x, 1^n, 1^s)$.
    \end{adjustwidth}
\end{definition}

\begin{remark}
    \label{remark:local-global-G}
    In \cref{definition:mitigator-security-general}, $\cG$ represents the output of the mitigator. This is true for both local and global mitigators. Local and global mitigators differ in their use of randomness: a global mitigator uses its random coins and random samples to select one deterministic function $\cX \to \cY$, which maps each $x \in \cX$ to some label in $\cY$. In contrast, a local mitigator uses its random coins and random samples to map just a single input $x$ to a single label $y$. A technicality that results from this difference is that for a global mitigator, $\cG$ (the distribution of $M^{f, \cD}(1^n, 1^s)$) is a distribution over deterministic functions $\cX \to \cY$. In contrast, for a local mitigator, $g(x) = M^{f, \cD}(x, 1^n, 1^s)$ is a single fixed randomized mapping $\cX \to \distribution{\cY}$.\footnote{
        Generally, in both cases $\cG$ is a distribution over randomized functions (i.e., a distribution over functions, and each function may itself be a randomized mapping; $\cG \in \distribution{(\distribution{\cY})^\cX}$). In the case of global mitigators the randomized functions are degenerate ($\cG$ is a distribution over deterministic functions); in the case of local mitigators the distribution over functions is degenerate ($\cG$ assigns a probability of $1$ to a single fixed mapping $\cX \to \distribution{\cY}$ that maps each $x \in \cX$ to a distribution over labels).
    }\footnote{
        If we desired to have a closer formal match between the ``type'' of $\cG$ for local and global mitigation, we could easily define $\cG$ in both cases as a distribution over deterministic functions. To do that, we would define $\cG$ in the case of local mitigation as the distribution of the function $g$ such that $g(x) = M^{f}(x,Z,r,1^n,1^s)$, where $Z \sim \cD^k$ is an i.i.d.\ sample, $k \in \bbN$ is the maximum number of samples used by $M$, and $r$ is the vector of random coins used by $M$. For any fixed $(Z,r)$, $g$ is a deterministic function; $\cG$ is the distribution over functions $g$ corresponding to a random choice of $(Z,r)$. This formulation is not meaningfully different from the formulation we use in this paper, and all results would be the same. We opt for the formulation as in \cref{remark:local-global-G} simply because the notation is slightly cleaner without explicit parameters of $(Z,r)$.
    }
\end{remark}

\subsection{Specific Instantiations of Mitigation Security}

The quality of the security guaranteed by \cref{definition:mitigator-security-general} hinges crucially on the choice of the distribution dissimilarity function $\lambda$. 
We now provide two instantiations of \cref{definition:mitigator-security-general} that correspond to specific choices of $\lambda$. 

Perhaps the most natural choice of a dissimilarity function, which provides a very strong guarantee of security, is to use the total variation distance, as in the following definition. 

\begin{definition}[$\TV$ Security for Global Mitigation]
    \label{definition:tv-security}
    Let $\cX$ and $\cY$ be sets. For every index $n \in \bbN$, let $\cX_n \subseteq \cX$ be a set and let $\bbD_n \subseteq \distribution{\cX_n \times \cY}$ be a collection of distributions. Let $\bbD = \{\bbD_n\}_{n \in \bbN}$. Let $L^{(0)},L^{(1)}:\cY\times\cY\to\bbR_{\geq 0}$ be loss functions, and let $\varepsilon_0,\varepsilon_1 \geq 0$. A global mitigator $M$ is \ul{$(L^{(0)},\varepsilon_0) \to (L^{(1)},\varepsilon_1)$ total variation secure for distributions $\bbD$} if:
    \begin{flalign*}
        &
        \qquad \exists \text{ a negligible function } \mu \in \negligible
        &
        \\
        &
        \qquad \quad 
        \forall \text{ index } n \in \bbN ~ \forall \text{ population distribution } \cD \in \bbD_n
        &
        \\
        & 
        \qquad \quad\quad
        \exists \text{ output distribution } \cG^\ideal_\cD \in \distribution{\cY^{\cX_n}} 
        &
        \\
        & 
        \qquad \quad\quad\quad
        \forall \text{ security parameter } s \in \bbN
        &
        \\
        &
        \qquad \quad\quad\quad\quad
        \forall \text{ function } f \in  \cY^{\cX_n} \text{ with loss } L^{(0)}_\cD(f) \leq \varepsilon_0
        &
        \\[0.3em]
        &
        \qquad \quad\quad\quad\quad\quad
        \text{The following two conditions hold}:
        &
    \end{flalign*}
    \begin{enumerate}[leftmargin=9em]
        \item{
            \label{item:tv-security-tv-independence}
            \textbf{Security}. 
            $
                \TVf{\cG, \cG^\ideal_\cD} \leq \mu(s)
            $, and
        }
        \item{
            \label{item:tv-security-low-loss}
            \textbf{Accuracy}. 
            $\PPP{
                g \sim \cG
            }{
                L^{(1)}_\cD(g) \leq \varepsilon_1
            } \geq 1-\mu(s)$,
        }
    \end{enumerate}
    \begin{adjustwidth}{7em}{0pt}
        where $\cG$ is the distribution of the function $M^{f, \cD}(1^n, 1^s)$.
    \end{adjustwidth}
\end{definition}

\vspp

When the label space $\cY$ is a metric space, it makes sense to consider a more nuanced notion of security. Here, the guarantee is that the mitigator outputs a label $y^*$ that, while not indistinguishable from the ``ideal'' label, is promised to be ``close enough'' to the ideal label according to the metric on $\cY$. 
Following is a formalization of this notion of security for local mitigators with label space $\cY = \bbR$.

\begin{definition}[Cutoff Loss Security]
    \label{definition:cutoff-loss-security}
    Let $\cX$ be a set. For every index $n \in \bbN$, let $\cX_n \subseteq \cX$ be a set and let $\bbD_n \subseteq \distribution{\cX_n \times \bbR}$ be a collection of distributions. Let $\bbD = \{\bbD_n\}_{n \in \bbN}$, and let $\varepsilon, \delta_0, \delta_1 \geq 0$. A local mitigator $M$ is \ul{$(\varepsilon, \delta_0 \to \delta_1)$-cutoff loss secure for distributions $\bbD$} if:
    \begin{flalign*}
        &
        \qquad 
        \exists \text{ a negligible function } \mu \in \negligible
        &
        \\
        &
        \qquad \quad
        \forall \text{ index } n \in \bbN ~ \forall \text{ population distribution } \cD \in \bbD_n
        &
        \\
        & 
        \qquad \quad\quad
        \exists \text{ a function } \gD^\ideal: \cX_n \to \bbR
        &
        \\
        & 
        \qquad \quad\quad\quad
        \forall \text{ security parameter } s \in \bbN 
        &
        \\
        & 
        \qquad \quad\quad\quad\quad
        \forall \text{ function } f \in  \bbR^{\cX_n} \text{ with } \PPP{(x,y) \sim \cD}{\left| f(x) - y \right| \geq \delta_0} \leq \varepsilon 
        &
        \\[0.3em]
        &
        \qquad \quad\quad\quad\quad\quad
        \text{The following two conditions hold:}
        &
    \end{flalign*}
    \begin{enumerate}[leftmargin=9em]
        \item{
            \label{item:loss-security-low-loss}
            \textbf{Accuracy}.  $\displaystyle \PPPunder{\substack{(x,y) \sim \cD \\ y^* \gets M^{f,\cD}(x, 1^n, 1^s)}}{\left| y^* - y \right| \geq \delta_1} \leq \varepsilon + \mu(s)$.
        }
        \item{
            \label{item:loss-security-f-independence}
            \textbf{Cutoff Loss Security}. $\displaystyle \forall x^* \in \cX:
                \!\!\!{\PPPunder{y^* \gets M^{f,\cD}(x^*,1^n, 1^s)}{
                    \left| y^* - \gD^\ideal(x^*) \right| \geq \delta_1
                } \leq \mu(s)}$.
        }
    \end{enumerate}
    \hspace*{6.7em} Furthermore, we say that $M$ is \ul{unbiased} if $M$ satisfies the following guarantee:
    \begin{enumerate}[leftmargin=9em]
        \setcounter{enumi}{2}
        \item{
            \label{item:loss-security-unbiased}
            \textbf{Mean Security}. $M$ is an unbiased estimator of $\gD^\ideal$ such that
            \[
            \forall x^* \in \cX_n: \:
            \EEE{y^* \gets M^{f,\cD}(x^*,1^n, 1^s)}{
                y^*
            } = \gD^\ideal(x^*).
            \]
        }
    \end{enumerate}
\end{definition}

\subsection{The Cutoff Dissimilarity Function}
\label{section:cutoff-loss-discussion-general}

\cref{definition:cutoff-loss-security} is essentially obtained by instantiating \cref{definition:mitigator-security-general} with the choices $L^{(0)} = \LossCutoffNameEmpty{\delta_0}$, $L^{(1)} = \LossCutoffNameEmpty{\delta_1}$, and $\lambda = \CutDist_{\delta_1}$, as in the following definition.\footnote{
    The mean security requirement (\cref{item:loss-security-unbiased}) is an optional additional guarantee that does not follow from \cref{definition:mitigator-security-general}, and makes sense only when $\cY = \bbR$. 
}

\begin{definition}
    \label{definiton:cutoff-loss-dissimilarity}
    Let $\cX \subseteq \bbR^n$ be a set and let $\delta \geq 0$. The \ul{$\delta$-cutoff distribution dissimilarity function for randomized functions $\cX \to \distribution{\bbR}$} is defined as follows. For any randomized functions $f, f': ~ \cX \to \Delta(\bbR)$,
    \[ 
        \CutDist_{\delta}(f, f') = \sup_{x \in \cX} \: \PP{ \left| f(x) - f'(x)\right| > \delta}, 
    \]
    where the probability is over the randomness of $f(x)$ and $f'(x)$.
\end{definition}

However, \cref{definition:cutoff-loss-security} is slightly simplified compared to a strict application of \cref{definition:mitigator-security-general}, as follows.

\begin{itemize}
    \item{
        The distribution dissimilarity function $\lambda$ in \cref{definition:mitigator-security-general} is defined over \emph{distributions} of randomized functions. However, as explained in \cref{remark:local-global-G}, a local mitigator outputs a fixed randomized function (i.e., a degenerate, singleton distribution over randomized functions). So our definition simplifies accordingly, to a notion of dissimilarity defined for fixed randomized functions.
    }
    \item{
        If we were to apply \cref{definition:mitigator-security-general} with loss $L^{(1)} = \LossCutoffNameEmpty{\delta_1}$ directly, \cref{item:loss-security-low-loss} in \cref{definition:cutoff-loss-security} would instead state that
        \begin{equation}\label{eqn:ideal-loss-accuracy} 
            \PPPunder{g \sim \cG}{\PPPunder{(x,y) \sim \cD}{|g(x) - y| \geq \delta_1} \geq \varepsilon } \leq \mu(s),
        \end{equation}
        where $\cG$ is the distribution of the function $g(x) = M^{f, \cD}(x, 1^n, 1^s)$, so sampling $g \sim \cG$ means sampling both (i) the internal randomness needed by $M$, and (ii) the samples from $\cD$ needed by $M$. We view this definition as slightly cumbersome, so instead we adopt \cref{item:loss-security-low-loss}. Note that \cref{eqn:ideal-loss-accuracy} implies \cref{item:loss-security-low-loss} by a union bound.
    }
\end{itemize}

\subsubsection{Security Implications of the Cutoff Dissimilarity Function}
\label{section:cutoff-loss-discussion-security}

We now show that cutoff loss security implies a notion of security that we call \emph{weak local $\TV$ security}.

\begin{definition}
    \label{definiton:local-TV-dissimilarity}
    Let $\cX, \cY$ be sets. The \ul{local $\TV$ distribution dissimilarity function for randomized functions $\cX \to \distribution{\cY}$} is defined as follows. For any randomized functions $f, f': ~ \cX \to \Delta(\cY)$,
    \[ 
        \LocalTV(f, f') = \sup_{x \in \cX} \: \TV\left(f(x), f'(x)\right). 
    \]
\end{definition}

We claim that \emph{randomized rounding} allows us to convert $\CutDist$ bounds into weak $\LocalTV$ bounds, while hurting the accuracy only by a bounded amount. Specifically, we have the following lemma.
\begin{lemma}
    For every index $n \in \bbN$, let $\cX_n$ be a set, and let $\bbD_n \subseteq \distribution{\cX_n \times \bbR}$ be a collection of distributions. Let $\bbD = \{\bbD_n\}_{n \in \bbN}$, and let $\varepsilon, \delta_0, \delta_1 \geq 0$. Suppose $M$ is an efficient mitigator that is $(\varepsilon, \delta_0 \to \delta_1)$-cutoff loss secure for distributions $\bbD$. Then, for all $\beta > 0$, there is an efficient mitigator $M'$ with the following modified accuracy and loss security properties, with the same order of quantifiers as in \cref{definition:cutoff-loss-security}:
    \begin{enumerate}
        \item \textbf{Accuracy}.  $\displaystyle \PPPunder{\substack{(x,y) \sim \cD \\ y^* \gets M'^{f,\cD}(x, 1^n, 1^s)}}{\left| y^* - y \right| \geq \delta_1 (1 + \beta)} \leq \varepsilon + \mu(s)$.
        \item \textbf{Local $\TV$ Security}. $\LocalTV(g, \gD^{\ideal}) \leq \mu(s) + \frac{1}{\beta}$, where $g : \cX \to \distribution{\bbR}$ is given by $g(x) = M'^{f, \cD}(x, 1^n, 1^s)$.
    \end{enumerate}
\end{lemma}

\DeclarePairedDelimiter\floor{\lfloor}{\rfloor}
\begin{proof}
    Let $\alpha = \delta_1 \cdot \beta \in \bbR$
    be a rounding parameter, and let $\floor{\cdot }_{\alpha} : \bbR \to \alpha \bbZ $ be the function that rounds downwards to an integer multiple of $\alpha$, i.e., $\floor{x}_{\alpha} = \floor{\frac{x}{\alpha}} \cdot \alpha \in \alpha \bbZ$.
    \newcommand{\oldg}{\gD^{\mathsf{old}\text{-}\ideal}}
    The new mitigator $M'$ simply runs $M$ to get some value $y \in \bbR$, samples a random offset $b \sim \uniform{[0, \alpha)}$, and outputs $\floor{y + b}_\alpha$. Suppose $\oldg$ was the previous ideal function that had cutoff loss security 
    We define the new $\gD^{\ideal}$ as $\gD^{\ideal}(x) = \floor{\oldg(x) + b}_\alpha$ where $b \sim \uniform{[0, \alpha)}$.

    By direct properties of the floor function, it is clear that accuracy can only get worse by an additive factor of $\alpha = \delta_1 \cdot \beta$. To see that $M'$ satisfies local $\TV$ security, recall that by \cref{item:loss-security-f-independence} in \cref{definition:cutoff-loss-security}, we have
    \[\displaystyle \forall x^* \in \cX:
                \!\!\!{\PPPunder{y^* \gets M^{f,\cD}(x^*,1^n, 1^s)}{
                    \left| y^* - \oldg(x^*) \right| \geq \delta_1
                } \leq \mu(s)}.\]
    Assuming $|y^* - \oldg(x^*)| < \delta_1$, then we know
    \[ \TV\left(\floor{y^* + b}_{\alpha}, \floor{\oldg(x^*) + b}_{\alpha} \right) \leq \frac{\delta_1}{\alpha} = \frac{1}{\beta}, \]
    as by a coupling argument, the only way the two quantities can differ is when $b$ lands in an interval that has length at most $\delta_1$. Therefore, by a union bound,
        \[\displaystyle \forall x^* \in \cX:
                 \TV\left( M'^{f, \cD}(x^*, 1^n, 1^s), \gD^{\ideal}(x^*) \right) \leq \mu(s) + \frac{1}{\beta}.\]
    That is, $\LocalTV(g, \gD^{\ideal}) \leq \mu(s) + \frac{1}{\beta}$, where $g : \cX \to \distribution{\bbR}$ is given by $g(x) = M'^{f, \cD}(x, 1^n, 1^s)$.
\end{proof}

\section{Some Initial Observations on Secure Backdoor Mitigation}
\label{section:initial-observations}

\subsection{Secure Backdoor Mitigation Cannot be Distribution Free}

\cite{DBLP:journals/jmlr/Feldman09} showed a result on the power of membership queries in agnostic learning, which has significant implications for our work. In secure backdoor mitigation, the defender receives a potentially-backdoored function $\tilde{f}$ from an untrusted party, and queries this function at locations of its choosing in order to learn a ``clean'' (``nearly canonical'') function $g$ that has good accuracy. For this process to make sense, we insist that backdoor mitigation be more efficient than learning a ``clean'' function from scratch using random samples. 

But why would it be possible for backdoor mitigation to be more efficient? The basic reason is that when learning from scratch, the learner has access only to \emph{i.i.d.\ random samples.}, whereas in backdoor mitigation, the defender also has \emph{oracle (membership query) access} to the (suspect but still useful) function $\tilde{f}$. This raises the question of when, if at all, can learning using oracle access to a labeling function be more efficient than learning using i.i.d.\ samples? \cite{DBLP:journals/jmlr/Feldman09} showed that in general, for distribution-free learning, oracle access does not help, as in the following theorem.

\begin{theorem}[Theorem 6  in \citealp{DBLP:journals/jmlr/Feldman09}]
    \label{theorem:feldman}
    Let $m,q,t: [0,1]^2 \to \bbN$. Let $\cX$ be a set, and let $\cH \subseteq \pmo^\cX$ be a class of functions with VC dimension $d \in \bbN$. Assume $A$ is an algorithm such that for every $\varepsilon,\delta \in [0,1]$, $A$ agnostically PAC learns $\cH$ with parameters $(\varepsilon,\delta)$. $A$ uses $m(\varepsilon,\delta)$ i.i.d.\ samples from a distribution $\cD$ that generates labeled examples of the form $(x,f(x))$ for some labeling function $f: \cX \to \pmo$, as well as using $q(\varepsilon,\delta)$ oracle queries to $f$, and $A$ runs in time $t(\varepsilon,\delta)$. 
    
    Then there exists an algorithm $A'$ that uses only i.i.d.\ samples from $\cD$ (and does not require oracle access), such that $A'$ agnostically PAC learns $\cH$ with parameters $(\varepsilon,\delta)$ with respect to the same types of distributions $\cD$. $A'$ uses 
    \[
        m'(\varepsilon,\delta) = \BigO{\frac{d\log(d/\varepsilon) + \log(1/\delta)}{\varepsilon^2}}
    \]
    i.i.d.\ samples from $\cD$, and $A'$ runs in time 
    $
    t'(\varepsilon,\delta) = t(\varepsilon/2,\delta/2) + m'(\varepsilon,\delta)
    $.
\end{theorem}

\begin{proof}[Proof Sketch]
    Take a sample $S$ consisting of $m'(\varepsilon,\delta)$ i.i.d.\ samples from $\cD$. By uniform convergence, $S$ is an $\varepsilon/2$-sample for $\cD$, namely, $|\PPP{Z \sim \uniform{S}}{Z \in E} - \PPP{Z \sim \cD}{Z \in E}| \leq \varepsilon/2$ for every event $E$. Now simulate $A$ with parameters $(\varepsilon/2,\delta/2)$ to learn the distribution $\uniform{S}$ and obtain a classifier $h: \cX \to \pmo$ (we have full knowledge of $S$, so we can simulate oracle queries to the labeling function of~$\uniform{S}$). Then $\LossZeroOne{\cD}{g} \leq \LossZeroOne{\uniform{S}}{g} + \varepsilon/2 \leq \varepsilon$, where the first inequality holds because $S$ is an $\varepsilon/2$-sample and the second follows from the correctness of $A$.
\end{proof}

The conclusion is that, for general distributions, learning with oracle access cannot be faster than learning from random samples, up to constant factors. 
To see that this implies that backdoor mitigation cannot in general be faster than learning from random samples, simply consider the case where the potentially-backdoored $\tilde{f}$ is completely honest, namely, $\tilde{f}$ is precisely the labeling function of the population distribution~$\cD$. This amounts to learning $\cD$ with oracle access to the true labeling function, which, as we saw, is no easier in general than learning $\cD$ from random samples.\footnote{
    One could still hope, for instance, that if the defender has whitebox access to $\tilde{f}$ then they could mitigate faster than learning with random samples. In this paper we focus on mitigation with blackbox access. 
}

However, \cref{theorem:feldman} does not imply that oracle access is not useful when learning \emph{with respect to specific families of distributions}, e.g., learning with respect to distributions that have a uniform marginal on the domain (a case we focus on in this paper). The reason is that the reduction in the proof of \cref{theorem:feldman} executes $A$ on the distribution $\uniform{S}$ -- which is very far from uniform on the domain $\cX$.

In fact, the \emph{learning parity with noise} (LPN) assumption, which is a widely-believed assumption in modern cryptography, posits that in some cases (e.g., learning parity functions), learning from noisy random samples is computationally infeasible, but learning with oracle access to a labeling functions is easy (e.g., for parity functions efficient learning is provided by the Goldreich--Levin algorithm, see \cref{theorem:goldreich-levin}). 

This is one justification for our focus in this paper on backdoor mitigation with respect to specific ``nice'' classes of distributions. See further discussion in \cref{section:on-our-assumptions}.

\subsection{Lower Bound for General Mitigation}

Here, we show that there are collections of distributions $\bbD$ where mitigation is not more efficient than directly (re-)learning from random samples. For concreteness, we will consider \cref{definition:tv-security}. The intuition here is that the only advantage global mitigation has, as compared to directly learning, is \emph{membership queries} from (a possibly corrupted) hypothesis $f$. However, there are natural choices of population distributions %
for which membership queries and sample access are equally powerful. In particular, if we look at functions with \emph{no structure}, then seeing labeled examples, either by samples or membership queries, does not help prediction for new, unseen inputs.

\cite{christiano2024backdoordefenselearnabilityobfuscation} show something similar in the \emph{white box} setting using indistinguishability obfuscation and puncturable pseudorandom functions. However, since all of our techniques are in the black-box setting, we demonstrate the significantly simpler result, where the mitigator gets only input-output access to $f$.

Formally, in \cref{definition:tv-security}, consider $\cX_n = \{0,1\}^n$, $\cY = \{0,1\}$. For a function $h : \cX_n \to \cY$, let $\cD_h = \uniform{\{(x, h(x))\}_{x \in \cX_n}}$. Let $\bbD_n = \{\cD_h : h \in \cY^{\cX_n} \}$. Let $L^{(0)} = L^{(1)} = \LossZeroOneNameEmpty$.

\begin{lemma}[Informal]\label{mitigator-def-lower-bound}
    In the above setting, for any $\varepsilon_1 > 0$ and negligible function $\mu$, suppose $M$ is a mitigator such that
    \[ \PP{\LossZeroOne{\cD_h}{M^{h, \cD_h}(1^n, 1^s)} \leq \varepsilon_1} \geq 1 - \mu(s),\]
    for a randomly chosen $h : \cX_n \to \cY$. Suppose that $M$ uses $k \leq 2^n/4$ samples from $\cD_h$ and makes $q \leq 2^n/4$ queries to the oracle $h$. Then, there is a learner $M'$ using $O(k + q)$ samples in expectation such that
    \[ \PP{\LossZeroOne{\cD_h}{M'^{\cD_h}(1^n, 1^s)} \leq \varepsilon_1} \geq 1 - \mu(s).\]
\end{lemma}

We emphasize that we are not using the full power of the mitigator in \cref{mitigator-def-lower-bound}, making the lower bound only stronger. In particular, we are only using \cref{item:tv-security-low-loss} when plugging in the true hypothesis $h$ as the membership oracle, and we are not using \cref{item:tv-security-tv-independence} at all. 

\begin{proof}[Proof sketch of \cref{mitigator-def-lower-bound}]
Since $\cD_h$ has no structure and we are considering the $\LossZeroOneNameEmpty$ loss, an optimal choice of mitigator $M$ with $k$ samples and $q$ queries to $f$ would be as follows. Given $k$ samples, $M$ would query $q$ distinct points in $\{0,1\}^n$ not present in the samples. Then, $M$ can store a lookup table corresponding to all sampled and queried pairs. For these input values, it would output whatever is in its table, and for everything else, it would randomly guess an element of $\{0,1\}$. Let $T$ be the number of distinct elements in the lookup table. It is clear that $T \leq k + q$.

A learner $M'$ without membership queries can emulate this process except for the $q$ queries to $h$. Instead, it can use samples from $\cD_h$ until it has seen $T$ distinct examples. In this case, the performance with respect to $\LossZeroOneNameEmpty$ will be at least as good as $M$. It remains to see how many samples $M'$ needs to generate $T$ distinct examples.

Let $T'$ denote the number of samples $M'$ draws. Let $Z'$ denote the expected number of distinct elements seen in $T'$ samples. For $T' \leq 2^n$, by linearity of expectation, we have
\[ \EE{Z'} = 2^n \left( 1 - \left(1 - \frac{1}{2^n} \right)^{T'} \right) \geq 2^n \left( 1 - e^{-\frac{T'}{2^n}} \right) \geq 2^n \left(1 - \left(1 - \frac{T'}{2 \cdot 2^n} \right) \right) = \frac{T'}{2},\]
where we have used the inequality $e^{-x} < 1-\frac{x}{2}$ for all $x \in [0,1]$. Therefore, setting $T' = 2(k + q) \leq 2^n$, we have $T' \geq 2T$, and therefore, $\EE{Z'} \geq T$.
\end{proof}

    \else
        
    \fi

    \ifdraftcompile
        \section{Global Mitigation}
\label{section:global-mitigation}

In this section we show a mitigation result for potentially-backdoored functions $\pmo^n\to\pmo$, which relies on two main assumptions:
\begin{enumerate}
    \item{
        The potentially-backdoored function has low population loss.
    }
    \item{
        The population distribution is ``nice'' in the sense that the marginal over the domain is uniform, and the labels are close to a Fourier-heavy function. (However, the labels need not be deterministic. Namely, for an $x$ in the domain, it is possible that both the labels $1$ and $-1$ have positive probability).  
    }
\end{enumerate} 

\subsection{Fourier Analysis Preliminaries}
\label{section:fourier-analysis-preliminaries}

Following are some basic notions from Fourier analysis of Boolean functions, which are used in this paper. See \cite{DBLP:books/daglib/0033652} for a comprehensive introduction.

\paragraph*{The expectation inner product.} 
Given a measure space, let $\cL_2$ be the set of real-valued random variables over the measure space with finite second moment. Then $\cL_2$ is a vector space over the reals. The function $\iprod{\cdot,\cdot}: \: \cL_2 \times \cL_2 \to \bbR$ given by
\[
    \iprod{X,Y} = \EE{XY}
\]
is an inner product.
In particular, $\| X \| = \sqrt{\iprod{X,X}}$ is a norm.

\paragraph*{Fourier characters and coefficients on the Boolean hypercube.} Let $n \in \bbN$ and let $\cX = \bbR^n$. We identify any function $f: ~ \cX \to \bbR$ with the random variable $f(X)$, where $X$ is uniform over $\cX$. In particular, $\iprod{f,g} = \EEE{X \sim \uniform{\cX}}{f(X)g(X)}$ for any $f,g \in \bbR^{\cX}$.

For any $S \subseteq [n]$, the \emph{character} of $S$ is the function $\cX \to \pmo$ given by
\[
    \chi_S(x) = \prod_{i \in S}x_i,
\]
where $x = (x_1,\dots,x_n)$.

The set $\left\{\chi_S(x): ~ S \subseteq [n]\right\}$ is an orthonormal basis of the space of all functions $\cX \to \bbR$. In particular, every function $f: ~ \cX \to \bbR$ has a unique representation
\[
    f(x) = \sum_{S \subseteq [n]} \widehat{f}(S)\chi_S(x)    
\]
where $\widehat{f}(S) = \iprod{f,\chi_S}$ is the \emph{$S$-coefficient}, or \emph{weight of $S$} in $f$. We indicate sets of characters with specified weights using notation such as $\widehat{f}^{\: \geq \alpha} = \left\{S \subseteq [n]: \: \left|\widehat{f}(S)\right| \geq \alpha\right\}$ and $\widehat{f}^{\: = \beta} = \left\{S \subseteq [n]: \: \left|\widehat{f}(S)\right| = \beta\right\}$. 

\begin{theorem}[Parseval's identity]
    For any $n \in \bbN$ and $f: ~ \pmo^n \to \bbR$,
    \[
        \|f\|^2 = \iprod{f,f} = \EEE{x \sim \uniform{\pmo^n}}{f(x)^2} = \sum_{S \subseteq [n]}\widehat{f}(S)^2.    
    \]
\end{theorem}

\subsection{Mitigation for Fourier-Heavy Functions}

\begin{definition}
    \label{definition:fourier-heavy-sparse}
    Let $n,t \in \bbN$, $\cX = \pmo^n$, and $\tau \geq 0$. The class of \ul{Fourier $\tau$-heavy} functions is
    \[
        \cH_{\geq \tau} = \Big\{f \in \bbR^\cX: \: 
        \left(
            \forall S \subseteq [n]: \:
                \left|\widehat{f}(S)\right| \geq \tau ~ \lor ~ \widehat{f}(S) = 0
        \right)
        \Big\}.
    \]
    Similarly, the class of \ul{binary-valued Fourier $\tau$-heavy} functions is
        \[
        \cH_{\geq \tau}^{\pm1} = \Big\{f \in \{\pm 1\}^\cX: \: 
        \left(
            \forall S \subseteq [n]: \:
                \left|\widehat{f}(S)\right| \geq \tau ~ \lor ~ \widehat{f}(S) = 0
        \right)
        \Big\} \subseteq \cH_{\geq \tau}.
    \]
    The class of \ul{Fourier $t$-sparse} function is
    $
        \cH_{t} = \Big\{f \in \bbR^\cX: \: 
        \left|\widehat{f}^{> 0}\right| \leq t
        \Big\}
    $.
\end{definition}

\begin{figure}[!ht]
    \centering
    \includegraphicsorstandalone[width=0.7\linewidth]{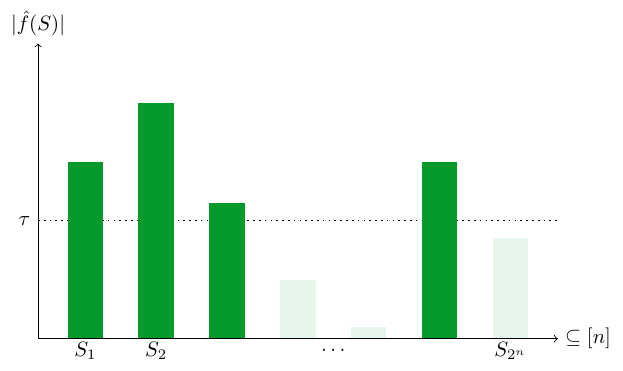}
    \caption{
        In a $\tau$-heavy function $f$, all non-zero Fourier coefficients have absolute value at least $\tau$.
    }
    \label{fig:enter-label}
\end{figure}

\begin{theorem}
    [$\TV$ global mitigation for Fourier-heavy functions]
    \label{theorem:fourier-heavy-mitigation}
    Let $n,s \in \bbN$, let $\tau > 0$, $\varepsilon_0 \leq (\tau/6)^2$ and $\varepsilon_1 > \varepsilon_0$, let $\cX = \pmo^n$. Let $\bbD \subseteq \distribution{\cX \times [-1,1]}$ be a collection of distributions $\cD$ with uniform marginal on $\cX$ such that $\LossSquare{\cD}{\cH_{\geq \tau}} \leq \varepsilon_0$. 
    Then \cref{algorithm:mitigator-fourier} is a global mitigator that is
    \[
        \left(
            \LossSquareNameEmpty, \varepsilon_0
        \right)
        \to
        \left(
            \LossSquareNameEmpty, \varepsilon_1
        \right)
    \]
    total variation secure for $\bbD$ (\cref{definition:tv-security}). 
    \cref{algorithm:mitigator-fourier} uses
    \[
        m = \BigO{\frac{
                s+\log(1/\tau)
            }{
                \tau^2(\varepsilon_1-\varepsilon_0)
            }
        }
    \]
    i.i.d.\ samples from a distribution $\cD \in \bbD$ and $\poly{n,1/\tau,s}$ oracle queries to an arbitrary function $f: \cX \to [-1,1]$ with loss $\LossSquare{\cD}{f} \leq \varepsilon_0$. 
\end{theorem}

\begin{remark}
    The assumption in \cref{theorem:fourier-heavy-mitigation} that $\varepsilon_0$ is strictly less than $\tau^2$ (e.g., $\varepsilon_0 \leq (\tau/6)^2$) appears to be necessary when using our Goldreich--Levin-based technique. Suppose that instead, we took $\varepsilon_0 = \tau^2$. Consider a function $y: \cX \to \bbR$ with two distinct coefficients $S_1,S_2 \subseteq [n]$ such that $\widehat{y}(S_i) = \tau$ for $i \in \{1,2\}$. Let $f_{- i}(x) = \sum_{S \subseteq [n]: \: S \neq S_i}\widehat{y}(S)\chi_S(x)$. Then for $\cD = \uniform{\{(x,y(x)): \: x \in \cX\}}$, the losses are $\LossSquare{\cD}{f_{-1}} = \LossSquare{\cD}{f_{-2}} = \tau^2 \leq \varepsilon_0$. Hence, the algorithm would be required to return the same output distribution on query access to either of $f_{-1}$ and $f_{-2}$. However, executing the Goldreich--Levin algorithm on $f_{-i}$ will not recover $S_i$. The missing coefficient (that is not recovered) depends on the function $f_{-i}$ being queried. This suggests that if increasing the upper bound beyond $\tau^2$ is possible, then doing so requires new ideas. \qed
\end{remark}

\begin{corollary}
    \label{cor:binary-fourier-heavy-mitigation}
    Let $n,s \in \bbN$, let $\tau > 0$, $\varepsilon_0 \leq (\tau/12)^2$ and $\varepsilon_1 > 4\varepsilon_0$, let $\cX = \pmo^n$. Let $\bbD \subseteq \distribution{\cX \times \{\pm 1\}}$ be a collection of distributions $\cD$ with uniform marginal on $\cX$ such that $\LossZeroOne{\cD}{\cH_{\geq \tau}^{\pm 1}} \leq \varepsilon_0$. Then, the composition of \cref{algorithm:mitigator-fourier} and $\sign(\cdot)$ is a global mitigator that is
    \[ \left( \LossZeroOneNameEmpty, \varepsilon_0 \right) \to \left( \LossZeroOneNameEmpty, \varepsilon_1 \right)  \]
    total variation secure for $\bbD$ (\cref{definition:tv-security}). The algorithm uses
        \[
        m = \BigO{\frac{
                s+\log(1/\tau)
            }{
                \tau^2(\varepsilon_1-4\varepsilon_0)
            }
        }
    \]
    i.i.d.\ samples from a distribution $\cD \in \bbD$ and $\poly{n,1/\tau,s}$ oracle queries to an arbitrary function $f: \cX \to \{\pm 1\}$ with loss $\LossZeroOne{\cD}{f} \leq \varepsilon_0$. 
\end{corollary}
\begin{proof}[Proof of \cref{cor:binary-fourier-heavy-mitigation}]
    This follows from \cref{lemma:ell-2-loss-vs-zero-one} and \cref{theorem:fourier-heavy-mitigation}, when applied with $4 \varepsilon_0$ instead of $\varepsilon_0$. Specifically,
    \begin{align*} 
    \LossSquare{\cD}{\cH_{\geq \tau}} &= 4 \LossZeroOne{\cD}{\cH_{\geq \tau}} \leq 4 \LossZeroOne{\cD}{\cH^{\pm 1}_{\geq \tau}} \leq 4 \varepsilon_0,
    \\\LossSquare{\cD}{f} &= 4\LossZeroOne{\cD}{f} \leq 4 \epsilon_0,
    \end{align*}
    where we have used the fact that $\cH_{\geq \tau}^{\pm 1} \subseteq \cH_{\geq \tau}$. Therefore, \cref{algorithm:mitigator-fourier} is a global mitigator that is $(\LossSquareNameEmpty, 4\varepsilon_0) \to (\LossSquareNameEmpty, \varepsilon_1)$ total variation secure for $\bbD$. Applying \cref{item:tv-security-low-loss} of \cref{definition:tv-security}, as guaranteed by \cref{theorem:fourier-heavy-mitigation}, we have
    \[\PPP{g \sim \cG}{
                \LossSquare{\cD}{g} \leq \varepsilon_1
            } \geq 1-\mu(s),\]
    where $\cG$ is the distribution of the function $M^{f, \cD}(1^n, 1^s)$. Since $\LossZeroOne{\cD}{\sign(g)} \leq \LossSquare{\cD}{g}$ by \cref{lemma:ell-2-loss-vs-zero-one}, it follows that
        \[\PPP{g \sim \cG}{
                \LossZeroOne{\cD}{\sign(g)} \leq \varepsilon_1
            } \geq 1-\mu(s),\]
    as desired.
\end{proof}

\begin{algorithmFloat}[H]
    \begin{ShadedBox}
        \textbf{Assumptions:}
        \begin{itemize}
            \item{
                $s,n \in \bbN$; $\tau > 0$; $\varepsilon_0 < \varepsilon_1$; $\varepsilon_0 \leq (\tau/6)^2$; $\cX = \{\pm 1\}^n$.
            }
            \item{
                $\cD \in \distribution{\cX \times [-1,1]}$ with $\cD_{\cX} = \uniform{\cX}$ and $\LossSquare{\cD}{\cH_{\geq \tau}} \leq \varepsilon_0$.
            }
            \item{
                The algorithm has oracle access to an arbitrary function $f \in \bbR^{\cX}$ (not necessarily in $\cH_{\geq \tau}$) such that $\LossSquare{\cD}{f} \leq \varepsilon_0$
            }
            \item{
                $m = \BigO{(s+\log(1/\tau))/\!\left(\tau^2(\varepsilon_1-\varepsilon_0)\right)}$.
            }
            \item{
                $Z = \bigl((x_1,y_1),\dots,(x_m,y_m)\bigr) \sim \cD^m$.
            }
        \end{itemize}

        \vspp

        \textsc{FourierHeavyMitigator}($Z$,$\tau$,$n$):
        \vsp
        \begin{algorithmic}
            \State Execute the Goldreich--Levin algorithm (\cref{theorem:goldreich-levin}) using oracle access to $f$ to\\ 
            \quad obtain a collection $\cS \subseteq 2^{[n]}$ such that $\PP{\widehat{f}^{\:\geq2\tau/3} \subseteq \cS \subseteq \widehat{f}^{\:\geq \tau/2}} \geq 1-\negligible(s)$
            \vsp

            \For $S \in \cS$:
                \vsp

                \State $\widebar{g}(S) \: \gets \: \frac{1}{m}\sum_{i \in [m]} \chi_S(x_i)\cdot y_i$

            \EndFor
            \vsp
            \State {\bfseries return} the function $g(x) = \sum\limits_{S \in \cS} \widebar{g}(S) \cdot \chi_S(x)$
        \end{algorithmic}
    \end{ShadedBox}
    \caption{An independent global mitigator for Fourier-heavy functions.}
    \label{algorithm:mitigator-fourier}
\end{algorithmFloat}

\subsubsection{Proof of Theorem~\ref{theorem:fourier-heavy-mitigation}}

We start with some basic observations.

\begin{notation}
    For $n \in \bbN$, $\cX = \pmo^n$ and $\cD \in \distribution{\cX \times \bbR}$, denote $\yD(x) = \EEE{(X,Y) \sim \cD}{Y \: | \: X = x}$.
\end{notation}

\begin{claim}[Loss decomposition]
    \label{claim:loss-decomposition}
    Let $\cX$ be a set, and let $\cD \in \distribution{\cX \times \bbR}$. Then for any function $r: ~ \cX \to \bbR$,
    \begin{align*}
        \EEE{(x,y) \sim \cD}{(r(x)-y)^2} 
        &= \EEE{x \sim \cD_\cX}{(r(x)-\yD(x))^2} + V(D),
    \end{align*}
    where $V(\cD) = \EEE{x \sim \cD_{\cX}}{\VVar{y \sim \cD_{Y|X=x}}{y}} = \EEE{x \sim \cD_\cX}{\EEE{y \sim \cD_{Y|X=x}}{y^2}-\EEE{y \sim \cD_{Y|X=x}}{y}^2}$.
\end{claim}

\begin{proof}
    \hfill$\begin{aligned}[t]
        \EEE{(x,y) \sim \cD}{(r(x)-y)^2}
        &=
        \EEE{(x,y) \sim \cD}{r(x)^2-2r(x)y + y^2}
        \\
        &= 
        \EEE{x \sim \cD_\cX}{r(x)^2-2r(x)\yD(x) + \EEE{y \sim \cD_{Y|X=x}}{y^2}}
        \\
        &= 
        \EEE{x \sim \cD_\cX}{r(x)^2-2r(x)\yD(x) + \yD(x)^2 + \EEE{y \sim \cD_{Y|X=x}}{y^2}-\yD(x)^2}
        \\
        &= 
        \EEE{x \sim \cD_\cX}{(r(x) - \yD(x))^2} + V(\cD).
        \hspace*{10em}
        \qed
    \end{aligned}$
    \let\qed\relax
\end{proof}

\begin{fact}
    \label{claim:fourier-of-fd}
    Let $n \in \bbN$, $\cX = \pmo^n$ and $\cD \in \distribution{\cX \times \bbR}$ with $\cD_\cX = \uniform{\cX}$. Then
    \[
        \forall S \subseteq [n]:
        ~ 
        \widehat{\yD}(S) = \EEE{(x,y)\sim\cD}{y\cdot\chi_S(x)}.
    \]
\end{fact}

\begin{proof}
    \hfill$\begin{aligned}[t]
        \widehat{\yD}(S)
        &=
        \iprod{
            \yD, \chi_S
        }
        =
        \EEE{
            x \sim \uniform{\cX}
        }{
            \EEE{y \sim \cD_{Y|X=x}}{y}
            \cdot
            \chi_S(x)
        }
        =
        \EEE{(x,y)\sim\cD}{
            y\cdot\chi_S(x)
        }.
    \end{aligned}$
\end{proof}

We are now prepared to prove \cref{theorem:fourier-heavy-mitigation}.

\begin{proof}[Proof of \cref{theorem:fourier-heavy-mitigation}]

    First, by the Goldreich--Levin theorem (\cref{theorem:goldreich-levin}), $\poly{n,1/\tau, s}$ queries to $f$ are indeed sufficient to guarantee that 
    \begin{equation}
        \label{eq:goldreich-levin-guarantee}
        \PP{\widehat{f}^{\:\geq2\tau/3} \subseteq \cS \subseteq \widehat{f}^{\:\geq \tau/2}} \geq 1-\negligible(s).
    \end{equation}
    By Parseval's identity and the inclusion $\cS \subseteq \widehat{f}^{\:\geq \tau/2}$ in \cref{eq:goldreich-levin-guarantee}, it follows that 
    \begin{equation}
        \label{eq:bound-on-size-of-S}
        \PP{|\cS| \leq 4/\tau^2} \geq 1-\negligible(s)
    \end{equation} 
    Second, by \cref{claim:fourier-of-fd} and Hoeffding's inequality, for every $S \in \cS$,
    \begin{equation}
        \label{eq:g-bar-s-good-estimate}
        \PPP{Z \sim \cD^m}{\left|\widebar{g}(S)-\widehat{\yD}(S)\right| \leq \Delta} \geq 1-\frac{e^{-s}}{4/\tau^2},
    \end{equation}
    where $\Delta = \sqrt{\frac{\tau^2(\varepsilon_1-\varepsilon_0)}{4}}$, and we used the fact that $m \geq \OmegaOf{(s + \ln(1/\tau)) / \Delta^2}$.
    From \cref{eq:bound-on-size-of-S,eq:g-bar-s-good-estimate} and the union bound,
    \begin{equation}
        \label{eq:coefficient-estimation-delta}
        \PP{\forall S \in \cS: ~ \left|\widebar{g}(S)-\widehat{\yD}(S)\right| \leq \Delta} \geq %
        1-\negligible(s),
    \end{equation}
    where the probability is over the choice of the sample $Z$ and the randomness of the Goldreich--Levin algorithm.

    Third, let $\epsH = \LossSquare{\cD}{\cH_{\geq \tau}} \leq \varepsilon_0$, and let $h \in \cH_{\geq \tau}$ such that $\LossSquare{\cD}{h} = \epsH$.\footnote{
        To see that such an $h$ exists, identify each function in $\cH_{\geq \tau}$ with its vector of Fourier coefficients in $\bbR^{2^n}$. Note that $\cH_{\geq \tau}$ is a compact set in $\bbR^{2^n}$. By \cref{claim:loss-decomposition} and Parseval's identity, the function $\LossSquareName{\cD}$ is continuous. Existence follows from the extreme value theorem.
    } 
    We show that 
    \begin{equation}
        \label{eq:h-hat-geq-tau-eq-s}
        \PP{\widehat{h}^{\geq \tau} = \cS} \geq 1-\negligible(s),
    \end{equation}
    where the probability is over the randomness of Goldreich--Levin. This follows from the high agreement between $f$ and $h$, as follows.
    \begin{align*}
        \label{eq:f-h-agreement}
        \sum_{S \subseteq [n]}\left(
            \widehat{f}(S)
            -
            \widehat{h}(S)
        \right)^2
        &=
        \EEE{(x,y)\sim\cD}{
            (f(x)-h(x))^2
        }
        \\
        &=
        \EEE{(x,y)\sim\cD}{
            (f(x)-y + y-h(x))^2
        }
        \\
        &=
        \big\|(f-y)+(y-h)\big\|^2
        \\
        &\leq \left(
            \big\|f-y\big\|
            +
            \big\|y-h\big\|
        \right)^2
        \\
        &= \left(
            \left(\LossSquare{\cD}{f}\right)^{1/2}
            +
            \left(\LossSquare{\cD}{h}\right)^{1/2}
        \right)^2
        \leq
        4\varepsilon_0.
    \end{align*}
    In particular, 
    \begin{equation*}
        \forall S \subseteq [n]: \: \left|
        \widehat{f}(S)
            -
        \widehat{h}(S)
        \right|
        \leq 
        2\sqrt{\varepsilon_0}
        \leq
        \tau/3.
    \end{equation*}
    Consequently, for any $S$, if $\widehat{h}(S)\geq \tau$ then $\widehat{f}(S)\geq 2\tau/3$, so
    \begin{equation}
        \label{eq:h-hat-tau-inf-hat-two-thirds}
        \widehat{h}^{\geq \tau} \subseteq 
        \widehat{f}^{\:\geq 2\tau/3}.
    \end{equation}
    On the other hand, for any $S$, if $\widehat{f}(S)\geq \tau/2$ then $\widehat{h}(S) > 0$. Because $h \in \cH_{\geq \tau}$, this implies $\widehat{h}(S) \geq \tau$, so
    \begin{equation}
        \label{eq:f-hat-half-in-h-hat-tau}
        \widehat{f}^{\:\geq \tau/2} \subseteq \widehat{h}^{\geq \tau}.
    \end{equation}
    Combining \cref{eq:h-hat-tau-inf-hat-two-thirds,eq:f-hat-half-in-h-hat-tau,eq:goldreich-levin-guarantee} yields 
    \[
        \PP{\widehat{h}^{\geq \tau} \subseteq \widehat{f}^{\:\geq2\tau/3} \subseteq \cS \subseteq \widehat{f}^{\:\geq \tau/2} \subseteq \widehat{h}^{\geq \tau}} \geq 1-\negligible(s),
    \]
    as desired.

    Fourth, we consider the losses of $h$ and $g$.
    \begin{align}
        \LossSquare{\cD}{h}
        &=
        \EEE{X \sim \cD_\cX}{(h(X)-\yD(X))^2} + V(\cD)
        \tagexplain{By \cref{claim:loss-decomposition}}
        \nonumber
        \\
        &=
        \sum_{S \subseteq [n]}\left(\widehat{h}(S)-\widehat{\yD}(S)\right)^2
        +
        V(\cD),
        \tagexplain{Parseval's identity}
        \label{eq:loss-of-h}
        \\[1em]
        \LossSquare{\cD}{g}
        &=
        \sum_{S \in \cS}\left(\widehat{g}(S)-\widehat{\yD}(S)\right)^2
        +
        \sum_{S \in 2^{[n]}\setminus\cS}\left(\widehat{g}(S)-\widehat{\yD}(S)\right)^2
        +
        V(\cD)
        \tagexplain{ditto}
        \nonumber
        \\
        &\leq
        |\cS|\cdot\Delta^2 
        +
        \sum_{S \in 2^{[n]}\setminus\cS}\left(\widehat{g}(S)-\widehat{\yD}(S)\right)^2
        +
        V(\cD)
        \tagexplain{%
            By \cref{eq:coefficient-estimation-delta}, w.p.\ $1-\negligible(s)$%
        }
        \nonumber
        \\
        &=
        (\varepsilon_1-\varepsilon_0)
        +
        \sum_{S \in 2^{[n]}\setminus\cS}\left(\widehat{g}(S)-\widehat{\yD}(S)\right)^2
        +
        V(\cD)
        \tagexplainmultiline{%
            By \cref{eq:bound-on-size-of-S}, w.p.\ $1-\negligible(s)$%
        }
        \nonumber
        \\
        &=
        (\varepsilon_1-\varepsilon_0)
        +
        \sum_{S \in 2^{[n]}\setminus\cS}\left(\widehat{h}(S)-\widehat{\yD}(S)\right)^2
        +
        V(\cD)
        \tagexplainmultiline{%
            $\widehat{g}(S) = \widehat{h}(S) = 0$ for $S \notin \cS$
            \\
            \vphantom{1}~~by \cref{eq:h-hat-geq-tau-eq-s} and $h \in \cH_{\geq\tau}$,
            \\
            \vphantom{1}~~w.p.\ $1-\negligible(s)$%
        }
        \nonumber
        \\
        &\leq
        (\varepsilon_1-\varepsilon_0)
        +
        \sum_{S \subseteq [n]}\left(\widehat{h}(S)-\widehat{\yD}(S)\right)^2
        +
        V(\cD)
        \nonumber
        \\
        &=
        \varepsilon_1-\varepsilon_0 + \epsH \leq \varepsilon_1.
        \tagexplain{By \cref{eq:loss-of-h} and choice of $h$}
        \nonumber
    \end{align}
    This shows that \cref{algorithm:mitigator-fourier} satisfies the accuracy requirement (\cref{item:tv-security-low-loss}) in \cref{definition:tv-security}. 
    
    Fifth, for the security requirement (\cref{item:tv-security-tv-independence} in \cref{definition:tv-security}), observe that \cref{algorithm:mitigator-fourier} outputs a function $g$ which is completely determined by (i.e., is a function of) the tuple $(Z, \cS)$. This gives the following Markov chain:
    \begin{equation}
        \label{eq:markov-chain}
        f - \cS - (Z, \cS) - g.\\
    \end{equation}
    However, \cref{eq:h-hat-geq-tau-eq-s} states that $\cS $ equals the specific value $\widehat{h}^{\geq \tau}$ with probability $1-\negligible(s)$, regardless of the specific choice of $f$, and therefore the dependence on $f$ is negligible.
    
    More formally, let $\cG$ and $\cG^\ideal_\cD$ denote the distribution of the random variable $g$ when \cref{algorithm:mitigator-fourier} is executed with oracle access to function $f$ and $h$, respectively. For any set $A$, \cref{eq:markov-chain} implies 
    \[
        \PPP{g \sim \cG}{g \in A \: \Big| \: \cS = \widehat{h}^{\geq \tau}}
        =
        \PPP{g \sim \cG^\ideal_\cD}{g \in A \: \Big| \: \cS = \widehat{h}^{\geq \tau}} = p_A.
    \]
    So
    \begin{flalign*}
        &
        \TVf{\cG,\cG^\ideal_\cD}
        =
        \sup_A
        \Big|
            \PPG{g \in A} - \PPGideal{g \in A}
        \Big|
        &
        \\
        &
        \quad
        =
        \sup_A
        \Bigg|
            \PPG{g \in A \: \Big| \: \cS = \widehat{h}^{\geq \tau}}\PPG{\cS = \widehat{h}^{\geq \tau}} 
            + 
            \PPG{g \in A \: \Big| \: \cS \neq \widehat{h}^{\geq \tau}}\PPG{\cS \neq \widehat{h}^{\geq \tau}}
            \\
            &\qquad\qquad 
            - \PPGideal{g \in A \: \Big| \: \cS = \widehat{h}^{\geq \tau}}\PPG{\cS = \widehat{h}^{\geq \tau}} 
            -
            \PPGideal{g \in A \: \Big| \: \cS \neq \widehat{h}^{\geq \tau}}\PPG{\cS \neq \widehat{h}^{\geq \tau}}
        \Bigg|
        &
        \\
        &
        \quad
        =
        \sup_A
        \Big|
            p_A\cdot(1-\negligible(s)) + \negligible(s)
             - 
            p_A\cdot(1-\negligible(s)) - \negligible(s)
        \Big|
        \leq 
        \negligible(s),
        &
    \end{flalign*}
    as desired.
\end{proof}

    \else
        
    \fi

    \ifdraftcompile
        \section{Local Mitigation}

In this section we show mitigation results for potentially-backdoored functions $\bbR^n \to \bbR$, which rely on two main assumptions:
\begin{enumerate}
    \item{
        The potentially-backdoored model has low $\ell_2$ population loss.
    }
    \item{
        The population distribution is ``nice'' in the sense that the marginal over the domain $\bbR^n$ is uniform on the $n$-dimensional unit ball, and the labels are close to a linear or polynomial function $\bbR^n \to \bbR$. (However, the labels need not be deterministic.)  
    }
\end{enumerate}

\subsection{Local Mitigation Preliminaries}
\label{section:local-mitigation-preliminaries}

Below, we describe the notions we need regarding convexity and probability measures in Euclidean space.

\paragraph*{Convex sets and degeneracy.} 
For $n \in \bbN$, a set $C \subseteq \bbR^n$ is \emph{convex} if for all $x, y \in C$, the line segment between $x$ and $y$ is fully contained within $C$, i.e., for all $\alpha \in [0,1]$, $\alpha x + (1-\alpha) y \in C$. Throughout, we will work with \emph{nondegenerate} convex sets $C$, in the sense that the standard Lebesgue measure of $C$ is non-zero. (In other words, $C$ is ``full-dimensional''.) Furthermore, we work with bounded convex sets $C$, in the sense that $C$ is contained in some sufficiently large Euclidean ball.

For a bounded convex set $C$, we use the fact that for all $x \neq y \in C$, there exists a unique point $p$ on the boundary of $C$ such that the ray starting at $x$ that goes through $y$ intersects $p$. More formally, letting
\[ \alpha^* = \sup \left\{\alpha \in \bbR_{\geq 0} : x + \alpha (y-x) \in C\right\}, \]
we have
\[ p = x + \alpha^* \cdot (y-x). \]
Note that $\alpha^*$ exists and is finite since $C$ is bounded and $x \neq y$. Moreover, we assume that algorithms can efficiently compute such a point $p$ given $x$ and $y$. 

\paragraph*{Probability, measures, and forms.} We work with the standard Lebesgue measure, unless explicitly stated otherwise. For a measurable (in fact, convex) set $C \subseteq \bbR^n$, the uniform distribution over $C$ is with respect to the Lebesgue measure. That is, the probability density function is given by the standard Lebesgue measure, normalized by the volume of $C$. We sometimes use the variable names $\eta(\cdot)$ and $\mu(\cdot)$ to refer to differential forms corresponding to probability measures in $\bbR^n$. Note that $\eta(\cdot)$ and $\mu(\cdot)$ include the differential terms (e.g., ``$dx$'').

\subsection{Basic Mitigation for Linear Functions}
\label{section:basic-local-linear-mitigation}

\begin{theorem}
    \label{theorem:basic-linear-mitigation}
    Let $\delta \geq 0$ and $\varepsilon \in [0,\nicefrac{1}{100}]$. 
    For every index $n\in \bbN$, let $B_n = B(\mathbf{0},1) \subseteq \bbR^n$ be the unit ball; 
    let $\cX_n \subseteq B_n$ be a (nondegenerate) convex set; 
    let $\cH_n$ be the set of affine functions $\bbR^n \to \bbR$;
    and let $\bbD_n = \bbD_{n,\varepsilon,\delta/20n}$ be the collection of distributions $\cD$ with uniform marginal $\cD_{\cX_n} = \uniform{\cX^n}$ such that $\LossCutoff{\delta/20n}{\cD}{\cH_n} \leq \varepsilon$.
    Let $\bbD = \{\bbD_n\}_{n \in \bbN}$. 
    Then \cref{algorithm:linear-mitigation-basic} defines a local mitigator $M$ that is 
    \[
        \left(
            \varepsilon, \: \delta/(20n) \to \delta
        \right)
    \]
    cutoff loss secure (\cref{definition:cutoff-loss-security}) for distributions $\bbD$, where $s \in \bbN$ is the security parameter.
    
    In particular, there exists $\mu \in \negligible$ such that for every index $n \in \bbN$ and every distribution $\cD \in \bbD_n$ there exists a function $\gD^\ideal: ~ \cX_n \to \bbR$ such that for any arbitrary (possibly malicious) function $f:~ \cX_n \to \bbR$ with loss $\LossCutoff{\nicefrac{\delta}{20n}}{\cD}{h} \leq \varepsilon$, and for every $s \in \bbN$, the mitigator $M$ satisfies:
    \begin{enumerate}
        \item{
            \textbf{Accuracy}. 
            $\PPP{(x,y) \sim \cD, \: y^* \gets M^{f,\cD}(x,1^n, 1^s)}{\left|y^* - y\right| > \delta} \leq \varepsilon +\mu(s)$, and 
        }
        \item{
            \textbf{Cutoff Loss Security}. %
            $
                \forall x^* \in \cX_n: ~
                \PPP{y^* \gets M^{f,\cD}(x^*,1^n, 1^s)}{
                    \left|
                        y^*
                        - 
                        \gD^\ideal(x^*)
                    \right| > \delta
                } \leq \mu(s)
            $.
        }
    \end{enumerate}
    Furthermore, \cref{algorithm:linear-mitigation-basic} uses a total of $\BigO{s}$ oracle queries to $f$, does not use random samples from $\cD$, and runs in time $\BigO{sn}$, assuming unit runtime cost for each arithmetic operation on the representations of real numbers involved in the computation. However, $M$ is not guaranteed to be unbiased.
\end{theorem}

In particular, the number of samples and queries required in \cref{theorem:basic-linear-mitigation} is independent of~$n$. The runtime is optimal in the sense that merely performing $m = \ThetaOf{s}$ queries to $f$ on points $x_1,\dots,x_m \in \cX_n \subseteq \bbR^n$ requires a runtime of $\OmegaOf{sn}$.

\begin{algorithmFloat}[!ht]
    \begin{ShadedBox}
        \textbf{Assumptions:}
        \begin{itemize}
            \item{
                $n \in \bbN$; $\delta \geq 0$; $\varepsilon \in [0,\nicefrac{1}{100}]$.
            }
            \item{
                $\cX \subseteq B(\mathbf{0},1) \subseteq \bbR^n$; $\cX$ is convex and nondegenerate.
            }
            \item{
                $\cD \in \distribution{\cX \times \bbR}$ with $\cD_{\cX} = \uniform{\cX}$.
            }
            \item{
                $\cH$ is the class of affine linear functions $\bbR^n \to \bbR$.
            }
            \item{
                $\exists h \in \cH: ~ \LossCutoff{\nicefrac{\delta}{20n}}{\cD}{h} \leq \varepsilon$.
            }
            \item{
                The algorithm has oracle access to an arbitrary function $f: ~ \cX \to \bbR$ (not necessarily in $\cH$) such that $\LossCutoff{\nicefrac{\delta}{20n}}{\cD}{f} \leq \varepsilon$.
            }
            \item{
                $s \in \bbN$ is a security parameter; $m = 320s$.
            }
            \item{
                $x^* \in \cX$ is arbitrary.
            }
        \end{itemize}

        \vspp

        \BasicLinearMitigator($n$, $m$, $x^*$):
        \vsp
        \begin{algorithmic}

            \State $\cI \gets $ empty set
            \vspp

            \For $i \in [m]$:
                \vsp
                \State sample $x_i \sim \uniform{\cX}$
                \vsp
                
                \State $x_i' \gets \Resamp(x^*$, $x_i$, $n$) 
                \Comment{See \cref{algorithm:resampling}}
                \vsp

                \State $\displaystyle \lambda_i \gets \frac{\|x_i-x^*\|_2}{\|x_i-x^*\|_2-\,\|x_i'-x^*\|_2}$
                \vspp

                \If $|\lambda_i| \leq 4n$:
                    \vsp

                    \State $y_i \gets f(x_i)$; ~ $\tilde{y}_i' \gets f(x_i')$
                    \vsp

                    \State $g_i \gets (1-\lambda_i) y_i + \lambda_i \tilde{y}_i'$
                    \vsp
                
                    \State $\cI \gets \cI \cup \{i\}$
                \EndIf
            \EndFor
            \vspp

            \State $g \gets $ median of $\{g_i\}_{i \in \cI}$
            \vsp
            \State \textbf{output} $g$
        \end{algorithmic}
    \end{ShadedBox}
    \caption{A basic local independent mitigator for linear functions $\bbR^n \to \bbR$.}
    \label{algorithm:linear-mitigation-basic}
\end{algorithmFloat}

\begin{algorithmFloat}[!ht]
    \begin{ShadedBox}
        \textbf{Assumptions:}
        \begin{itemize}
            \item{
                $n \in \bbN$; $\cX \subseteq B(\mathbf{0},1) \subseteq \bbR^n$; $\cX$ convex.
            }
            \item{
                $x^* \in \cX$ is arbitrary.
            }
            \item{
                $x$ was sampled uniformly from $\cX$.
            }
        \end{itemize}

        \vspp

        $\Resamp$($x^*$, $x$, $n$):
        \vsp
        \begin{algorithmic}
            \State Let $\ell$ be the line segment from $x^*$ to a point on the boundary of $\cX$ such that\\ 
            \quad\quad $\ell$ passes through $x$, and let $t \in (0,2]$ be the length of $\ell$.
            \vsp

            \State Sample $r' \in [0,t]$ according to density function $q(r) = \frac{n}{t^n}r^{n-1}$.
            \vsp
            
            \State Let $x'$ be the point on $\ell$ at distance $r'$ from $x^*$.
            \vsp

            \State \textbf{output} $x'$
        \end{algorithmic}
    \end{ShadedBox}
    \caption{An algorithm for sampling two points with uniform marginals on the unit sphere, such that the points are on a straight line with an arbitrary input $x^*$.}
    \label{algorithm:resampling}
\end{algorithmFloat}

\subsubsection{Correlated Sampling Lemma}

\begin{lemma}[Correlated Sampling]
    \label{lemma:coupled-sampling}
    Let $n \in \bbN$, let $\cX \subseteq \bbR^n$ be any bounded (nondegenerate) convex set, and let $U = \uniform{\cX}$ be the uniform distribution on $\cX$. Let $x^* \in \cX$. Consider the following procedure for generating a point $x' \in \cX$:
    \begin{enumerate}
        \item{
            Sample $x \sim U$.
        }
        \item{
            Let $x' \gets$ \emph{$\Resamp$}($x^*$, $x$, $n$) using \cref{algorithm:resampling}.
        }
    \end{enumerate}
    Then $x^*$, $x$ and $x'$ are on a straight line, and furthermore, the random variables $x$ and $x'$ are equal in distribution.
\end{lemma}

\begin{proof}
    It is clear from construction of $\Resamp$ that $x^*, x$, and $x'$ all lie on the same straight line, so it suffices to argue that the distributions of $x$ and $x'$ are identically distributed.
    
    Consider a spherical coordinate system centered at $x^*$ (instead of the origin), where the $\ell_2$ distance from $x^*$ is denoted by $r$. (Alternatively, this can be described by the usual centered spherical coordinate system, where we translate the whole space by $-x^*$.) Let $\cA$ denote the set $\cX$ written in these spherical coordinates. Since $U$ is the uniform distribution over $\cX$, we can write the density of $U$ as a volume element in spherical coordinates as
    \begin{equation}\label{eq:main-polar-decomposition} dV(\btheta, r) = \1[(\btheta, r) \in \cA] \cdot \eta(\btheta) \cdot r^{n-1} dr ,
    \end{equation}
    where $dV$ is proportional to the standard volume form and $\eta = \eta(\btheta)$ is proportional to the standard volume form on the $(n-1)$-dimensional sphere, where $\btheta$ parameterizes points on the sphere and denotes directions relative to its center (in our case, $x^*$). More concretely,
    \[ \eta(\btheta) \propto d\theta_1 d\theta_2 \cdots d\theta_{n-1} \cdot  \prod_{i=1}^{n-1} \sin^{n-i-1}(\theta_i), \]
    where crucially, $\eta$ does not depend on $r$.

    We now invoke Lemma~\ref{lemma:joint-probability-splitting}, using the fact that the density $dV$ splits into products of $\btheta$ and $r$ measures, as in~\eqref{eq:main-polar-decomposition}. Let $x$ be decomposed into  $\btheta(x)$ and $r(x)$, and similarly for $x'$.

    Using the notation of Lemma~\ref{lemma:joint-probability-splitting}, 
    let
    \[ \mu_{\Theta}(\btheta) = \eta(\btheta),\;\; \mu_{R}(r) = r^{n-1} dr,\]
    and
    \[ \cA_{|\btheta} = \{r \in \bbR_{\geq 0} : (r, \btheta) \in \cA\}. \]
    We have
    \[ \widehat{\mu}_{R|\btheta}(r) = \frac{\1[r \in \cA_{|\btheta}]}{\int_{r' \in \bbR} \1[r' \in \cA_{|\btheta}] \cdot \mu_R(r')} \cdot \mu_R(r). \]
    Using convexity and the notation of $\Resamp$, we know that for $\btheta = \btheta(x)$, we have $\cA_{|\btheta} = [0, t]$, where $t$ is the same as in $\Resamp$, namely the length of the line segment $\ell$ with endpoints $x^*$ and the boundary of $\cX$, passing through $x$. Plugging in our measures, we get
    \begin{align*} \widehat{\mu}_{R|\btheta}(r) &= \frac{\1[r \in [0, t]]}{\int_{r' \in \bbR} \1[r' \in [0,t]] \cdot (r')^{n-1} dr'} \cdot r^{n-1} dr
    \\&= \1[r \in [0, t]] \cdot \frac{r^{n-1} dr}{\int_0^t (r')^{n-1} dr'}
    \\&= \begin{cases} \frac{n}{t^n} r^{n-1} dr &\text{if }r \in [0,t],\\
    0 & \text{if }r \notin [0,t].
    \end{cases}
    \end{align*}
    This exactly matches the density $q(\cdot)$ in $\Resamp$. Therefore, by Lemma~\ref{lemma:joint-probability-splitting}, the joint distributions $(\btheta(x), r(x))$ and $(\btheta(x), r(x'))$ are identical. Since $\btheta(x) = \btheta(x')$, it follows that the joint distributions $(\btheta(x), r(x))$ and $(\btheta(x'), r(x'))$ are identical, which implies that $x$ and $x'$ are identically distributed, as desired.
\end{proof}

\begin{lemma}\label{lemma:joint-probability-splitting}
    Let $\cX$ and $\cY$ be nice\footnote{Formally, we require $\cX$ and $\cY$ to be standard Borel spaces equipped with $\sigma$-finite measures. Since we will only use this when $\cX$ is the sphere and bounded $\cY \subseteq [0, \infty)$, both equipped with the standard Lebesgue measure, we do not elaborate on the most general requirements on $\cX$ and $\cY$.} bounded subsets of $\bbR^{n}$ and $\bbR$, respectively. Suppose $\mu$ is a joint probability measure on $\cX \times \cY$ that can be written as
    \[ \mu(x,y) = \1[(x,y) \in \cA] \cdot \mu_{\cX}(x) \cdot \mu_{\cY}(y) , \]
    for some measures $\mu_{\cX}$ on $\cX$, $\mu_{\cY}$ on $\cY$, and some measurable set $\cA \subseteq \cX \times \cY$. Consider the following sampling procedure:
    \begin{enumerate}
        \item Sample $(x, y_0) \sim \mu$. (We will not use $y_0$.)
        \item Let $\cA_{|x} := \{y \in \cY : (x,y) \in \cA\}$. For the above $x \in \cX$, consider the probability measure on $\cY$ given by
        \[ \widehat{\mu}_{\cY|x}(y) := \frac{\1[y \in \cA_{|x}]}{\int_{y' \in \cY} \1[y' \in \cA_{|x}] \cdot \mu_{\cY}(y')} \cdot \mu_{\cY}(y).\]
        Sample $y_1 \sim \widehat{\mu}_{\cY|x}$ and output $(x, y_1)$.
    \end{enumerate}
    Then, the distribution of $(x, y_1)$ is exactly $\mu$.
\end{lemma}

\begin{proof}
Consider the probability measures
    \begin{align*} 
        \widehat{\mu}_{\cX}(x) &:= \left( \int_{y \in \cY} \1[y \in \cA_{|x}] \cdot \mu_{\cY}(y) \right) \cdot \mu_{\cX}(x),
        \\
        \widehat{\mu}_{\cY|x}(y) &:= \frac{\1[y \in \cA_{|x}]}{\int_{y' \in \cY} \1[y' \in \cA_{|x}] \cdot \mu_{\cY}(y')} \cdot \mu_{\cY}(y), \;\;\;\text{for fixed $x \in \cX$}.
    \end{align*}
    Notice that for all $(x,y) \in \cX \times \cY$, we have the equality
    \[ \widehat{\mu}_{\cX}(x) \cdot \widehat{\mu}_{\cY|x}(y) = \1[(x,y) \in \cA] \cdot \mu_{\cX}(x) \cdot \mu_{\cY}(y) = \mu(x,y). \]
    Note that by the assumption that $\mu$ is a probability measure (i.e., integrates to $1$), one can immediately observe via Fubini's theorem that $\widehat{\mu}_{\cX}$ and $\widehat{\mu}_{\cY|x}$ are probability measures for any $x \in {\cX}$ (i.e., both integrate to $1$).

    Consider some measurable subset $S \subseteq \cA$. The probability of event $S$ under probability measure $\mu$ is given by
    \begin{align}\int_{(x,y) \in S} \mu(x,y) &= \int_{(x,y) \in {\cX} \times \cY} \1[(x,y) \in S] \cdot \mu(x,y)
    \nonumber\\\label{eqn:mu-XY} &= \int_{(x,y) \in {\cX} \times \cY} \1[(x,y) \in S] \cdot \widehat{\mu}_{\cX}(x) \cdot \widehat{\mu}_{\cY|x}(y). 
    \end{align}
    On the other hand, consider the sampling process defined in the statement. The marginal density measure on $\cX$ is given at a point $x \in \cX$ by
    \begin{align*} \int_{y \in \cY} \mu(x,y) &= \int_{y \in \cY} \1[(x,y) \in \cA] \cdot \mu_{\cX}(x) \cdot \mu_{\cY}(y)
    \\&= \left( \int_{y \in \cY} \1[y \in \cA_{|x}]  \cdot \mu_{\cY}(y) \right) \cdot \mu_{\cX}(x)
    \\&= \widehat{\mu}_{\cX}(x).
    \end{align*}
    By construction of our sampling process, we know that for fixed $x \in \cX$, we sample $y_1 \sim \widehat{\mu}_{\cY|x}$. Therefore, the probability of event $S$ under our sampling process is given by
    \[ \int_{(x, y) \in S} \widehat{\mu}_{\cX}(x) \cdot \widehat{\mu}_{\cY|x}(y) = \int_{(x,y) \in \cX \times \cY} \1[(x,y) \in S] \cdot \widehat{\mu}_{\cX}(x) \cdot \widehat{\mu}_{\cY|x}(y),\]
    which exactly matches our expression in \cref{eqn:mu-XY}, as desired.
\end{proof}

\begin{claim}
    \label{claim:bound-on-lambda}
    In the notation of \cref{lemma:coupled-sampling}, let
    \[
        r = \|x-x^*\|_2,
        \quad
        r' = \|x'-x^*\|_2,
    \]
    and 
    \[
        \lambda = \frac{r}{r-r'},
        \quad
        \lambda' = \frac{r'}{r'-r}.
    \]
    Then
    \[
        \PP{\max\left\{|\lambda|,|\lambda'|\right\} \leq 4n} \geq \frac{1}{8}.
    \]
\end{claim}

\begin{proof}
    By \cref{lemma:coupled-sampling} and the details of \cref{algorithm:resampling}, $r$ and $r'$ are each sampled independently from the interval $(0,t]$ according to the density function $q(r) = \frac{n}{t^n}r^{n-1}$, where $t$ is the length of the line $\ell$ that contains $x^*$, $x$ and $x'$. In particular, for any $p \in [0,1]$, take $\rho = \rho(p) \in [0,t]$ such that the following holds.
    \begin{align*}
        p = \int_{0}^{\rho}\!q(r)\,dr = \frac{n}{t^n}\left[\frac{r^n}{n}\right]_0^{\rho} = \left(\frac{\rho}{t}\right)^n 
        ~
        \Longrightarrow
        ~
        \rho(p) = p^{1/n}\cdot t.
    \end{align*}
    In particular, $\rho(\nicefrac{1}{4}) = (\nicefrac{1}{4})^{1/n}\cdot t$ and $\rho(\nicefrac{3}{4}) = (\nicefrac{3}{4})^{1/n}\cdot t$. This implies that 
    \begin{equation*}
        \PP{
            \Big|
                r-r'
            \Big| 
            \geq
            t\left(
                \left(
                    \frac{3}{4}
                \right)^{1/n}
                \!-
                \left(
                    \frac{1}{4}
                \right)^{1/n}
            \right)
        }
        \geq
        2\left(
            \frac{1}{4}
        \right)^{\!2}
        =
        \frac{1}{8}.
    \end{equation*}
    Hence, with probability at least $1/8$,
    \begin{align*}
        \max \left\{\big|\lambda\big|,\big|\lambda'\big|\right\}
        &= 
        \frac{
            \max \left\{|r|,|r'|\right\}
        }{
            |r-r'|
        } 
        \leq 
        \frac{t}{t\left(
            \left(
                \frac{3}{4}
            \right)^{1/n}
            \!-
            \left(
                \frac{1}{4}
            \right)^{1/n}
        \right)}
        =
        \frac{4^{1/n}}{3^{1/n}-1}
        =
        \frac{4^{1/n}}
        {
            e^{\ln(3)/n}-1
        }
        \\
        &=
        \frac{4^{1/n}}
        {
            \left(
                1 + \frac{\ln(3)}{n} + \sum_{k=2}^\infty\frac{\left(\ln(3)/n\right)^k}{k!}
            \right)-1
        }
        \leq
        \frac{4^{1/n}}{\left(\frac{\ln(3)}{n}\right)} \leq 4n,
    \end{align*}
    as desired.
\end{proof}

\subsubsection{Proof of Theorem~\ref{theorem:basic-linear-mitigation}}

The following claim says that in \cref{algorithm:linear-mitigation-basic}, for most $i \in \cI$, the functions $h$ and $f$, and the labels from the distribution $\cD$, are all close to each other for the points $x_i$ and~$x_i'$.

\begin{claim}
    \label{claim:delta-delta-bound-given-e}
    Let $\varepsilon,\delta \geq 0$. 
    Let $n$, $\cX$, $x$, $x'$ be as in \cref{lemma:coupled-sampling}. Let $\cD \in \distribution{\cX \times \bbR}$ be a distribution with $\cD_{\cX} = \uniform{\cX}$, and let $h,f: \: \cX \to \bbR$ such that $\LossCutoff{\delta}{\cD}{h} \leq \varepsilon$ and $\LossCutoff{\delta}{\cD}{f} \leq \varepsilon$. Let $y,y' \in \bbR$ be random variables such that $(x,y),(x',y') \sim \cD$. Let $\lambda$, $\lambda'$ be as in \cref{claim:bound-on-lambda} and let 
    \begin{align*}
        \Delta_y &= h(x)-y; \qquad~ \Delta_y' = h(x')-y'
        \\
        \Delta_f &= h(x)-f(x); \quad \Delta_f' = h(x')-f(x').
    \end{align*} 
    Let $E$ be the event $\big\{\big|\lambda\big| \leq 4n\big\}$ or $\big\{\max\left\{|\lambda|,|\lambda'|\right\} \leq 4n\big\}$.
    Then 
    \[
        \PP{\max\left\{
            |\Delta_y|,|\Delta_y'|
        \right\}
        > \delta
        \:
        \lor 
        \:
        \max\left\{
            |\Delta_f|,|\Delta_f'|
        \right\}
        > 2\delta
        ~
        \Big|
        ~
        E\: }
        \leq
        32\varepsilon.
    \]
\end{claim}

\begin{proof}
    Seeing as $\LossCutoff{\delta}{\cD}{h} \leq \varepsilon$ and $\LossCutoff{\delta}{\cD}{f} \leq \varepsilon$, a union bound yields
    \[
        \PP{
            |h(x)-y|\leq \delta
            ~ 
            \land 
            ~
            |h(x')-y'|\leq \delta
            ~ 
            \land 
            ~
            |f(x)-y|\leq \delta
            ~ 
            \land 
            ~
            |f(x')-y'|\leq \delta
        } \geq 1-4\varepsilon.
    \]
    By the triangle inequality,
    \begin{equation}
        \label{eq:delta-delta-bound}
        \PP{
            \max\left\{
                |\Delta_y|,|\Delta_y'|
            \right\}
            > \delta
            ~ 
            \land 
            ~
            \max\left\{
                |\Delta_f|,|\Delta_f'|
            \right\}
            > 2\delta
        } \geq 1-4\varepsilon.
    \end{equation}
    By \cref{claim:bound-on-lambda},
    \begin{equation}
        \label{eq:bound-on-e}
        \PP{E} \geq \frac{1}{8}.
    \end{equation}
    By \cref{eq:delta-delta-bound,eq:bound-on-e},
    \begin{flalign*}
        &
        \PP{
            \max\left\{
                |\Delta_y|,|\Delta_y'|
            \right\}
            > \delta
            ~ 
            \lor
            ~
            \max\left\{
                |\Delta_f|,|\Delta_f'|
            \right\}
            > 2\delta
            ~
            \Big|
            ~
            E\: 
        }
        &
        \\[0.5em]
        &
        \qquad\qquad
        \leq
        \frac{
            \PP{
                \max\left\{
                    |\Delta_y|,|\Delta_y'|
                \right\}
                > \delta
                ~ 
                \lor
                ~
                \max\left\{
                    |\Delta_f|,|\Delta_f'|
                \right\}
                > 2\delta
                }
        }{
            \PP{
                E
            }
        }
        \leq 32\varepsilon,
        &
    \end{flalign*}
    as desired.
\end{proof}

\begin{proof}[Proof of \cref{theorem:basic-linear-mitigation}]
    The $\BigO{s}$ bound on the sample and query complexity is immediate from the construction of \cref{algorithm:linear-mitigation-basic}.  The $\BigO{sn}$ bound on the runtime is also immediate under the assumption that arithmetic operations incur unit cost, where we recall that computing a median of a list of length $m$ can be done in time $\BigO{m}$ \citep{DBLP:journals/jcss/BlumFPRT73}. It remains to show correctness of \cref{algorithm:linear-mitigation-basic}. 

    Fix an affine linear function $h: \: \bbR^n \to \bbR$ such that $\LossCutoff{\nicefrac{\delta}{20n}}{\cD}{h} \leq \varepsilon$. Let $\gD^\ideal = h$, and let $y^\ideal = h(x^*)$. For each $i \in [m]$, let $\ell_i \subseteq \bbR^n$ be a line segment from $x^*$ to the unit sphere that passes through $x_i$ and $x_i'$ (such a line exists by \cref{lemma:coupled-sampling}). 
    For any point $x\in\ell_i$, let $r(x) = \|x-x^*\|_2$. Let $\alpha_i,\beta_i \in \bbR$ such that 
    \[
        h_i(r) = \alpha_ir + \beta_i
    \]
    is the $1$-dimensional restriction of $h$ to $\ell_i$, namely, $h_i(r(x))=h(x)$ for all $x \in \ell_i$. Note that $\beta_i = y^\ideal$.
    By \cref{lemma:coupled-sampling}, each $x'_i \sim \uniform{\cX}$. Hence, we may define random variables $y_1,y'_1,\dots,y_m,y'_m \in \bbR$ such that for each $i\in[m]$, $(x_i,y_i) \sim \cD$ and $(x_i',y_i') \sim \cD$.

    Fix $i \in [m]$. Let $\Delta_{y,i} = h(x_i)-y_i$ and $\Delta_{f,i}' = h(x_i')-f(x_i')$.
    Hence,
    \begin{align}
        \label{eq:gi}
        g_i 
        &= 
        (1-\lambda_i) y_i + \lambda_i f(x_i')
        \nonumber
        \\
        &=
        (1-\lambda_i) (h(x_i) - \Delta_{y,i}) + \lambda_i (h(x_i') - \Delta_{f,i}')
        \nonumber
        \\
        &=
        \big(
            (1-\lambda_i)h(x_i) + \lambda_i h(x_i')
        \big)
        -
        \big(
            (1-\lambda_i)\Delta_{y,i} + \lambda_i \Delta_{f,i}'
        \big)
    \end{align}
    Consider each term separately.
    \begin{align}
        \label{eq:gi-linear-term}
        (1-\lambda_i)h(x_i) + \lambda_i h(x_i')
        &=
        (1-\lambda_i) h_i(r_i) 
        + 
        \lambda_i h_i(r_i')
        \tagexplain{Let $r_i = r(x_i)$, $r_i' = r(x_i')$}
        \nonumber
        \\
        &=
        (1-\lambda_i) \left(\alpha_ir_i + \beta_i\right)
        + 
        \lambda_i \left(\alpha_ir_i' + \beta_i\right)
        \nonumber
        \\
        &=
        \beta_i + \alpha_i\left((1-\lambda_i)r_i + \lambda_ir_i'\right)
        \nonumber
        \\
        &=
        \beta_i + \alpha_i\left(\left(1-\frac{r_i}{r_i-r_i'}\right)r_i + \frac{r_i}{r_i-r_i'}\cdot r_i'\right)
        \nonumber
        \\
        &=
        \beta_i = y^\ideal,
    \end{align}
    and,
    \begin{align}
        \label{eq:gi-error-term}
        \Big|
            (1-\lambda_i) \Delta_{y,i} 
            + 
            \lambda_i \Delta_{f,i}' 
        \Big|
        &\leq
        \Big|1-\lambda_i\Big|\cdot|\Delta_{y,i}| 
            + 
        |\lambda_i|\cdot|\Delta_{f,i}'|
        \nonumber
        \\
        &\leq
        \left(
            2|\lambda_i|+1
        \right)
        \cdot
        \max\{|\Delta_{y,i}|,|\Delta_{f,i}'|\} 
        \nonumber
        \\
        &\leq 
        9n
        \cdot
        \max\{|\Delta_{y,i}|,|\Delta_{f,i}'|\},
    \end{align}
    where the final inequality holds when $|\lambda_i|\leq 4n$. 
    Seeing as $\LossCutoff{\nicefrac{\delta}{20n}}{\cD}{h} \leq \varepsilon$ and $\LossCutoff{\nicefrac{\delta}{20n}}{\cD}{f} \leq \varepsilon$, \cref{claim:delta-delta-bound-given-e} implies that 

    \begin{equation}
        \label{eq:bound-delta-i-givenlambda-i}
        \PP{
            \max\{|\Delta_{y,i}|,|\Delta_{f,i}'|\} > 
            \nicefrac{\delta}{10n}
            ~~
            \Bigg|
            ~~
            |\lambda_i| \leq 4n
        }
        \leq 
        32\varepsilon.
    \end{equation}

    By \cref{claim:bound-on-lambda},
    \begin{equation}
        \label{eq:lambda-i-samll-whp}
        \PP{|\lambda_i| \leq 4n} \geq \frac{1}{8}.
    \end{equation}

    Combining \cref{eq:gi,eq:gi-linear-term,eq:gi-error-term,eq:bound-delta-i-givenlambda-i,eq:lambda-i-samll-whp}, we conclude that with probability at least $1/8$, $|\lambda_i|\leq 4n$, and furthermore, 
    \begin{equation}
        \label{eq:gi-y-delta-given-lambda}
        \PP{
            \big|
                g_i-y^\ideal
            \big|
            \leq 0.9\delta
            ~~ \Bigg| ~~
            |\lambda_i|\leq 4n
        } \geq 1-32\varepsilon \geq \frac{2}{3}.
    \end{equation}
    Additionally, by \cref{eq:lambda-i-samll-whp} and Hoeffding's inequality, for $m \geq 320s$, with probability at least $1-2e^{-s}$, there are at least $20s$ samples $(x_i,x_i')$ such that $|\lambda_i| \leq 4n$, namely $|\cI| \geq 20s$. Moreover, by \cref{eq:gi-y-delta-given-lambda} and Hoeffding's inequality, if $|\cI| \geq 20s$ then with probability at least $1-2e^{-s}$, at least half of the values in $\{g_i\}_{i \in \cI}$ satisfy $|g_i-y^\ideal| \leq 0.9\delta$, and this implies that the median $g$ of $\{g_i\}_{i \in \cI}$ also satisfies $|g-y^\ideal| \leq 0.9\delta$. By a union bound,
    \[
        \PP{|g-y^\ideal| \geq 0.9\delta} \leq 4e^{-s},
    \]
    This holds for any $x^* \in \cX$,
    namely,
    \begin{equation}
        \label{eq:g-y-star-close}
        \forall x^* \in \cX: ~ \PP{|g-\gD^\ideal(x^*)| \geq 0.9\delta} \leq 4e^{-s}.
    \end{equation}
    This establishes \cref{item:loss-security-f-independence} in \cref{definition:cutoff-loss-security}. 
    
    Furthermore, seeing as $\gD^\ideal = h$ such that $\LossCutoff{\nicefrac{\delta}{20n}}{\cD}{h} \leq \varepsilon$, it follows from \cref{eq:g-y-star-close}, the triangle inequality $|g-y| \leq |g-h(x)| + |h(x)-y|$ and a union bound that
    \[
        \PPP{(x,y)\sim\cD}{\Big|g-y\Big| \geq \delta} \leq \varepsilon + 4e^{-s}.
    \]
    This establishes \cref{item:loss-security-low-loss} in \cref{definition:cutoff-loss-security}, as desired.~\qedhere
\end{proof}

\subsection{Improved Mitigation for Linear Functions}
\label{section:improved-local-linear-mitigation}

\begin{definition}[Distributions with Benign Noise]
    \label{definition:benign-noise}
    Let $n\in \bbN$, let $\sigma \geq 0$, let $\cX \subseteq B(\mathbf{0},1) \subseteq \bbR^n$ be a nondegenerate convex set contained in the unit ball, and let $U = \uniform{\cX}$ be the uniform distribution on $\cX$. 
    Let $\cQ \in \SubG{\sigma^2}$ be a real-valued distribution that is symmetric about $0$. 
    Let $\cH$ be a class of functions $\bbR^n \to \bbR$.
    For each $h \in \cH$, let $\cD_{h,\cX,\cQ}$ be a distribution over pairs $(x,y) \in \cX \times \bbR$ generated in the following manner:
    \begin{enumerate}
        \item{
            Sample $x \sim U$,
        }
        \item{
            Sample $\eta \sim \cQ$,
        }
        \item{
            Set $y = h(x) + \eta$.
        }
    \end{enumerate}
    The collection of \ul{distributions with labels from $\cH$ and benign noise $\cQ$} is $\bbD_{\cH,\cX,\cQ} = \{\cD_{h,\cX,\cQ}: ~ h \in \cH\}$.
\end{definition}

\begin{algorithmFloat}[!ht]
    \begin{ShadedBox}
        \textbf{Assumptions:}
        \begin{itemize}
            \item{
                $n$, $\cX$, $\Hlinear$ and $\bbD_{\linear,\cX,\cQ}$ as in \cref{definition:benign-noise}.
            }
            \item{
                $\delta \geq 0$; $\varepsilon \in [0,\nicefrac{1}{10}]$.
            }
            \item{
                The algorithm has random sample access to $\cD \in \bbD_{\linear,\cX,\cQ}$.
            }
            \item{
                $\exists h \in \Hlinear: ~ \LossCutoff{\delta/n}{\cD}{h} \leq \varepsilon$.
            }
            \item{
                The algorithm has oracle access to an arbitrary function $f \in \bbR^{\cX}$ (not necessarily in $\Hlinear$) such that $\LossCutoff{\delta/n}{\cD}{f} \leq \varepsilon$.
            }
            \item{
                $s \in \bbN$ is a security parameter.
            }
            \item{
                $x^* \in \cX$ is arbitrary.
            }
        \end{itemize}

        \vspp

        \AdvancedLinearMitigator($x^*$, $n$, $s$):
        \vsp
        \begin{algorithmic}
            \State sample $(x_1,y_1),\dots,(x_s,y_s) \sim \cD^s$
            \vspp

            \State $\cI \gets $ empty set
            \vspp

            \For $i \in [s]$:
                \vsp
                
                \State $x_i' \gets$ \textsc{ResamplingProcedure}($x^*$, $x_i$, $n$) 
                \Comment{See \cref{algorithm:resampling}}
                \vspp

                \State $r_i \gets \|x_i-x^*\|_2$; ~ $r_i' \gets \|x_i'-x^*\|_2$ 
                \vsp

                \State $\displaystyle \lambda_i \gets \frac{r_i}{r_i-r_i'}; ~ \lambda_i' \gets \frac{r_i'}{r_i'-r_i}$
                \vspp

                \If $|\lambda_i| \leq 4n 
                ~ \land ~
                |\lambda_i'| \leq 4n$:
                    \vsp

                    \State $\cI \gets \cI \cup \{i\}$
                    \vsp

                    \State $b_i \gets (f(x_i)-y_i)\cdot\lambda_i'$
                    \vsp

                    \State $g_i \gets (1-\lambda_i) y_i + \lambda_i f(x_i')$
                \EndIf
            \EndFor
            \vspp

            \State $\left\{i_1,i_2,\dots,i_{|\cI|}\right\} \gets \cI$; ~ $\sHalf \gets \left\lfloor|\cI|/2\right\rfloor$
            \vsp

            \State $g \gets \textsc{MeanOfMedians}\left(\left\{g_{i_{2t}} - b_{i_{2t-1}}: ~ t \in [\sHalf]\right\}\right)$
            \Comment See \cref{algorithm:robust-mean}
            \vsp
            
            \State \textbf{output} $g$
        \end{algorithmic}
    \end{ShadedBox}
    \caption{An unbiased cutoff loss secure local mitigator for linear functions $\bbR^n \to \bbR$, with a trade-off between precision and sample complexity.}
    \label{algorithm:linear-mitigation-advanced}
\end{algorithmFloat}

\begin{theorem}
    \label{theorem:advanced-linear-mitigation}
    Let $\varepsilon \in [0,\nicefrac{1}{10}]$ and $\delta \geq 0$. 
    For every index $n\in \bbN$, let $B_n = B(\mathbf{0},1) \subseteq \bbR^n$ be the unit ball; 
    let $\cX_n \subseteq B_n$ be a (nondegenerate) convex set; 
    let $\cH_n$ be the set of affine linear functions $\bbR^n \to \bbR$;
    let $\sigma_n \leq (\delta/n)\cdot(2\ln(2/\varepsilon))^{-1/2}$; 
    let $\cQ_n \in \SubG{\sigma_n^2}$ be a real-valued subgaussian distribution that is symmetric about $0$;
    let $\bbD_n = \bbD_{\cH_n,\cX_n,\cQ_n}$ be the set of distributions with affine linear labels and benign noise, as in \cref{definition:benign-noise}; 
    and let $\bbD = \{\bbD_n\}_{n \in \bbN}$.\footnote{
        In particular, by \cref{claim:concentration-sum-of-subgaussians} and the choice of $\sigma_n$, $\PPP{q \sim \cQ_n}{|q| \geq \delta/n} \leq \varepsilon$. So for every $\cD \in \bbD_n$ there exists an affine function $h \in \cH_n$ such that $\LossCutoff{\delta/n}{\cD}{h} \leq \varepsilon$.
    }
    
    Then \cref{algorithm:linear-mitigation-advanced} defines a local mitigator $M$ that is an
    \begin{equation}
        \label{eq:improved-local-mitigation-with-s}
        \left(
            \varepsilon, 
            ~
            \frac{1}{n}\cdot\delta
        \:
        \longrightarrow 
        \:
        \left(
            \frac{1}{n}
            +
            \frac{\ln(s)}{s^{1/4}}
        \right)\cdot\delta
        \right)
    \end{equation}
    unbiased cutoff loss secure mitigator for distributions $\bbD$ (satisfying \cref{item:loss-security-low-loss,,item:loss-security-f-independence,,item:loss-security-unbiased} in \cref{definition:cutoff-loss-security}), where %
    $s \in \bbN$ is the security parameter.

    In particular, if $s = \ln(n)\sqrt{n}$ and $n$ is large enough, the mitigator is
    \begin{equation}
        \label{eq:improved-local-mitigation-without-s}
        \left(
            \varepsilon, 
            ~
            \frac{1}{n}\cdot\delta
        \:
        \longrightarrow 
        \:
            \frac{1}{n^{1/10}}\cdot\delta
        \right)
    \end{equation}
    unbiased and cutoff loss secure. \cref{algorithm:linear-mitigation-advanced} uses $s$ samples and $2s$ oracle queries, and runs in time $\BigO{sn}$, assuming unit runtime cost for each arithmetic operation on the representations of real numbers involved in the computation.
\end{theorem}

\subsubsection{
    Proof of \texorpdfstring{%
            \cref{theorem:advanced-linear-mitigation}
        }{%
            Advanced Linear Mitigation
        }%
}

\begin{proof}[Proof of \cref{theorem:advanced-linear-mitigation}]
    Fix $n \in \bbN$, $\cD \in \bbD_n$, and let $h \in \cH_n$ such that $\cD = \cD_{h,\cX_n,\cQ_n}$ as in \cref{definition:benign-noise}. By \cref{claim:concentration-sum-of-subgaussians} and the choice of $\sigma_n$, $\LossCutoff{\delta/n}{\cD}{h} \leq \varepsilon$. 
    Let $f: \cX_n \to \bbR$ such that $\LossCutoff{\delta/n}{\cD}{f} \leq \varepsilon$.
    Fix $i \in \cI$.
    Let $\Delta_{f,i} = f(x_i)-h(x_i)$ and $\Delta_{f,i}' = f(x_i')-h(x_i')$. Let $\ell_i$ be the line from $x^*$ to the unit sphere that contains $x_i$ and $x_i'$. Let $r(x) = \|x-x^*\|_2$, and let $\alpha_i,\beta_i \in \bbR$ such that $h_i(r)=\alpha_i \cdot r + \beta_i$ is the $1$-dimensional restriction of $h$ to $\ell_i$, namely, $h_i(r(x))=h(x)$ for all $x \in \ell_i$. Note that $\beta_i = h(x^*)$, and $y_i = h(x_i)+\eta_i$, where $\eta_i \sim \cQ_n \in \SubG{\sigma_n^2}$ with $\EE{\eta_i} = 0$. Similarly, we can define a random variable $y_i' = h(x_i')+\eta_i'$ where $\eta_i' \sim \cQ_n$ is independent.

    Consider the random variable~$g_i$.
    \begin{align*}
        \label{eq:naive-estimator-decomposition}
        g_i 
        &= 
        (1-\lambda_i) y_i + \lambda_i f(x_i')
        \\
        &
        = 
        (1-\lambda_i) (h_i(r_i) + \eta_i) + \lambda_i(h_i(r_i')+\Delta_{f,i}')
        \\
        &
        = 
        \underbrace{
            (1-\lambda_i)h_i(r_i)
            + 
            \lambda_i\cdot h_i(r_i')
        }_{h(x^*)}
        +
        \underbrace{
            (1-\lambda_i) \eta_i
        }_{\text{Noise I}}
        +
        \underbrace{
            \lambda_i\cdot \Delta_{f,i}'
        }_{(\star)}
        .
    \end{align*}
    Examine each term separately. The first term equals $h(x^*)$ for any $x_i,x_i' \in \ell_i$, as follows.
    \begin{align*}
        (1-\lambda_i)h_i(r_i) 
        + 
        \lambda_i\cdot h_i(r_i')
        &=
        (1-\lambda_i)(\alpha_i\cdot r_i + \beta_i)
        + 
        \lambda_i\cdot (\alpha_i\cdot r_i' + \beta_i)
        \\
        &=
        \beta_i
        +
        (1-\lambda_i)(\alpha_i\cdot r_i)
        + 
        \lambda_i\cdot \alpha_i\cdot r_i'
        \\
        &=
        \beta_i
        +
        \left(1-\frac{r_i}{r_i-r_i'}\right)(\alpha_i\cdot r_i)
        + 
        \frac{r_i}{r_i-r_i'}\cdot (\alpha_i\cdot r_i')
        \\
        &=\beta_i = h(x^*).
    \end{align*}
    The term $(\star)$ can be decomposed into a bias term and a noise term, as follows. 
    \begin{align*}
        (\star)
        &=
        \lambda_i\cdot \Delta_{f,i}'
        \\
        &=
        \lambda_i\cdot\big(f(x_i')-h(x')\big)
        \\
        &=
        \lambda_i\cdot\big(f(x_i')-y_i'+\eta_i'\big)
        \\
        &=
        \underbrace{
            \lambda_i\cdot\big(f(x_i')-y_i'\big) 
        }_{\text{Bias}}
        + 
        \underbrace{
            \lambda_i\cdot\eta_i'
        }_{\text{Noise II}}.
    \end{align*}
    In sum, we have the following central identity
    \begin{equation}
        \label{eq:gi-decomposition}
        \mathcolorbox{myLipicsLightGray}{
            g_i = h(x^*) + b_i' + N_i,
        }
    \end{equation}
    where
    $
        b_i'    = \lambda_i\cdot(f(x_i')-y_i')
    $
    is a bias term, and 
    $
        N_i = 
        (1-\lambda_i) \cdot\eta_i
        +
        \lambda_i\cdot\eta_i'
    $ is a noise term. We analyze the bias term and the noise term separately. 
    
    For the noise term $N_i$ in \cref{eq:gi-decomposition}, seeing as $\eta_i$ is symmetric about $0$ and $\eta_i \,\bot\, \lambda_i$, it follows that $N_i^0 = (1-\lambda_i) \cdot\eta_i$ is also symmetric about~$0$. A similar argument holds for $N_i^1 = \lambda_i\cdot\eta_i'$, and therefore the noise term $N_i = N_i^0 + N_i^1$ is distributed symmetrically about~$0$. 
    
    Furthermore, $1-\lambda_i = \lambda_i'$, and $\max\{|\lambda_i|, |\lambda_i'|\} \leq 4n$ for $i \in \cI$. Hence, by \cref{claim:subgaussian-times-bounded}, $N_i^0$ and $N_i^1$ satisfy $N_i^0,N_i^1 \in \SubG{16n^2\sigma_n^2}$. By \cref{claim:sum-of-subgaussians}, $N_i = N_i^0 + N_i^1 \in \SubG{\xi^2}$ where 
    \[
        \xi^2 
        = 
        2^2\cdot16n^2\sigma_n^2 
        \leq 
        64\cdot \frac{\delta^2}{2\log(2/\varepsilon)} 
        \leq
        64\delta^2.
    \]
    By \cref{claim:concentration-sum-of-subgaussians}, this implies that 
    \begin{equation}
        \label{eq:noise-subgaussian}
        \PP{\left|N_i\right| \geq 20\delta} 
        \leq
        2\expf{-\frac{(20\delta)^2}{2\xi^2}} \leq \frac{1}{10}.
    \end{equation}

    Next, for the bias term $b_i'$ in \cref{eq:gi-decomposition}, note that \cref{algorithm:linear-mitigation-advanced} uses quantities of the form $g_i-b_j$, where subtracting $b_j$ is intended to ``cancel out'' the $b_i'$ term in $g_i$. Consider the expression $b_i'-b_j$. For any fixed $j \neq i$, \cref{lemma:coupled-sampling} and the bound $\LossCutoff{\delta/n}{\cD}{f} \leq \varepsilon$ imply that
    \[
        \PP{\max\left\{\left|f(x_i')-y_i'\right|, \left|f(x_j)-y_j\right|\right\} > \frac{\delta}{n}
        ~~
        \Big|
        ~
        i,j \in \cI\: }
        \leq
        2\varepsilon.
    \]
    So 
    \begin{equation}
        \label{eq:bound-on-delta-b}
        \PP{\left|b_i'-b_j\right| > 4\delta
        ~~
        \Big|
        ~
        i,j \in \cI\: }
        \leq
        2\varepsilon.
    \end{equation}
    Furthermore, by \cref{lemma:coupled-sampling}, $(x_i',y_i') \stackrel{d}{=} (x_j,y_j)$, so
    \[
        b_j 
        =
        (f(x_j)-y_j)\cdot\frac{r_j'}{r_j'-r_j}
        \stackrel{d}{=}
        (f(x_i')-y_i')\cdot\frac{r_i}{r_i-r_i'}
        =
        b_i'
    \]
    Because $b_j \: \bot \: b_i'$, 
    this implies that $b_i'-b_j$ is distributed symmetrically about $0$.\footnote{
        Here we use the assumption $b_j \: \bot \: b_i'$, which holds because the samples for iteration $i$ are independent of the samples for iteration $j$. It might be tempting to consider a simpler algorithm that does not partition $\cI$ into two disjoint sets of size $\sHalf$, and instead simply uses estimates of the form $g_i-b_i$ for all $i \in \cI$. This leads to terms of the form $b_i'-b_i$, where $b_i'$ and $b_i$ are identically distributed but are not independent. To see why that might not be good enough, recall that for a joint distribution $(X,Y)$ with $X \stackrel{d}{=} Y$, the variables $(X,Y)$ might not be exchangeable, and in particular $X-Y$ might not be symmetric. For example, consider a distribution of $(X,Y)$ that is uniform over $\{(-1,0),(0,1),(1,-1)\}$. Then $X-Y$ is not symmetric. To summarize, by using $g_i-b_j$ with $i\neq j$, we get a symmetric distribution while avoiding questions of exchangeability.
    }

    Let $\cP_{g-b}$ be the distribution of $g_i-b_j$ for distinct $i,j \in \cI$, and let $\cP_\good$ be the conditional distribution of
    \[
        z \sim \cP_{g-b}\,\Big|\big(
            z \in 
            \left[
                h(x^*)-24\delta, 
                h(x^*)+24\delta
            \right] 
        \big).
    \]
    Putting the facts that $N_i$ and $b_i'-b_j$ are symmetric about $0$ together with \cref{eq:gi-decomposition,eq:noise-subgaussian,eq:bound-on-delta-b} gives that $\cP_{g-b}$
    can be written as a mixture
    \[
        \cP_{g-b} = (1-\alpha)\cdot\cP_\good + \alpha\cdot\cP_\bad,
    \]
    such that
    \begin{itemize}
        \item{
            $\alpha \leq \nicefrac{1}{10}+2\varepsilon \leq \nicefrac{1}{3}$,
        }
        \item{
            $\cP_{g-b}$ is distributed symmetrically about $h(x^*)$, and
        }
        \item{
            $\PPP{z \sim \cP_\good}{\left|z - h(x^*)\right| \leq 24\delta} = 1$.
        }
    \end{itemize}
    By \cref{theorem:robust-mean}, the estimate $g$ returned by \cref{algorithm:linear-mitigation-advanced} satisfies:
    \begin{enumerate}
        \item{
            $g$ is unbiased, i.e., $\EE{g} = h(x^*)$. So \cref{algorithm:linear-mitigation-advanced} is unbiased, satisfying \cref{item:loss-security-unbiased} in \cref{definition:cutoff-loss-security}.
        }
        \item{
            $g$ is concentrated, namely,
            \[
                \forall \eta \geq 0: ~ \PPP{\cS \sim \cD^s}{\Big| g - h(x^*) \Big| \geq \eta} 
                \leq
                4\expf{-\gamma \sqrt{s}},
            \] 
            where $\gamma = {\frac{1}{4}\cdot\min\left\{\frac{1}{100}, \frac{2\eta^2}{(24\delta)^2}\right\}}$. So for $\eta_s = \delta\ln(s)/s^{1/4}$ as in the statement,
            \begin{align}
                \label{eq:advanced-mitigator-concentration}
                \PPP{\cS \sim \cD^s}{\Big| g - h(x^*) \Big| \geq \eta_s} 
                &\leq
                4\expf{
                    -\frac{\ln(s)^2}{2\cdot(24)^2\cdot s^{1/2}}
                    \cdot \sqrt{s}
                }
                \leq
                4\expf{
                    -\frac{\ln(s)^2}{2\cdot(24)^2}
                }
                \in \negligible(s).
                \footnotemark\\[-2em]\nonumber
            \end{align}%
            \footnotetext{%
                    \cref{eq:advanced-mitigator-concentration} is for the case $\gamma < 1/400$. For the case $\gamma = 1/400$, the upper bound is clearly negligible in $s$.
            }%
            This shows that \cref{algorithm:linear-mitigation-advanced} satisfies cutoff loss security (\cref{item:loss-security-f-independence} in \cref{definition:cutoff-loss-security}) with parameter $\delta_1 = \eta_s$, which is better than required in \cref{eq:improved-local-mitigation-with-s}.
            Additionally, combining \cref{eq:advanced-mitigator-concentration} with $\LossCutoff{\delta/n}{\cD}{h} \leq \varepsilon$ gives
            \[
                \PPPunder{\substack{(x,y) \sim \cD \\ y^* \gets M^{f,\cD}(x, 1^n, 1^s)}}{\Big| y^* - y \Big| \geq \frac{\delta}{n} + \eta_s} 
                \leq 
                \PPPunder{(x,y) \sim \cD}{\Big|y - h(x)\Big| \geq \frac{\delta}{n} ~ \lor ~ \Big|h(x) - y^*\Big| \geq \eta_s}
                \leq
                \varepsilon + \negligible(s).
            \]
            So \cref{algorithm:linear-mitigation-advanced} satisfies the accuracy requirement (\cref{item:loss-security-low-loss} in \cref{definition:cutoff-loss-security}) with parameters 
            $
                \delta/n
                    \to
                \delta\cdot(1/n + \ln(s)/s^{1/4})
            $
            as in \cref{eq:improved-local-mitigation-with-s}.
            In particular, for $s = \ln(n)\sqrt{n}$, %
            \begin{align*}
                \frac{\delta}{n} + \eta_s 
                &= 
                \delta\cdot\left(
                    \frac{1}{n}
                    + \frac{\ln(s)}{s^{1/4}}
                \right)
                = 
                \delta\cdot\left(
                    \frac{1}{n}
                    + 
                    \frac{\ln\big(\ln(n)\sqrt{n}\big)}{\big(\ln(n)\sqrt{n}\big)^{1/4}}
                \right)
                \\
                &
                \leq 
                \delta\cdot\left(
                    \frac{1}{n}
                    + 
                    \frac{\ln(n)}{n^{1/8}}
                \right)
                \leq 
                \delta\cdot\frac{1}{n^{1/10}},
            \end{align*}
            where the last inequality holds for $n$ large enough. Hence, \cref{algorithm:linear-mitigation-advanced} also satisfies the accuracy requirement with parameters 
            $
                \delta/n
                    \to
                \delta/n^{1/10}
            $
            as in \cref{eq:improved-local-mitigation-without-s}.
        }
    \end{enumerate}
    Thus, \cref{algorithm:linear-mitigation-advanced} satisfies \cref{item:loss-security-low-loss,,item:loss-security-f-independence,,item:loss-security-unbiased} in \cref{definition:cutoff-loss-security} with the parameters in \cref{eq:improved-local-mitigation-with-s,eq:improved-local-mitigation-without-s}, as desired.
\end{proof}

\subsection{Mitigation for Multivariate Polynomials}
\label{section:local-polynomial-mitigation}

\begin{definition} For $d+1$ values $v_0, \cdots, v_d \in \bbR$, we define the Vandermonde matrix $\Vand(v_0, \cdots, v_d)$ to be the matrix $V \in \bbR^{(d+1) \times (d+1)}$ given by
\[ V = \begin{bmatrix}
    1 & v_0 & v_0^2 & \cdots & v_0^d\\
    1 & v_1 & v_1^2 & \cdots & v_1^d\\
    \vdots & \vdots & \vdots & \ddots & \vdots\\
    1 & v_d & v_d^2 & \cdots & v_d^d
\end{bmatrix}. \]
Symbolically, for all $i, j \in \{0, \cdots, d\}$, $V_{i,j} = v_i^j$.
\end{definition}

We recall the definition of the infinity norm $\norm{\cdot}_{\infty}$ on matrices, induced by the infinity norm $\norm{\cdot}_{\infty}$ on vectors. 
\begin{definition}
For a matrix $M \in \bbR^{\ell \times \ell}$, the infinity norm of $M$, denoted $\norm{M}_{\infty}$, is defined by
\[ \norm{M}_{\infty} = \sup_{v \in \bbR^n \setminus \{0\}} \frac{\norm{Mv}_{\infty}}{\norm{v}_{\infty}}, \]
where we use the standard $\ell_{\infty}$ norm for vectors $u \in \bbR^n$, formally given by
\[ \norm{u}_{\infty} = \max_{i \in [n]} |u_i|. \]
\end{definition}
We now reference an upper bound on the infinity norm of the inverse of Vandermonde matrices.
\begin{theorem}[Theorem 1 of \citealp{gautschi1962inverses}]\label{inverse-vandermonde-math}
    Let $V = \Vand(v_0, \cdots, v_d)$, where $v_i \neq v_j$ for all $i \neq j \in \{0, \cdots, d\}$. Then,
    \[ \norm{V^{-1}}_{\infty} \leq \max_{0 \leq j \leq d} \prod_{\substack{0 \leq i \leq d\\i \neq j}} \frac{1 + |v_i|}{|v_i - v_j|}. \]
\end{theorem}

\begin{figure}[H]
    \centering
    \includegraphicsorstandalone[width=0.5\linewidth]{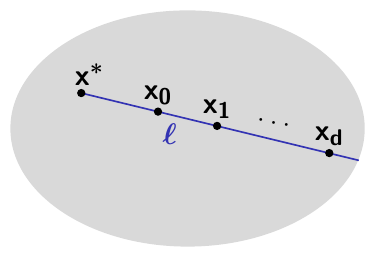}
    \caption{%
        \cref{algorithm:polynomial-mitigation-basic} uses the {\Resamp} of \cref{lemma:coupled-sampling} to sample \( d + 1 \) points on a straight line with \( x^* \) such that each point has a uniform marginal.
    }
    \label{figure:correlated-sampling-polynomial}
    \vspace{1em}
\end{figure}

In Algorithm~\ref{algorithm:polynomial-mitigation-basic}, we give pseudocode to describe the mitigation algorithm used for multivariate polynomials.

\begin{algorithmFloat}
    \begin{ShadedBox}
        \textbf{Assumptions:}
        \begin{itemize}
            \item{
                $n, d \in \bbN$; $\varepsilon \in [0,\nicefrac{1}{20d}]$; $\delta_1 \geq 0$; $\delta_0 = \nicefrac{\delta_1}{4 \cdot (80 nd^2)^d}$; $\cX \subseteq B(\mathbf{0},1) \subseteq \bbR^n$; $\cX$ convex.
            }
            \item{
                $\cD \in \distribution{\cX \times \bbR}$ with $\cD_{\cX} = \uniform{\cX}$.
            }
            \item{
                $\cH$ is the class of polynomial functions $\bbR^n \to \bbR$ of total degree at most $d$.
            }
            \item{
                $\exists h \in \cH: ~ \LossCutoff{\delta_0}{\cD}{h} \leq \varepsilon$.
            }
            \item{
                The algorithm has oracle access to an arbitrary function $f \in \bbR^{\cX}$ (not necessarily in $\cH$) such that $\LossCutoff{\delta_0}{\cD}{f} \leq \varepsilon$.
            }
            \item{
                $s \in \bbN$ is a security parameter; $m = s$.
            }
            \item{
                $x^* \in \cX$ is arbitrary.
            }
        \end{itemize}

        \vspp

        \textsc{LocalPolynomialMitigator}($n$, $d$, $m$, $x^*$):
        \vsp
        \begin{algorithmic}

            \State Sample $x_1, \dots, x_m \gets U(\cX)$.
            \vspp

            \State $\cG \gets $ empty set
            \vspp

            \For $i \in [m]$:
                \vsp
                \State $\left(x_i^{(0)}, \widehat{y}_i^{(0)}\right) \gets (x_i, f(x_i))$\vsp 
                \For $j \in [d]$:
                    \vsp
                    \State $x_i^{(j)} \gets \Resamp$($x^*$, $x_i$, $n$) 
                    \Comment{See \cref{algorithm:resampling}}
                    \vsp
                    \State $\widehat{y}_i^{(j)} \gets f\left(x_i^{(j)}\right)$
                    \State Let $t_i$ be the length of the line segment $\ell_i$ from $x^*$ to the boundary of $\cX$ that\\ \quad \quad \quad  passes through $x_i$.
                    \State $r_i^{(j)} \gets \dfrac{\|x_i^{(j)}-x^*\|_2}{t_i} \in [0,1]$;
                \EndFor
                \vsp
                \State $R_i \gets \Vand\left(r_i^{(0)}, \cdots, r_i^{(d)}\right) \in \bbR^{(d+1) \times (d+1)}$
                \State $\widehat{y}_i \gets \left(\widehat{y}_i^{(0)}, \cdots, \widehat{y}_i^{(d)}\right) \in \bbR^{d+1}$
                \State $\widehat{\alpha}_i \gets R_i^{-1} \widehat{y}_i \in \bbR^{d+1}$
                \State $\cG \gets \cG \cup \widehat{\alpha}_i^{(0)}$, where $\widehat{\alpha}_i^{(0)} \in \bbR$ is the first coordinate of $\widehat{\alpha}_i$.
            \EndFor
            \vsp
            \State $g \gets $ median of $\cG$
            \vsp
            \State \textbf{output} $g$
        \end{algorithmic}
    \end{ShadedBox}
    \caption{A basic local independent mitigator for low degree polynomials $\bbR^n \to \bbR$.}
    \label{algorithm:polynomial-mitigation-basic}
\end{algorithmFloat}

We now prove that the resampled points in Algorithm~\ref{algorithm:polynomial-mitigation-basic} are sufficiently pairwise far, so that we can invoke the bound given in Theorem~\ref{inverse-vandermonde-math}.

\begin{claim}\label{claim:poly-mitigation-vandermonde-points-are-far-apart}
    In the notation of Algorithm~\ref{algorithm:polynomial-mitigation-basic}, for each $i \in [m]$,
    \[ \PP{ \forall j \neq j' \in \{0, \cdots, d\}\!\!: \left|r_i^{(j)} - r_i^{(j')} \right| \geq \frac{1}{ 40 nd^2}} \geq \frac{5}{6}. \]
\end{claim}

\begin{proof}
By construction of the {\Resamp} as in Algorithm~\ref{algorithm:resampling}, we know the density of each $t_i \cdot r_i^{(j)}$ is sampled i.i.d. (over $j \in \{0, \cdots, d\}$) from the density function $q(r) = \frac{n}{t_i^n} r^{n-1}$, where $t_i$ is the length of the line segment from $x^*$ to the boundary of $\cX$ that passes through $x$. By dividing by $t_i$, this density becomes $q'(r) := t_i \cdot q(r \cdot t_i) = n \cdot r^{n-1}$, with support $[0,1]$. That is, each $r_i^{(j)}$ is sampled i.i.d. (over $j \in \{0, \cdots, d\}$) from $q'$.

We partition $[0,1]$ into $20d^2$ consecutive intervals $I_1, \cdots, I_{20d^2}$ such that each interval has equal mass under $q'$ (i.e., we split up $q'$ into $20d^2$ quantiles). We arrange the intervals so that $0 \in I_1$ and $1 \in I_{20d^2}$. We now give a birthday-style argument that shows that with probability at least $5/6$, it holds that all $r_i^{(j)}$ (over $j \in \{0, \cdots d\}$, so $i$ is fixed) lie in distinct and non-neighboring intervals. Let $Z_i$ be the random variable denoting the number of $(j, j')$ pairs that lie in the same or neighboring intervals. By linearity of expectation,
\begin{align*} \EE{Z_i} &= \sum_{j \neq j' \in \{0, \cdots, d\}} \PP{r_i^{(j)}, r_i^{(j')} \text{ are in the same or neighboring intervals}}
\\&\leq \binom{d+1}{2} \cdot \frac{3}{20d^2} \leq \frac{1}{6}.
\end{align*}
Since $Z_i$ is non-negative, by Markov's inequality, $\PP{Z_i \geq 1} \leq 1/6$, so $\PP{Z_i = 0} \geq 5/6$. Therefore, with probability at least $5/6$, we know that all $r_i^{(j)}$ (over $j \in \{0, \cdots d\}$) lie in distinct and non-neighboring intervals.

Given that $Z_i = 0$, since all $r_i^{(j)}$ are separated from each other by at least one full interval, we know that 
\[ \min_{j \neq j'}{\left|r_i^{(j)} - r_i^{(j')} \right|} \geq \min_{k \in [20d^2]} |I_k|, \]
where we use the notation $|I_k| \in \bbR$ to denote the length of the interval $I_k$. By monotonicity of $q'$, we know that
\[ \min_{k \in [20d^2]} |I_k| = |I_{20d^2}|,\]
where the length $z := |I_{20d^2}|$ is given by the solution $z$ to the equation
\[ \int_{1-z}^1 n r^{n-1} dr = \frac{1}{20d^2}. \]
Since $n,d \geq 1$, solving this equation yields
\[z = 1 - \left( 1 - \frac{1}{20d^2} \right)^{1/n} \geq 1 - e^{-1/(20nd^2)} \geq 1 - \left(1 - \frac{1}{40nd^2} \right) = \frac{1}{40nd^2}.\]
Therefore, with probability at least $5/6$, we have
    \[ \min_{j \neq j'}{\left|r_i^{(j)} - r_i^{(j')} \right|} \geq \frac{1}{40nd^2}, \]
as desired.
\end{proof}

We finally prove that that Algorithm~\ref{algorithm:polynomial-mitigation-basic} is a local mitigator for degree-$d$ multivariate polynomials.

\begin{theorem}\label{theorem:basic-polynomial-mitigation}
    For any $n, d \in \bbN$, let $B_n = B(\mathbf{0},1) \subseteq \bbR^n$ be the unit ball, and let $\cX_n \subseteq B_n$ be a nondegenerate convex set. 
    Let $\cH_n$ be the set of multivariate polynomials $\bbR^n \to \bbR$ of total degree at most $d$. 
    Let $\varepsilon \in [0,\nicefrac{1}{20d}]$, $\delta_1 \in \bbR_{\geq 0}$ and $\delta_0 = \nicefrac{\delta_1}{ 4\cdot (80nd^2)^{d}}$. Let $\bbD_n = \bbD_{n,\varepsilon,\delta_0}$ be the collection of distributions $\cD$ with uniform marginal on $\cX_n$ such that $\LossCutoff{\delta_0}{\cD}{\cH_n} \leq \varepsilon$, i.e., there exists a multivariate polynomial $h : \bbR^n \to \bbR$ of total degree at most $d$ such that
    \[ \PPPunder{(x,y) \sim \cD}{\left| h(x) - y \right| > \delta_0} \leq \varepsilon. \]

    Then \cref{algorithm:polynomial-mitigation-basic} is a local mitigator that is $(\varepsilon, \delta_0 \to \delta_1)$-cutoff loss secure (per Definition~\ref{definition:cutoff-loss-security}) for distributions $\bbD_n$, where $s \in \bbN$ is the security parameter.

    Furthermore, \cref{algorithm:polynomial-mitigation-basic} uses a total of $s (d+1)$ oracle queries, and runs in time $s \cdot \poly{d, n}$, assuming unit runtime cost for each arithmetic operation on the representations of real numbers involved in the computation.
    
\end{theorem}

\begin{proof}
    Let $U = \uniform{\cX_n}$ be the uniform distribution on $\cX_n$. Fix a population distribution $\cD \in \bbD_{n,\varepsilon, \delta_0}$ and a multivariate polynomial $h : \bbR^n \to \bbR$ of total degree at most $d$ such that $\LossCutoff{\delta_0}{\cD}{h} \leq \varepsilon$. We set $\gD^\ideal(x) = h(x)$. Fix the adversarially chosen function $f : \bbR^n \to \bbR$ with loss $\LossCutoff{\delta_0}{\cD}{f} \leq \varepsilon$. That is,
    \begin{equation}\label{eqn:polynomial-loss-guarantee}
    \PPPunder{(x, y) \sim \cD}{|h(x) - y| \geq \delta_0} \leq \varepsilon,\;\;\;\;
    \PPPunder{(x, y) \sim \cD}{|f(x) - y| \geq \delta_0} \leq \varepsilon.
    \end{equation}
    By applying the triangle inequality and the union bound and looking at the marginal on $\cX_n$,
    \begin{equation}\label{eqn:polynomial-loss-guarantee-f-h}
    \PPP{x \gets U}{|f(x) - h(x)| \geq 2\delta_0} \leq 2\varepsilon.
    \end{equation}
    Let $y^\ideal = h(x^*)$. For each $i \in [m]$, consider the line $\ell_i : \bbR \to \bbR^n$ passing through $x^*$ and $x_i$, scaled so that $\ell_i(0) = x^*$ and $\ell_i(1)$ is on the boundary of $\cX_n$. Let $h_i : \bbR \to \bbR$, $h_i := h \circ \ell_i$ be the \emph{univariate} polynomial of $h$ restricted to the line $\ell_i$, which has degree at most $d$. Let $\alpha_i = (\alpha_i^{(0)}, \cdots, \alpha_i^{(d)}) \in \bbR^{d+1}$ be the coefficients such that
    \[ h_i(r) = \sum_{k = 0}^d \alpha_i^{(k)} r^k. \]
    By construction,
    \[ \alpha_i^{(0)}  = h_i(0) = h(\ell_i(0)) = h(x^*) = y^\ideal.\]
    Let $r_i^{(0)}, \cdots, r_i^{(d)} \in \bbR$ be as in Algorithm~\ref{algorithm:polynomial-mitigation-basic}. Note that for all $j \in \{0, \cdots, d\}$,
    \[ \ell_i\left(r_j^{(j)} \right) = x_i^{(j)}. \]
    
    Letting $R_i = \Vand\left(r_i^{(0)}, \cdots, r_i^{(d)}\right) \in \bbR^{(d+1) \times (d+1)}$ and $y_i = \left(h_i\left(r_i^{(0)} \right), \cdots, h_i\left(r_i^{(d)}\right) \right) \in \bbR^{d+1}$, we have the matrix equation
    \[ R_i \alpha_i = y_i,\]
    and therefore
    \[ \alpha_i = R_i^{-1} y_i.\]
    (Note that since $R_i$ is a Vandermonde matrix, it is invertible as all of the $r_i^{(j)}$ are distinct, which will be true with probability $1$.)

    Using the notation in our algorithm, since for all $j \in \{0, \cdots, d\}$, we have $y_i^{(j)} = h_i(r_i^{(j)}) = h(x_i^{(j)})$ and $\widehat{y}_i^{(j)} = f(x_i^{(j)})$, we know $y_i^{(d)} = h_i(r_i^{(j)}) = h(x_i^{(j)})$ and $\widehat{y}_i^{(j)} = f(x_i^{(j)})$, so by Eq.~\eqref{eqn:polynomial-loss-guarantee-f-h},
    \[\PP{\left|\widehat{y}_i^{(j)} - y_i^{(j)}\right| \geq 2\delta_0} \leq 2\varepsilon. \]
    By applying the union bound over $j \in \{0, \cdots, d\}$, we get
    \[\PP{\exists j \in \{0, \cdots, d\}, \left|\widehat{y}_i^{(j)} - y_i^{(j)}\right| \geq 2\delta_0} \leq (2d+2)\varepsilon \leq 4d\varepsilon \leq \frac{1}{5}. \] Therefore, with probability at least $4/5$,
    \[ \norm{\widehat{y}_i - y_i}_{\infty} \leq 2\delta_0. \]
    Recall that $\widehat{\alpha}_i = R_i^{-1} \widehat{y}_i$. Given the above, we have
    \begin{align*}
        \norm{\widehat{\alpha}_i - \alpha_i}_{\infty} = \norm{R_i^{-1} \widehat{y}_i - R_i^{-1} y_i}_{\infty} = \norm{R_i^{-1} (\widehat{y}_i - y_i)}_{\infty} \leq \norm{R_i^{-1}}_{\infty} \norm{\widehat{y}_i - y_i}_{\infty} \leq \norm{R_i^{-1}}_{\infty} 2 \delta_0.
    \end{align*}
    We now apply Theorem~\ref{inverse-vandermonde-math} along with Claim~\ref{claim:poly-mitigation-vandermonde-points-are-far-apart} (and the union bound) to see that with probability at least $1 - 1/5 - 1/6 = 19/30$,
    \begin{align*}
        \norm{R_i^{-1}}_{\infty} \leq \max_{0 \leq j \leq d} \prod_{\substack{0 \leq j' \leq d\\j' \neq j}} \frac{1 + \left|r_i^{(j)} \right|}{\left|r_i^{(j)} - r_i^{(j')}\right|} \leq \prod_{\substack{0 \leq j' \leq d\\j' \neq j}} \frac{2}{\frac{1}{40nd^2}} = (80nd^2)^d.
    \end{align*}
    Therefore, in this case,
    \[ \norm{\widehat{\alpha}_i - \alpha_i}_{\infty} \leq (80nd^2)^d \cdot 2\delta_0 = \frac{\delta_1}{2},\]
    and in particular, looking at the first coordinate,
    \[ \left|\widehat{\alpha}_i^{(0)} - y^\ideal \right|  = \left|\widehat{\alpha}_i^{(0)} - \alpha_i^{(0)} \right| \leq \frac{\delta_1}{2}. \]
    This, for each $i \in [m]$, with probability at least $19/30$, we add an estimate to $\cG$ that is $\delta_1/2$-close to $y^\ideal=h(x^*) = \gD^\ideal(x^*)$. By a standard Chernoff bound, it follows that $g$, the median of $\cG$ at the end of Algorithm~\ref{algorithm:polynomial-mitigation-basic}, satisfies
    \[ \PP{|g -  \gD^\ideal(x^*)| > \delta_1/2} \leq e^{-\frac{m \cdot (1/10^2)}{2}} \leq e^{-m/200} = 2^{-\Omega(s)}. \]
    This holds for any $x^* \in \cX_n$, so we (in particular) have
    \begin{equation}\label{eqn:mitigator-output-from-ideal}
        \forall x^* \in \cX_n: ~ \PP{|g-\gD^\ideal(x^*)| \geq \frac{3\delta_1}{4}} \leq 2^{-\Omega(s)}.
    \end{equation}
    This establishes \cref{item:loss-security-f-independence} in \cref{definition:cutoff-loss-security}. 

    To see \cref{item:loss-security-low-loss} in \cref{definition:cutoff-loss-security}, we can use the triangle inequality, the union bound, Eq.~\eqref{eqn:mitigator-output-from-ideal}, and Eq.~\eqref{eqn:polynomial-loss-guarantee} (since $\gD^\ideal(x^*) = h(x)$) to see that
    \[ \PPPunder{\substack{(x,y) \sim \cD \\g \gets M^{f}(x, 1^n, 1^s)}}{|g - y| \geq \delta_0 + \frac{3\delta_1}{4}} \leq \varepsilon + 2^{-\Omega(s)}. \]
    Since $\delta_0 + 3\delta_1/4 \leq \delta_1$, this establishes \cref{item:loss-security-low-loss} in \cref{definition:cutoff-loss-security}.

\end{proof}

\begin{remark}\label{remark:n-to-d-blowup-necessary}
While the $n^{O(d)}$ blowup in the $\delta$ error might seem undesirable, we show that this is unavoidable for a related notion of correction in the \emph{exact} setting, defined as follows. 

Let $\bbD_{n, \delta_0}$ be the collection of distributions $\cD$ with uniform marginal on $\cX_n$ such that $L_{\cD}^{>\delta_0}(\cH_n) = 0$, i.e., there exists a multivariate polynomial $h : \bbR^n \to \bbR$ of total degree at most $d$ such that $\PPP{(x, y) \sim \cD}{|h(x) - y| > \delta_0} = 0$. Instead of cutoff loss security (\cref{item:loss-security-f-independence} in \cref{definition:cutoff-loss-security}), suppose we want the stronger property that for all $f : \cX_n \to \bbR$ such that $\LossCutoff{\delta_0}{\cD}{f} \leq \varepsilon$,
\begin{equation}
    \label{eq:exact-recovery}
    \forall x^* \in \cX_n:
    \!\!\!{\PPPunder{y^* \gets M^{f}(x^*,1^n, 1^s)}{
                    \left| y^* - h(x^*) \right| \geq \delta_1
    } \leq \mu(s)},
\end{equation}
where the mitigator $M^{f}(x, 1^n, 1^s)$ has access to $f$ (but not $\cD$).\footnote{
    The difference between the requirement in \cref{eq:exact-recovery} and the requirement of \cref{item:loss-security-f-independence} in \cref{definition:cutoff-loss-security} is that in \cref{eq:exact-recovery}, $y^*$ must be close to the true labeling function $h$ (which is unknown to the defender), whereas in \cref{definition:cutoff-loss-security} $y^*$ must be close to a canonical function $\gD^\ideal$ chosen by the defender.
}

We claim two facts in this setting:
\begin{itemize}
    \item{Algorithm~\ref{algorithm:polynomial-mitigation-basic} achieves the security requirement of \cref{eq:exact-recovery}.  This follows directly from the analysis given in the proof of \cref{theorem:basic-polynomial-mitigation}. 
    }
    \item The multiplicative blowup of $\delta_1 = n^{\Omega(d)} \delta_0$ is unavoidable.
\end{itemize}

We now explain why the blowup is unavoidable, assuming that the mitigator only has oracle access to the potentially-backdoored $\tilde{f}$ (but does not have random samples from the population distribution).\footnote{
    A slightly more elaborate argument can show that the blowup is unavoidable even if the mitigator does have random samples from the population distribution.
}
To see this, let $d \in \bbN$ be even, and consider the following polynomials of degree at most $d$,
\[ h_0(x_1, \cdots, x_n) \equiv 0,\; \; \text{ and } \; \; h_1(x_1, \cdots, x_n) = \left(1 - \sum_{i = 1}^n x_i^2 \right)^{d/2},\]
along with the uniform distribution $U = \uniform{B}$ over the unit ball in $\bbR^n$. At the origin, i.e., $x^* = \mathbf{0}$, we have $h_b(x^*) = b$ for $b \in \{0,1\}$. However, with high probability over the unit ball, these polynomials are very close:
\begin{align*} \PPPunder{x \sim U}{\big|h_1(x)\big| \geq \frac{1}{n^{d/4}}} = \PPPunder{x \sim U}{1 - \sum_{i = 1}^n x_i^2 \geq \frac{1}{\sqrt{n}}} &=\PPPunder{x \sim U}{\sqrt{\sum_{i = 1}^n x_i^2} \leq \left(1 - \frac{1}{\sqrt{n}} \right)^{1/2}}
\\& = \left( 1 - \frac{1}{\sqrt{n}} \right)^{n/2} \leq e^{-\sqrt{n}/2}.
\end{align*}
That is, there are two polynomials of degree at most $d$ that are additively $1/n^{d/4}$-close with probability at least $1 - e^{-\sqrt{n}/2}$, yet they differ additively at the origin by $1$. Therefore, we can fix $f = h_0 \equiv 0$ and examine the distribution of $M^f(\mathbf{0}, 1^n, 1^s)$. Considering $\cD$ to be exactly given by $h_0$ or $h_1$, in both cases, we have shown that $\LossCutoff{\delta_0}{\cD}{f} \leq e^{-\sqrt{n}/2}$ for $\delta_0 = 1/n^{d/4}$. However, to satisfy our stronger security notion, we would need
\[ \PP{
                    \left| M^{f}(\mathbf{0}, 1^n, 1^s) - h_b(\mathbf{0}) \right| \geq \delta_1
                } \leq \mu(s) \]
for both values $b \in \{0,1\}$. Since $h_b(\mathbf{0}) = b$ and $\mu$ is negligible, this is only possible if $\delta_1 = \Omega(1)$. Therefore, $\delta_1 = n^{\Omega(d)} \cdot \delta_0$, as desired. \qed
\end{remark}

    \else
        
    \fi

    \ifdraftcompile
        \section{Robust Mean Estimation}
\label{section:robust-mean-estimation}

We analyze a simple algorithm for robust mean estimation in the Huber contamination model \citep{huber1964robust,huber1981robust}. 
Our algorithm (\cref{algorithm:robust-mean}) computes a mean-of-medians, which is the reverse of the extensively-studied median-of-means estimator \citep[among numerous others]{nemirovskij1983problem,jerrum1986random,DBLP:conf/icml/LaforgueSC21}. 
\cref{algorithm:robust-mean} uses an additional symmetry assumption on the distribution, and obtains a slower convergence rate than the standard median-of-means of estimator. 
The advantage of \cref{algorithm:robust-mean}, is that it has better robustness to arbitrary noise, as illustrated in the following section. 

We remark that \cref{algorithm:robust-mean} was recently studied by \cite*{DBLP:conf/nips/ZhongHYW21} for heavy-tailed distributions; however, they did not analyze its robustness to arbitrary noise, as we do here.\footnote{%
    The mean-of-medians algorithm was also used by \cite{DBLP:conf/nips/0004WWY023}.
} 
Additionally, we note that one could also use other more advanced techniques for robust mean estimation. For example, the algorithm of \cite*{DBLP:conf/nips/NovikovST23} gives better convergence performance. However, these more-advanced algorithms are considerably more complex. In the context of designing a security mechanism as in this paper, the simplicity of \cref{algorithm:robust-mean} can be a significant advantage.

\subsection{Why not use the standard Median-of-Means estimator?}
\label{section:why-not-standard-mom}

The standard median-of-means estimator partitions a sample of size $m=kb$ into $k$ batches of size $b$ each. For each batch $i \in [k]$, it computes the empirical average of the samples in the batch. Finally, it outputs the median of these averages. Namely, the estimator is 
\begin{equation}
    \label{eq:mom-estimator}
    \widehat{\mu} = \opFunction{median}{\frac{1}{b}\sum_{j \in [b]}x^{1}_j,\dots,\frac{1}{b}\sum_{j \in [b]}x^{k}_j},
\end{equation}
where batch $i$ is $x^i = (x^i_1,\dots,x^i_b)$.

To see why we opt not to use this estimator, consider a mixture distribution 
\[
    \cD = \frac{2}{3}\cdot U + \frac{1}{3}\cdot\cQ,
\] 
where $U = \uniform{\{-1,1\}}$ is the uniform distribution on $\{-1,1\}$, and $\cQ$ is a distribution of arbitrary noise, say $\PPP{x \sim \cQ}{x = c} = 1$ for some large $c$. The location parameter we are interested in recovering is $\mu = \EE{U} = 0$. We show that $\widehat{\mu}$ as in \cref{eq:mom-estimator} does not converge to $\mu$, specifically,
\[
    \PP{|\widehat{\mu}-\mu| \geq 1} \xrightarrow{m \to \infty} 1.
\]
Consider two cases.
\begin{itemize}
    \item{Case I: $b = 1$.
        In this case, $\widehat{\mu}$ is a median of $m$ values, each of which is in $\{-1,1,c\}$. If $m$ is odd, then $\widehat{\mu} \in \{-1,1,c\}$, so $\PP{|\widehat{\mu}-\mu| \geq 1} = 1$. Otherwise, if $m$ is even, and we define the median as some value in the range $[x_{t},x_{t+1}]$ for $t = \left\lfloor t/2 \right\rfloor$, then $\PP{|\widehat{\mu}-\mu| = 1} \geq \PP{x_{t} = x_{t+1} = 1} \xrightarrow{m \to \infty} 1$.
    }
    \item{
        Case II: $b \geq 2$. In this case, for each $i \in [k]$, batch $i$ contains at least one sample from $\cQ$ with probability $1-(2/3)^b > \frac{1}{2}$. Hence, as $\PP{B} \xrightarrow{m \to \infty} 1$, where $B$ is the event that a strict majority of the batches each contain at least one item equal to $c$. This implies that $\PP{|\widehat{\mu}-\mu| \geq \frac{c-b+1}{b}} \xrightarrow{m \to \infty} 1$ for $c$ arbitrarily large.
    }
\end{itemize}
We see that in both cases, the median-of-means does not converge to $\mu = 0$. This limitation is overcome by \cref{algorithm:robust-mean}, as we show in the next section.

\subsection{The Mean-of-Medians estimator}

\begin{algorithmFloat}[!ht]
    \begin{ShadedBox}
        \textbf{Assumptions:}
        \begin{itemize}
            \item{
                $\cS = \left\{x_1,\dots,x_{|\cS|}\right\} \subseteq \bbR$ is finite.
            }
        \end{itemize}

        \vspp

        \textsc{MeanOfMedians}($\cS$):
        \vsp
        \begin{algorithmic}
            \State $b \gets \left\lfloor\! \sqrt{|\cS|} \right\rfloor$
            \Comment{There are $b$ batches, each of size $b$}
            \vsp

            \For $i \in [b]$:
                \vsp
                \State $m_i \gets $ median of $\left\{
                    x^i_1,x^i_2,\dots,x^i_b    
                \right\}$, where $x^{i}_j = x_{(i-1)b+j}$
                \vsp

            \EndFor

            \State \textbf{output} $\frac{1}{|b|}\sum_{i \in [b]}m_i$
        \end{algorithmic}
    \end{ShadedBox}
    \caption{A simple algorithm for robust mean estimation.}
    \label{algorithm:robust-mean}
\end{algorithmFloat}

\begin{theorem}
    \label{theorem:robust-mean}
    Let $\cP,\cQ \in \distribution{\bbR}$ be distributions, let $\alpha \geq 0$, and let 
    \[
        \cD = (1-\alpha)\cdot\cP + \alpha\cdot\cQ
    \] 
    be a mixture distribution. Assume that:
    \begin{enumerate}[label=(\alph*)]
        \item{
            $\cP$ has mean $\mu \in \bbR$, and there exist $B, \beta \geq 0$ such that $\PPP{x \sim \cP}{|x - \mu| > B} < \beta$.
        }
        \item{
            \label{item:alpha-beta-small}
            $\alpha$ and $\beta$ are small, such that $(1-\alpha)(1-\beta) \geq 2/3$.
        }
        \item{
            $\cD$ is symmetric about $\mu$, namely, for any measurable set $A \subseteq \bbR$, $\cD(\mu+A) = \cD(\mu-A)$, where $\mu\pm A = \{\mu \pm a: ~ a \in A\}$.
        }
    \end{enumerate} 
    Let $m \in \bbN$, let $\cS \sim \cD^m$, and let 
    \[
        \widehat{\mu} = \emph{\textsc{MeanOfMedians}}(\cS)
    \]
    as in \cref{algorithm:robust-mean}. Then:
    \begin{enumerate}
        \item{
            \label{item:robust-mean-unbiased}
            $\widehat{\mu}$ is unbiased, i.e., $\EEE{\cS \sim \cD^m}{\widehat{\mu}} = \mu$.
        }
        \item{
            \label{item:robust-mean-concentrated}
            $\widehat{\mu}$ is concentrated, i.e.,
            $
                \displaystyle
                \forall \varepsilon \geq 0: ~ \PPP{\cS \sim \cD^m}{\Big| \widehat{\mu} - \mu \Big| \geq \varepsilon} 
                \leq
                4\expf{-\gamma \sqrt{m}}
            $ for $\gamma = \min\left\{\frac{1}{100}, \frac{2\varepsilon^2}{B^2}\right\}$ and $m$ large enough.
        }
    \end{enumerate}
\end{theorem}
\vsp

Note that in the theorem statement, $\cQ$ might not have a mean.

\subsubsection{Proof of \texorpdfstring{%
        \cref{theorem:robust-mean}
    }{%
        Robust Mean Estimation
    }%
}

\begin{proof}[Proof of \cref{theorem:robust-mean}]
    First, we show that $\widehat{\mu}$ is unbiased. Fix $i \in [b]$. Seeing as $\cD$ is symmetric about $\mu$, for any measurable $A \subseteq \bbR$, $\cD(A) = \cD(2\mu-A)$. Consequently, for every $j \in [b]$, $x^i_j \stackrel{d}{=} 2\mu - x^i_j$. Because the $x^i_j$ variables are independent, this implies that 
    \begin{align}
        \label{eq:symmetry-argument-median}
        \EE{m_i} 
        &= 
        \EEE{x^{\:i}_{1:b}\sim \cD^b}{
            \opFunction{median}{x^i_1,\dots,x^i_b}
        }
        \nonumber
        \\
        &= 
        \EEE{x^{\:i}_{1:b}\sim \cD^b}{
            \opFunction{median}{2\mu - x^i_1,\dots,2\mu - x^i_b}
        }
        \tagexplain{$x^i_j \stackrel{d}{=} 2\mu - x^i_j$}
        \nonumber
        \\
        &= 
        \EEE{x^{\:i}_{1:b}\sim \cD^b}{
            2\mu-
            \opFunction{median}{x^i_1,\dots,x^i_b}
        }
        \nonumber
        \\
        &=
        2\mu-\EE{m_i}.
    \end{align}
    Hence, 
    \begin{equation}
        \label{eq:mi-unbiased}
        \forall i \in [b]: ~ \EE{m_i} = \mu,
    \end{equation}
    so $\EE{\frac{1}{|b|}\sum_{i \in [b]}m_i} = \mu$, establishing \cref{item:robust-mean-unbiased} in the theorem.

    Second, we show that $\widehat{\mu}$ is concentrated. We can express the distribution $\cD$ as a mixture
    \begin{equation}
        \label{eq:mixture-good-bad}
        \cD = (1-\alpha)(1-\beta)\cdot\cP^{\good} + (1-\alpha)\beta\cdot\cP^{\bad} + \alpha\cdot\cQ,
    \end{equation}
    where $\cP^{\good} = \left(\cP~\Big|~ |x - \mu| \leq B\right)$ and $\cP^{\bad} = \left(\cP~\Big|~ |x - \mu| > B\right)$.
    
    Fix $i \in [b]$. We may assume that for every $j \in [b]$ there is an indicator $g^i_j \sim \Ber{(1-\alpha)(1-\beta)}$ such that if $g^i_j = 1$ then $x^i_j \sim \cP^{\good}$, and if $g^i_j = 0$ then $x^i_j \propto (1-\alpha)\beta\cdot\cP^{\bad} + \alpha\cdot\cQ$. 
    
    Observe that if a strict majority of $\{x^i_j: ~ j \in [b]\}$ is in the interval $I_{\mu \pm B} = [\mu - B, \mu + B]$, then the median $m_i$ is in $I_{\mu \pm B}$ as well. Namely,
    \[
        \PP{m_i \in I_{\mu \pm B} ~ \big| ~ E_i} = 1,
    \]
    where $E_i$ is the event where $\left|\{j \in [b]: ~ x^i_j \in I_{\mu \pm B}\}\right|/b > \nicefrac{1}{2}$.
    \begin{align}
        \label{eq:majority-xij-in-interval}
        \PP{\neg E_i}
        &\leq 
        \PP{\sum_{j \in [b]}g^i_j \leq \frac{1}{2}} 
        \nonumber
        \\
        &=
        \PP{\left|
            \EE{\sum_{j \in [b]}g^i_j}-\sum_{j \in [b]}g^i_j
        \right| \geq \frac{1}{6}}
        \tagexplain{By \cref{item:alpha-beta-small}}
        \nonumber
        \\
        &\leq
        2\expf{-2b\cdot\left(\frac{1}{6}\right)^2}.
        \tagexplain{Hoeffding's inequality}
    \end{align}
    Observe that $E_i$ is an event that is symmetric with respect to $\mu$. Formally, if $x^i = (x^i_1,\dots,x^i_b)$ and $2\mu-x^i = (2\mu-x^i_1,\dots,2\mu-x^i_b)$, then
    \begin{equation}
        \label{eq:ei-symmetric}
        x^i \in E_i ~ \iff ~
        (2\mu-x^i) \in E_i.
    \end{equation}
    This implies that the distribution $\cD^b|E_i$ is symmetric about $\mu$. Namely, for any measurable $A \in \bbR^b$, %
    \begin{align*}
        \PPP{x^i \sim \cD^b}{x^i \in A ~ | ~ E_i}
        &=
        \frac{
            \PPP{x^i \sim \cD^b}{x^i \in A \cap E_i}
        }{
            \PPP{x^i \sim \cD^b}{x^i \in E_i}
        }
        \\
        &=
        \frac{
            \PPP{x^i \sim \cD^b}{(2\mu-x^i) \in A \cap E_i}
        }{
            \PPP{x^i \sim \cD^b}{x^i \in E_i}
        }
        \tagexplain{Symmetry of $\cD$ about $\mu$}
        \\
        &=
        \frac{
            \PPP{x^i \sim \cD^b}{(2\mu-x^i) \in A  ~ \land ~  (2\mu-x^i) \in E_i}
        }{
            \PPP{x^i \sim \cD^b}{x^i \in E_i}
        }
        \\
        &=
        \frac{
            \PPP{x^i \sim \cD^b}{x^i \in (2\mu-A)  ~ \land ~  x^i \in E_i}
        }{
            \PPP{x^i \sim \cD^b}{x^i \in E_i}
        }
        \tagexplain{By \cref{eq:ei-symmetric}}
        \\
        &=
        \PPP{x^i \sim \cD^b}{x^i \in (2\mu-A) ~ | ~ E_i}
        \\
        &=
        \PPP{x^i \sim \cD^b}{(2\mu-x^i) \in A ~ | ~ E_i}.
    \end{align*}
    Namely, $x^i|E_i \stackrel{d}{=} (2\mu-x^i)|E_i$.
    By the same argument as in \cref{eq:symmetry-argument-median}, this implies that 
    \begin{equation}
        \label{eq:mean-mi-given-ei}
        \EE{m_i ~ | ~ E_i} = \mu.
    \end{equation}
    Let $E = \cap_{i \in [b]}E_i$. Then for any $\varepsilon \geq 0$
    \begin{align*}
        \PP{\left|\mu - \frac{1}{|k|}\sum_{i \in [k]}m_i\right| \geq \varepsilon} 
        &\leq
        \PP{\neg E} 
        +
        \PP{\left|\frac{1}{|k|}\sum_{i \in [k]}(m_i-\mu)\right| \geq \varepsilon ~ \Big| ~ E\:}
        \\
        &\leq
        \PP{\neg E} 
        + 
        2\expf{-\frac{2b\varepsilon^2}{B^2}}
        \tagexplain{Hoeffding's inequality, \cref{eq:mean-mi-given-ei}}
        \\
        &\leq
        2b\cdot \expf{-\frac{b}{18}} + 2\expf{-\frac{2b\varepsilon^2}{B^2}}
        \tagexplain{By \cref{eq:majority-xij-in-interval}, union bound}
        \\
        &\leq
        2\cdot \expf{-\frac{b}{100}} + 2\expf{-\frac{2b\varepsilon^2}{B^2}}
        \tagexplain{For $b \geq 102$}
        \\
        &\leq
        4\expf{-\gamma b},
    \end{align*}
    where $\gamma = \min\left\{\frac{1}{100}, \frac{2\varepsilon^2}{B^2}\right\}$, as desired.
\end{proof}

    \else
        
    \fi

    \ifdraftcompile
        \section{Future Directions}
\label{section:future-work}

We have provided preliminary evidence that techniques based on random self-reducibility can be effective at backdoor mitigation. There are two directions for future works that could try to advance this idea towards practical applications. 

One central direction is to search for additional families of distributions $\bbD$ with random self-reducibility properties that are suitable for backdoor mitigation. We have seen that if the label distribution is close to a linear, polynomial, or $\tau$-heavy function, then secure backdoor mitigation is possible. But do there exist broader families of distributions that appear commonly in real-world data and are also conducive for backdoor mitigation?

A second avenue for exploration is to take advantage of the representation of the proposed ML model $\tilde{f}$. Our constructions treat $\tilde{f}$ as a black-box, and make no assumptions on how it is implemented or represented. But perhaps one could obtain better results by either \emph{(i)} assuming that $\tilde{f}$ belongs to some class of functions, e.g., it is implemented by a neural network of a certain architecture; or, moreover \emph{(ii)} using whitebox access to the representation of $\tilde{f}$ (the code or parameters of the model) in order to obtain more sophisticated forms of random self-reducibility?

    \else
        \section{Future Directions}
\label{section:future-work}

We have provided preliminary evidence that techniques based on random self-reducibility can be effective at backdoor mitigation. There are two directions for future works that could try to advance this idea towards practical applications. 

One central direction is to search for additional families of distributions $\bbD$ with random self-reducibility properties that are suitable for backdoor mitigation. We have seen that if the label distribution is close to a linear, polynomial, or $\tau$-heavy function, then secure backdoor mitigation is possible. But do there exist broader families of distributions that appear commonly in real-world data and are also conducive for backdoor mitigation?

A second avenue for exploration is to take advantage of the representation of the proposed ML model $\tilde{f}$. Our constructions treat $\tilde{f}$ as a black-box, and make no assumptions on how it is implemented or represented. But perhaps one could obtain better results by either \emph{(i)} assuming that $\tilde{f}$ belongs to some class of functions, e.g., it is implemented by a neural network of a certain architecture; or, moreover \emph{(ii)} using whitebox access to the representation of $\tilde{f}$ (the code or parameters of the model) in order to obtain more sophisticated forms of random self-reducibility?

    \fi

\hfil

\paragraph{Acknowledgments.} JS, NV and VV were supported in part by NSF CNS-2154149 and a Simons Investigator Award. JS would like to thank Ido Nachum and Shay Moran for helpful conversations. 

\hfil

\newpage

\phantomsection

\addcontentsline{toc}{section}{References}
\bibliographystyle{plainnat}
\bibliography{paper}

\appendix

\phantomsection

\addcontentsline{toc}{section}{Appendices}

    \ifdraftcompile

\section{Goldreich--Levin Theorem}\label{appendix:goldreich-levin}

\begin{theorem}[\citealp{DBLP:conf/stoc/GoldreichL89}; Section 3.5 in \citealp{DBLP:books/daglib/0033652}]
    \label{theorem:goldreich-levin}
    There exist an algorithm $A$ and a polynomial $p$ as follows. For any $n \in \bbN$, any $\tau,\delta \in (0,1]$, and any function $f:~ \pmo^n \to \pmo$, if $A$ is executed with oracle access to $f$ then $A$ terminates in time $p(n,1/\tau, \log(1/\delta))$ and outputs a list $L \subseteq 2^{[n]}$ such that with probability at least $1-\delta$, for all $S \subseteq [n]$:
    \begin{enumerate}
        \item{
            $\left|\widehat{f}(S)\right| \geq \tau ~ \implies ~ S \in L$, and
        }
        \item{
            \label{item:gl-lower-bound}
            $S \in L ~ \implies ~ \left|\widehat{f}(S)\right| \geq 3\tau/4$.
        }
    \end{enumerate}
\end{theorem}

Note that the theorem is typically stated with a constant of $\tau/2$ in \cref{item:gl-lower-bound}, while we have chosen a stronger statement with a constant of $3\tau/4$. The theorem holds for any constant fraction of $\tau$.

\section{Miscellaneous Fourier Analysis Results}

We use the norm notation $\|\cdot\|$ defined in \cref{section:fourier-analysis-preliminaries}.

\begin{fact}
    \label{fact:loss-expectation}
    Let $\cD \in \distribution{\cX \times \pmo}$, and let $f:\cX \to \pmo$. Then,
    \begin{equation*}
        \Loss{\cD}{f}=\PPP{(x,y) \sim \cD}{f(x) \neq y} = \frac{1}{4}\cdot\EEE{(x,y) \sim \cD}{\left(f(x) - y\right)^2} = \frac{1}{4}\left\|f - y\right\|^2.
    \end{equation*}
\end{fact}

\begin{claim}
    \label{lemma:low-loss-functions-fourier-close}
    Let $\varepsilon \in [0,1]$, let $\cX = \pmo^n$, let $\cD \in \distribution{\cX \times \pmo}$ have uniform marginal on $\cX$, let $h: ~ \cX \to \pmo$ such that $\Loss{\cD}{h} \leq \varepsilon$, and let $r: ~ \cX \to \bbR$ such that $\EEE{(x,y) \sim \cD}{\left(r(x)-y\right)^2} \leq 4\varepsilon$. Then
    \[
        \sum_{S \subseteq [d]} \left(\widehat{r}(S)-\widehat{h}(S)\right)^2 \leq 16\varepsilon.
    \]
\end{claim}

\begin{proof}[Proof of \cref{lemma:low-loss-functions-fourier-close}]
    \begin{align*}
        \left\|r(x)-h(x)\right\|
        &\leq
        \left\|r(x)-y\right\|
        +
        \left\|h(x)-y\right\|
        \\
        &=
        \left(
            \EEE{(x,y) \sim \cD}{\left(r(x)-y\right)^2}
        \right)^{1/2}
        +
        \left(
            \EEE{(x,y) \sim \cD}{\left(h(x) - y\right)^2}
        \right)^{1/2}
        \\
        &=
        \left(
            \EEE{(x,y) \sim \cD}{\left(r(x)-y\right)^2}
        \right)^{1/2}
        +
        2\left(
            \PPP{(x,y) \sim \cD}{h(x) \neq y}
        \right)^{1/2}
        \tagexplain{By \cref{fact:loss-expectation}}
        \\
        &\leq
        4\sqrt{\varepsilon}.
        \tagexplain{Choice of $r$ and $h$}
    \end{align*}
    Hence,
    \begin{align*}\label{eq:loss-fourier-bound}
        16\varepsilon
        &\geq
        \left\|r(x)-h(x)\right\|^2
        \\
        &=
        \sum_{S \subseteq [d]} \left(\widehat{r}(S)-\widehat{h}(S)\right)^2.
        \tagexplain{Parseval's identity}
        \qedhere
    \end{align*}
\end{proof}

\begin{claim}[Section 4 in \citealp{mansour1994learning}]
    \label{lemma:remove-sign-loss}
    Let $d \in \bbN$, let $\cX$ be a set, let $\cD \in \distribution{\cX \times \pmo}$, and let $r:~ \cX \to \bbR$. Then
    \[
        \PPP{(x,y)\sim\cD}{
            \sign(r(x)) \neq y
        }
        \leq 
        \EEE{(x,y)\sim\cD}{
            \left(
                r(x)
                -
                y
            \right)^2
        }.
    \] 
\end{claim}

\begin{proof}[Proof of \cref{lemma:remove-sign-loss}]
    For a fixed $x \in \cX$ and $y \in \pmo$, consider two cases.
    \begin{itemize}
        \item{
            Case I: $\sign(r(x)) \neq y$. Then $\1(\sign(r(x)) \neq y) = 1 \leq |r(x) - y| \leq \left(r(x) - y\right)^2$.
        }
        \item{
            Case II: $\sign(r(x)) = y$. Then $\1(\sign(r(x)) \neq y) = 0 \leq \left(r(x) - y\right)^2$.
        }
    \end{itemize}
    Hence, for any $x \in \cX$ and $y \in \pmo$,
    \[
        \1(\sign(r(x)) \neq y) \leq \left(r(x) - y\right)^2.    
    \]
    Therefore,
    \begin{align*}
        \PPP{(x,y)\sim\cD}{
            \sign(r(x)) \neq y
        }
        &=
        \EEE{(x,y)\sim\cD}{
            \1\left(\sign(r(x)) \neq y\right)
        }
        \leq 
        \EEE{(x,y)\sim\cD}{
            \left(r(x) - y\right)^2
        }.
        \qedhere
    \end{align*}
\end{proof}

\begin{claim}\label{lemma:ell-2-loss-vs-zero-one}
    For all sets $\cX$, all distributions $\cD \in \distribution{\cX \times \{\pm 1\}}$, and all functions $f : \cX \to \bbR$, 
    \[ \frac{1}{4} \LossSquare{\cD}{\sign(f)} = \LossZeroOne{\cD}{\sign(f)} \leq \LossSquare{\cD}{f}.\]
\end{claim}
\begin{proof}
    The (left-hand) equality directly holds, as
    \begin{align*}  \frac{1}{4} \LossSquare{\cD}{\sign(f)} = \frac{1}{4} \EEE{(x,y)\sim\cD}{\left(\sign(f(x)) - y\right)^2} &=\EEE{(x,y)\sim\cD}{\indicator{\sign(f(x))\neq y}} 
    \\&=\PPP{(x,y)\sim\cD}{\sign(f(x))\neq y}
    \\&= \LossZeroOne{\cD}{\sign(f)},
    \end{align*}
    where we have used the fact that $(\sign(f(x)) - y)^2 \in \{0, 4\}$. The (right-hand) inequality is given in~\cref{lemma:remove-sign-loss}.
\end{proof}

\section{Subgaussian distributions}

\begin{definition}
    Let $\sigma \geq 0$ and let $X$ be a real-valued random variable. $X$ is \ul{subgaussian with variance proxy $\sigma^2$}, denoted $X \in \SubG{\sigma^2}$, if $\EE{X} = 0$ and
    \[
        \forall t \in \bbR: ~ \EE{e^{tX}} \leq e^{\frac{\sigma^2t^2}{2}}.
    \]
\end{definition}

\begin{claim}[Concentration for Subgaussian Random Variables]
    \label{claim:concentration-sum-of-subgaussians}
    Let $\sigma \geq 0$, and $X \in \SubG{\sigma^2}$. Then for any $t \geq 0$,
    \[
        \PP{\left|X\right| \geq t} 
        \leq
        2\expf{-\frac{t^2}{2\sigma^2}}.
    \] 
    Moreover, for $n \in \bbN$, independent variables $X_1,\dots,X_n \in \SubG{\sigma^2}$, and for any $a_1,\dots,a_n \in \bbR$ and $t \geq 0$,
    \[
        \PP{\left|\sum_{i \in [n]}a_iX_i\right| \geq t} 
        \leq
        2\expf{-\frac{t^2}{2\sigma^2\sum_{i \in n}a_i^2}}.
    \]
\end{claim}

\begin{claim}[Sum of Subgaussian Random Variables is Subgaussian]
    \label{claim:sum-of-subgaussians}
    Let $\sigma \geq 0$, $n \in \bbN$, and let $X_1,\dots,X_n \in \SubG{\sigma^2}$ for all $i \in [n]$. Then $Z = \sum_{i \in [n]}X_i \in \SubG{n^2\sigma^2}$.
\end{claim}

Note that the $X_i$'s in the claim are not necessarily independent.

\begin{proof}[Proof of \cref{claim:sum-of-subgaussians}]
    For any $t \in \bbR$,
    \begin{align*}
        \EE{e^{tZ}}
        &=
        \EE{e^{t(X_1 + \cdots + X_n)}}
        \\
        &=
        \EE{e^{
            \frac{1}{n}\sum_{i \in [n]}ntX_i}
        }
        \\
        &\leq 
        \frac{1}{n} \sum_{i \in [n]} \EE{e^{ntX_i}}
        \tagexplain{Jensen's inequality}
        \\
        &\leq 
        \frac{1}{n} \sum_{i \in [n]} e^{
            \frac{
                \sigma^2\cdot (nt)^2
            }{2}
        }
        =
        e^{
            \frac{
                n^2\sigma^2\cdot t^2
            }{2}
        },
        \tagexplain{$X_i \in \SubG{\sigma^2}$}
    \end{align*}
    as desired.
\end{proof}

\begin{claim}[Product of Subgaussian and Bounded Random Variables is Subgaussian]
    \label{claim:subgaussian-times-bounded}
    Let $\sigma,c \geq 0$, and let $X,Y \in \bbR$ be random variables such that $X \in \SubG{\sigma^2}$, and $\PP{|Y| \leq c} = 1$. Then $Z = XY \in \SubG{c^2\sigma^2}$.
\end{claim}

Note that $X$ and $Y$ need not be independent.

\begin{proof}[Proof of \cref{claim:subgaussian-times-bounded}]
    For any $t \in \bbR$,
    \begin{align*}
        \EE{e^{tZ}}
        &=
        \EE{e^{tXY}}
        \\
        &=
        \EEE{y \sim Y}{
            \EEE{X}{
                e^{(ty)\cdot X}
            }
        }
        \\
        &\leq
        \EEE{y \sim Y}{
            e^{\frac{\sigma^2\cdot(ty)^2}{2}}
        }
        \tagexplain{$X \in \SubG{\sigma^2}$}
        \\
        &\leq
        e^{\frac{c^2\sigma^2 t^2}{2}},
        \tagexplain{$|Y| \leq c$}
    \end{align*}
    as desired.
\end{proof}

    \else
        
    \fi

\end{document}